\documentclass[3p,a4paper]{elsarticle}
\usepackage[utf8]{inputenc}
\usepackage{subfigure}

\title{ Scientific Machine Learning with Kolmogorov-Arnold Networks 
}

\author[myUaddress]{Salah A. Faroughi \corref{mycorrespondingauthor}}
\author[myUaddress]{Farinaz Mostajeran}
\author[myUaddress]{Amin Hamed Mashhadzadeh}
\author[myUaddress2]{Shirko Faroughi}

\cortext[mycorrespondingauthor]{Corresponding author: salah.faroughi@utah.edu}

\address[myUaddress]{Energy \& Intelligence Lab, Department of Chemical Engineering, University of Utah, Salt Lake City, Utah  84112, USA
}

\address[myUaddress2]{Department of Mechanical Engineering, School of Engineering, Urmia University of Technology, Urmia, Iran
}

\date{\today}

\usepackage{mwe}

\usepackage{setspace} 
\usepackage{tabularx}
\usepackage{lineno,hyperref}
\usepackage{amsmath}
\usepackage{bm}
\usepackage{amssymb}
\usepackage[]{amsmath}
\usepackage{upgreek}
\usepackage{float}

\let\today\relax
\makeatletter
\def\ps@pprintTitle{%
    \let\@oddhead\@empty
    \let\@evenhead\@empty
    \def\@oddfoot{\footnotesize\itshape
         {Submitted preprint — October 2025} \hfill\today}%
    \let\@evenfoot\@oddfoot
    }
\makeatother

\usepackage{pgfplots}
\pgfplotsset{compat=1.5}
\usepgfplotslibrary{units}
\usetikzlibrary{spy}
\usetikzlibrary{pgfplots.groupplots}

\usepackage{epstopdf}
\usepackage{units}
\usepackage{lscape}
\usepackage{booktabs}
\usepackage{gensymb}
\usepackage{multirow}

\usepackage{graphicx}
\usepackage{caption}
\usepackage{subcaption}

\usepackage{fontawesome5}

\usepackage{booktabs}
\usepackage{multirow}
\usepackage{caption}
\usepackage{subcaption}

\usepackage{soul,color}
\soulregister\cite7
\soulregister\ref7
\soulregister\pageref7

\usepackage[flushleft]{threeparttable}
\usepackage{tikz}
\usepackage{placeins}
\usepackage{color}
\usepackage{hyperref}

\usepackage{stackengine}

\usepackage{array}
\usepackage{tabularx}
\usepackage{pdfpages}
\usepackage{natbib}
\biboptions{numbers,sort,compress} 
\usepackage{textcomp, gensymb}
\usepackage{booktabs} 

\definecolor{mydarkgreen}{RGB}{10,200,50} 

\begin{document}

\begin{abstract}

The field of scientific machine learning, which originally utilized multilayer perceptrons (MLPs), is increasingly adopting Kolmogorov–Arnold Networks (KANs) for data encoding. This shift is driven by the limitations of MLPs, including poor interpretability, fixed activation functions, and difficulty capturing localized or high-frequency features. KANs address these issues with enhanced interpretability and flexibility, enabling more efficient modeling of complex nonlinear interactions and effectively overcoming the constraints associated with conventional MLP architectures.  This review categorizes recent progress in KAN-based models across three  distinct perspectives: (i) data-driven learning, (ii) physics-informed modeling, and (iii) deep-operator learning. Each perspective is examined through the lens of architectural design, training strategies, application efficacy, and comparative evaluation against MLP-based counterparts.  By benchmarking KANs against MLPs, we highlight consistent improvements in accuracy, convergence, and spectral representation, clarifying KANs' advantages in capturing complex dynamics while learning more effectively. In addition to reviewing recent literature, this work also presents several comparative evaluations that clarify central characteristics of KAN modeling and hint at their potential implications for real-world applications. Finally, this review identifies critical challenges and open research questions in KAN development, particularly regarding computational efficiency, theoretical guarantees, hyperparameter tuning, and algorithm complexity. We also outline future research directions aimed at improving the robustness, scalability, and physical consistency of KAN-based frameworks.

\end{abstract}


\maketitle

\section{Introduction}\label{sec:Intro}

\begin{figure}[h!]
    \centering
    \includegraphics[width=1\linewidth]{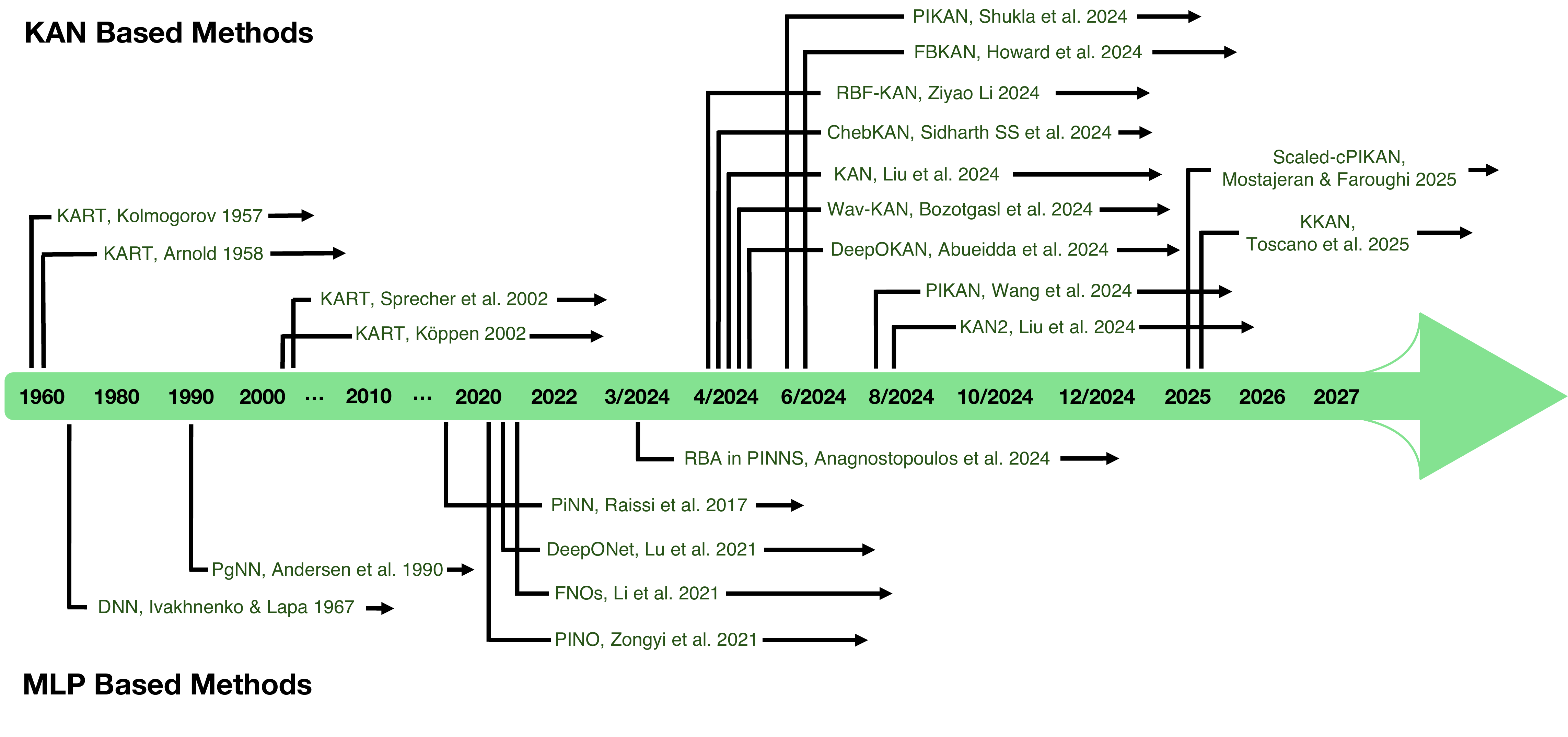}
    \caption{A schematic illustration highlighting the pivotal contributions in SciML that are established using MLPs or KANs. For MLP-based architectures, the following are listed: deep neural network (DNN)~\cite{ivakhnenko1967cybernetics}, physics-guided neural network (PgNN)~\cite{andersen2002artificial}, physics-informed neural network (PiNN)~\cite{raissi2019physics}, DeepONet~\cite{lu2021learning}, and Fourier neural operator~\cite{Li2020FNO}. For KAN-based architectures, the following are listed: Kolmogorov–Arnold representation theorem, Kolmogorov–Arnold networks, KART~\cite{kolmogorov1957representations, Arnold1958, sprecher2002space, koppen2002training}, RBF-KAN~\cite{li2024kolmogorov}, ChebKAN~\cite{ss2024chebyshev}, KAN~\cite{liu2024kan}, Wav-KAN~\cite{bozorgasl2405wav}, DeepOKAN~\cite{abueidda2025deepokan}, PIKAN~\cite{shukla2024comprehensive, wang2024kolmogorov}, FBKAN~\cite{howard2024finite},  KAN 2.0~\cite{liu2024kan2}, Scaled-cPIKAN~\cite{mostajeran2025scaled}, KKAN \cite{toscano2025kkans}.
    }
    \label{fig:PIKAN_revolution}
\end{figure}

Scientific Machine Learning (SciML) is rapidly transforming the landscape of scientific computing by integrating machine learning with physicochemical first principles, mathematical models, and observational  data~\cite{thiyagalingam2022scientific, raissi2019physics, karniadakis2021physics, raissi2017physicsinformeddeeplearning, cai2021physics,  doi:10.1137/24M1643232, doi:10.1137/24M1646455}. It enables the development of flexible, data-efficient models that capture complex and non-linear behaviors across a wide range of natural and industrial systems, with successful applications in science and engineering; e.g., fluid mechanics~\cite{jin2021nsfnets, MAO2020112789, JAGTAP2022111402, faroughi2022meta, pawar2024geo, LOU2021110676, SHU2023111972, OVADIA2025117982, faroughi2024physics}, solid mechanics~\cite{jin2023recent, haghighat2020deep, zhu2025extended, zeng2024adaptive,  kumar2022machine},  materials science~\cite{CIESIELSKI2025113417, datta2023multihead, kiyani2025predicting, NIELSEN2002177, kiyani2025crack}, climate modeling~\cite{PARTEE2022101707, Eyring2024, doi:10.1098/rsta.2020.0093, soltanmohammadi2023comparative, pawar2024esm,  NEURIPS2023_45fbcc01, ehteram2024read, mostajeran2025context}, and  other disciplines~\cite{li2022scientific, bedolla2020machine, tapeh2023artificial, celaya2025adaptive, celaya2025learning, celaya2024solutions, yang2025pinn, maceachern2021machine, mostajeran2025solving, he2024phase, darcet2023vision, grandits2024neural}. Traditionally, the backbone of most SciML methods has been multilayer perceptrons (MLPs)~\cite{JAGTAP2020109136, li2024physics}. Despite their success and widespread use, MLP-based approaches face persistent challenges, including poor interpretability~\cite{cranmer2023interpretable}, limited expressivity due to fixed activation functions~\cite{glorot2010understanding, goodfellow2016deep, raghu2017expressive}, and difficulty in capturing complex, nonlinear dependencies~\cite{hirsch2024neural, cruz2025state, mostajeran2024epi}, requiring deeper networks. These challenges have sparked growing interest in Kolmogorov–Arnold Networks (KANs), a class of neural architectures inspired by the Kolmogorov–Arnold representation theorem~\cite{kolmogorov1957representations, Arnold1958, arnold1959representation, liu2024kan}. Research in KANs can be broadly divided into three major modeling approaches: data-driven learning, which relies solely on observational or experimental data~\cite{kou2021data, bahramian2023data}; physics-informed learning, which incorporates governing equations into the training process~\cite{wang2024kolmogorov, howard2024finite, toscano2024inferring}; and deep-operator learning, which focuses on  mappings between infinite-dimensional function spaces~\cite{abueidda2025deepokan, kiyani2025predicting}. Fig.~\ref{fig:PIKAN_revolution} presents a schematic illustration highlighting the pivotal contributions in SciML that are established using MLPs or KANs. 

The Kolmogorov–Arnold Representation Theorem (KART)~\cite{kolmogorov1957representations, Arnold1958, arnold1959representation} establishes that any continuous multivariate function can be represented through nested compositions of continuous univariate functions. KART demonstrates how complex multidimensional relationships can be decomposed into simpler one-dimensional transformations. While KART provides a fundamental mathematical framework for function representation, its original formulation is non-constructive, the theorem proves existence but doesn't provide practical methods for determining the required univariate functions. This limitation makes direct application in machine learning challenging. To bridge this gap, KANs were introduced as a trainable neural architecture that mirrors the functional form of KART~\cite{liu2024kan, liu2024kan2}. In KANs, each univarite function in the decomposition is parametrized using basis expansions with coefficients learned from data. This setup transforms the symbolic structure into a computation graph where nonlinear operations are placed along edges, and summations are carried out at nodes~\cite{liu2024ikan, li2024kolmogorov, yang2024kolmogorov}. The resulting model can be interpreted as a two-layer network aligned with the original Kolmogorov–Arnold construction~\cite{girosi1989representation, ismailov2014approximation, schmidt2021kolmogorov, poluektov2023construction}. However, this shallow structure may be insufficient for learning complex mappings in practice, especially when smooth basis functions are used. To overcome this limitation, KANs have been generalized to deeper and wider configurations, extending their expressivity while preserving the core idea of dimension-wise adaptive function learning~\cite{basina2024kat, pant2025mlps}. KANs have since been applied to a wide range of problems in science and engineering, including data-driven discovery of physical laws~\cite{xu2024fourierkan, peng2024predictive, genet2024temporal}, solution of partial differential equations~\cite{shukla2020physics, mostajeran2024epi, mostajeran2025scaled}, and learning of operators from data~\cite{abueidda2025deepokan, kiyani2025predicting}. Beyond these domains, KANs have also been explored for tasks in image processing~\cite{azam2024suitability, cheon2024kolmogorov, seydi2024exploring},  audio signal classification~\cite{yu2024kan}, and many more applications~\cite{li2025u, shuai2025physics, karakacs2024novel, seydi2024unveiling, cheon2024demonstrating, panahi2025data}, demonstrating the flexibility of their functional design across diverse application areas.

The integration of KANs into SciML highlights a growing alignment between model architecture and the nature of the learning task, whether data-driven, physics-informed, or operator-based. In each case, the training objectives directly influence both model design and optimization dynamics.
Data-driven modeling plays a central role when physical mechanisms are incomplete, unknown, or difficult to express through explicit governing equations~\cite{kou2021data, williams2021evolution, bahramian2023data, ghorbani2025using, zhou2025askan, mollaali2025conformalized}.  In such cases, models are trained only on observational or experimental data to identify patterns linking inputs, system states, and outputs, commonly across regression, classification, and segmentation tasks. While MLPs have traditionally served as the default architecture in these tasks, their dense, black-box nature can limit interpretability and training stability. KANs, with their structured composition of univariate functions~\cite{liu2024kan, shukla2024comprehensive, ss2024chebyshev}, provide more structured approximation strategies and theoretical insights into convergence, especially under limited-data regimes~\cite{mostajeran2024epi, ejaz2024can, howard2024finite, panahi2025data}. 
When observational data is limited or physical consistency is critical, physics-informed learning provides a complementary approach by embedding known physical laws, typically a set of coupled partial differential equations (PDEs) and closure models, into the training process. 
This is valuable in settings where the governing equations are known but difficult to solve due to irregular geometries, high dimensionality, or sparse and noisy measurements in inverse problems~\cite{pang2019fpinns, YANG2021109913, shukla2021parallel, shukla2020physics}. 
In frameworks such as Physics-informed Neural Networks (PINNs)~\cite{raissi2017physicsinformeddeeplearning}, these constraints are enforced by incorporating PDE (or other closure model) residuals into the loss function~\cite{ karniadakis2021physics, cai2021physics, kharazmi2021hp}. However, such integration often presents trade-offs involving numerical stability, expressivity, and computational efficiency~\cite{raissi2019physics, nabian2020adaptive, cuomo2022scientific, psaros2023uncertainty, kashefi2025physics}.
Physics-informed KANs (PIKANs) address these challenges by replacing traditional MLP backbones with the structured KAN architecture~\cite{toscano2025pinns, jacob2024spikans, daryakenari2025representation, shuai2025physics, patra2024physics, guo2024physics, faroughi2025neural, xiapikans, kiyani2025optimizer, shukla2024comprehensive}, allowing for more control over the solution space and potentially tighter alignment with the underlying physics. Complementing both paradigms, deep-operator learning has gained attention as a general framework for approximating mappings between infinite-dimensional function spaces, rather than pointwise mappings~\cite{wen2022u, azizzadenesheli2024neural}.  
Architectures such as DeepONet~\cite{lu2021learning, lu2021deepxde} implement this through a branch–trunk decomposition to separately encode input functions and output domains, enabling learning across varying functional inputs~\cite{lu2022multifidelity, goswami2022physics, kobayashi2024improved, garg2023vb}. While early implementations relied on densely parameterized feedforward networks, extensions like DeepOKAN adopt KAN-based operators within this framework, further improving the expressiveness and training efficiency of DeepONet-like models~\cite{abueidda2025deepokan, kiyani2025predicting, pant2025mlps, wang2025efkan, yu2025deepoheat, toscano2025kkans}. By leveraging the KAN structure at the operator level, these models reduce sample complexity and improve approximation across broad function classes, especially in multi-fidelity and physics-informed settings~\cite{lu2021learning}.
Together, these approaches reflect an increasing incorporation of structure and prior knowledge into model design and training, forming a continuum that balances data availability, physical insight, and computational goals.

Several surveys have examined the development and applications of KANs from different perspectives \cite{somvanshi2025survey, ji2024comprehensive, essahraui2025kolmogorov}. For instance, \cite{ji2024comprehensive} presents an analytical overview that focuses on empirical comparisons between KANs and conventional MLPs, offering a detailed discussion of architectural variants, interpretability approaches, and performance trade-offs. In addition, \cite{essahraui2025kolmogorov} provides a structured overview of existing KAN architectures and their applications, emphasizing interpretability as a central feature and addressing key computational and implementation challenges. In contrast, to comprehensively evaluate the capabilities of KANs in scientific machine learning, this review is structured around three major contexts: data-driven KANs, physics-informed KANs, and deep-operator KANs.
For each class, we examine the architectural design principles, training strategies, and recent innovations aimed at improving performance, interpretability, and scalability. We further present representative applications and provide comparative analyses against established neural solvers such as MLPs, PINNs, and DeepONets. 
Alongside reviewing prior studies, the paper also presents several comparative analyses carried out to clarify central characteristics of KAN schemes and how they might be applied in practice.
In the final section, we outline current challenges and propose future directions, focusing on high-dimensional scalability, mesh-independent modeling, theoretical guarantees, and advanced basis selection schemes. By organizing the review in this manner, we aim to present a unified perspective on KANs, offering both conceptual clarity and practical insight for their adoption in different learning frameworks.

\section{Kolmogorov-Arnold Representation Theorem}
\label{sec:kolmogorov_arnold}

Kolmogorov’s superposition theorem, originally introduced in 1957 by Kolmogorov~\cite{kolmogorov1957representations, kolmogorov1961representation} and later refined by Arnold in a series of papers~\cite{Arnold1958, arnold1959representation}, is a fundamental result in approximation theory that addresses whether arbitrary multivariate continuous functions can be decomposed into simpler univariate functions. Kolmogorov demonstrated that any continuous real-valued function defined on an \(n\)-dimensional domain can be expressed as a finite composition of continuous univariate functions and the addition operation. Arnold further refined this result, particularly for the case of three variables, shedding light on the structural essence of these representations. Over the years, this theorem has been improved by several researchers, including Lorentz~\cite{lorentz1966approximation,lorentz1996constructive}, Sprecher~\cite{sprecher1965structure,sprecher1972improvement}, and Friedman~\cite{fridman1967improvement}, among others. Hecht-Nielsen’s interpretation~\cite{hecht1987counterpropagation,hecht1987kolmogorov,sprecher1965structure} showed limitations due to the non-smoothness of the inner functions and the dependency of the outer functions on the specific target function. K\r{u}rkov{\'a}~\cite{kuurkova1991kolmogorov,kuurkova1992kolmogorov} addressed some of these issues by introducing approximations using sigmoidal functions, enabling practical implementations. Sprecher~\cite{sprecher1996numerical,sprecher1997numerical} developed a numerical algorithm for a constructive proof of the theorem, though challenges regarding the continuity and monotonicity of the inner functions persisted. K\"{o}ppen~\cite{koppen2002training} proposed modifications to overcome these shortcomings, and Braun and Griebel~\cite{braun2009constructive} advanced this work by rigorously proving the existence, continuity, and monotonicity of the modified inner functions, providing a complete constructive proof based on Sprecher’s framework with K\"{o}ppen’s improvements. This theorem, often referred to as the Kolmogorov-Arnold Representation Theorem (KART), remains a cornerstone in the study of function approximation and has found applications in neural networks, despite its inherent complexities. Let $f: [0,1]^n \to \mathbb{R}$ be a continuous function. Then KART states that there exist continuous univariate functions
$
\varphi_k : \mathbb{R} \to \mathbb{R}
~\text{and}~
\psi_{k,j} : \mathbb{R} \to \mathbb{R},
$
for indices $k = 0, 1, \ldots, 2n$, and $j = 1, 2, \ldots, n$, such that, 
\begin{equation}\label{KART}
f(x_1, x_2, \ldots, x_n)
=
\sum_{k=0}^{2n}
\varphi_k\Bigl(
\sum_{j=1}^{n} \psi_{k,j}(x_j)
\Bigr)
\quad
\text{for all }(x_1, \ldots, x_n) \in [0,1]^n.
\end{equation}

Intuitively, this means that one can build a continuous function of $n$ variables by summing together only univariate inner functions $\psi_{k,j}$ (one for each dimension) inside univariate outer functions $\varphi_k$. Thus, \emph{any} continuous map $f$ from the $n$-dimensional space to the real line can be realized by a relatively simple set of one-dimensional building blocks and repeated additions. This canonical structure corresponds to what is referred to in KANs as the original depth-2 representation, where the first layer consists of learnable univariate functions $\psi_{k,j}$ applied independently to each input variable, and the second layer consists of univariate functions $\varphi_k$ composed with the linear combinations of the outputs from the first layer. In other words, a single-layer KAN, as originally inspired by KART, implements a two-stage functional composition: first applying input-wise transformations, and then aggregating them through summation and another univariate nonlinearity. This architecture forms the mathematical and conceptual foundation of KANs. The theorem is often considered a cornerstone in the theory of \emph{universal function approximation}~\cite{cybenko1989approximation, hornik1989multilayer, hornik1991approximation, barron1993universal},  because it ensures that high-dimensional continuous functions can be decomposed into low-dimensional components. Moreover, \cite{ismayilova2024kolmogorov, ismailov2023three} show that in the Kolmogorov two–hidden–layer neural network, the outer activation can inherit the continuity, discontinuity, and boundedness of the target function, allowing it to represent all types of multivariate functions, whether continuous, discontinuous, bounded, or unbounded. This insight underpins much of modern approximation theory~\cite{sprecher1965structure, braun2009constructive, yarotsky2017error}, reinforcing the idea that high-dimensional phenomena may be tractably approximated using compositions of simpler (often one-dimensional) functions.

Although the theorem predates modern machine learning by several decades, it provides a theoretical basis for understanding how complex functions (such as classification or regression functions in deep learning) can, in principle, be approximated by systematic compositions of simple functions. In particular, the idea that combining ``depth'' (stacking functions or sequential composition) and ``width'' (adding functions in parallel) can help approximate complex high-dimensional mappings is similar to several theoretical arguments about the universality of neural networks~\cite{telgarsky2016benefits, lu2017expressive, schmidt2020nonparametric}. In this sense, Kolmogorov-Arnold representations show why neural architectures built from repeated layers of simple univariate transformations~\cite{hecht1987kolmogorov, alonso2024mathematics, kilani2024kolmogorov} may achieve remarkable expressiveness in practice. Although this result is fundamentally a theorem in classical analysis, its implications continue to inspire research in high-dimensional approximation, function representations, and computational architectures for learning complex relationships in data. Further refinements and variations of Kolmogorov-Arnold's original ideas continue to appear in the mathematics of approximation theory~\cite{kuurkova1992kolmogorov, braun2009constructive, igelnik2003kolmogorov}, functional analysis~\cite{liu2015kolmogorov, tikhomirov1963kolmogorov}, and theoretical computer science~\cite{kilani2024kolmogorov, borenstein2006kolmogorov, edmunds2006approximation}.

\section{Kolmogorov-Arnold Networks}
\label{sec:kolmogorov_arnold_networks}

The exploration of the Kolmogorov-Arnold representation theorem in the context of neural networks has evolved significantly over the years. Early investigations by Sprecher \& Draghici~\cite{sprecher2002space} and Köppen~\cite{koppen2002training} focused on its applicability in network training, but these efforts were largely confined to the original depth-2, width-\((2n + 1)\) formulation, as presented in Eq.~\eqref{KART}. Challenges such as the non-smoothness of the inner functions and the difficulty of efficient training limited their practical implementation. Subsequent works, including Montanelli \& Yang~\cite{montanelli2020error} and Lai \& Shen~\cite{lai2021kolmogorov}, have explored the theoretical implications and potential benefits of the theorem, though they did not fully exploit modern optimization techniques like backpropagation. 
More studies have revisited these limitations from complementary perspectives: Fakhoury et al.~\cite{fakhoury2022exsplinet} mitigate non-smoothness and improve empirical trainability by replacing the original inner mappings with smooth, locally supported B-spline bases (explicitly differentiable for degree greater than two); by contrast, He~\cite{he2023optimal} provides a constructive theoretical analysis of the optimal expressive power of ReLU deep neural networks within the Kolmogorov framework, strengthening approximation guarantees but not directly addressing smoothness or practical training-stability. Thus, while some work combines architectural choices and empirical evaluation to mitigate the classical obstacles, other contributions provide complementary theoretical foundations that justify and guide modern implementations. 
Building on KART and the aforementioned foundational work, Kolmogorov–Arnold Networks~\cite{liu2024kan, liu2024kan2, li2024kolmogorov, bozorgasl2405wav,ss2024chebyshev} have emerged as a modern computational architecture. KANs implement the classical theorem stating that any continuous function of \(n\) variables can be expressed as nested sums and compositions of univariate functions. By leveraging this decomposition, KANs construct neural network-like structures that balance expressiveness and computational efficiency, offering a promising alternative to traditional MLPs for scientific computing and high-dimensional approximation tasks~\cite{ta2025af, he2024mlp}.
In practice, KANs can be extended beyond the classical depth-2 structure by stacking multiple layers of univariate transformations, enabling deeper architectures that support hierarchical function approximation and improved representational capacity~\cite{schmidt2021kolmogorov, liu2024kan, yang2024kolmogorov, liu2024kan2}.

\subsection{Two-Layer KANs}

A two-layer Kolmogorov-Arnold Network~\cite{girosi1989representation, ismailov2014approximation, schmidt2021kolmogorov, poluektov2023construction}  as expressed in Eq.~\eqref{KART} aims to approximate a function, $F: \mathbb{R}^n \to \mathbb{R},$ by representing it using a depth-2, width-\((2n + 1)\) architecture analogous to the Kolmogorov-Arnold superposition, where each $\psi_{k,j}$ and $\varphi_k$ is a univariate, continuous (and in practice differentiable) function. 
In a \textit{two-layer} Kolmogorov-Arnold Network: (i) The \emph{first layer} consists of functions $\psi_{k,j}:\mathbb{R}\to \mathbb{R}$ for $j=1,\dots,n$ and $k=0,\dots,2n$, whose outputs are summed over $j$ (for each fixed $k$); (ii) The \emph{second layer} consists of univariate functions $\varphi_k:\mathbb{R}\to \mathbb{R}$ that take the sum from the first layer as input, and are subsequently summed across $k=0,\dots,2n$ to produce a final scalar output. Hence, the original $n$-dimensional input is decomposed into univariate transformations, and the network’s trainable parameters correspond to the internal representations of these $\psi_{k,j}$ and $\varphi_k$ functions.  

\subsection{Multi-Layer KANs}

A multi-layer KAN~\cite{liu2024kan} is defined as,
\begin{equation}\label{eq:mlayerKAN}
  \phi_k(\cdot)
  \;=\;
  \phi_{k}^{(L)}
  \Bigl(
  \ldots 
  \phi_{k}^{(2)}
  \bigl(
    \phi_{k}^{(1)}(\cdot)
  \bigr)
  \Bigr),
\end{equation}
where each $\phi_{k}^{(m)}$ remains a univariate function, but organized in a layered, nested manner. This architecture extends the classical depth-2 superposition model by allowing each univariate component to be represented as a composition of multiple univariate transformations.
Consequently, each ``outer'' function $\varphi_k$ can be modeled as a small network composed of univariate layers, and each ``inner'' function $\psi_{k,j}$ can similarly be enhanced with additional depth. Liu et al.~\cite{liu2024kan} introduced the concept of ``Kolmogorov-Arnold depth", proposing that certain functions, which cannot be smoothly approximated with only two layers, may benefit from deeper nested structures. Their study also addresses practical challenges and offers enhancements such as adaptive grid methods~\cite{rigas2024adaptive}, hybrid KAN-MLP architectures~\cite{li2024hkan, chechkin2025hybrid}, and multi-head activation grouping~\cite{yang2024activation, toscano2025kkans} for improved computational performance. 
A recent theoretical advance by Kratsios et al.~\cite{kratsios2025kolmogorov} further reinforces the mathematical underpinnings of multi-layer KANs. They rigorously proved that residual multi-layer KANs (Res-KANs) can optimally approximate functions belonging to spaces~\cite{chui1992general, devore1988interpolation}, even on complex or fractal domains.
Furthermore, they bounded the pseudodimension of Res-KANs to obtain dimension-independent sample complexity estimates, showing that Res-KANs can efficiently learn smooth functions even in high-dimensional spaces.
These results establish a solid theoretical foundation supporting the expressive power and generalization ability of deep KAN architectures.
These developments underscore the theoretical and empirical advantages of deeper KANs. By preserving the core principle of univariate composition, multi-layer KANs offer increased modeling flexibility and computational efficiency through deeper, more expressive representations~\cite{liu2015kolmogorov, polar2021deep}.

\begin{figure}
    \centering
    \includegraphics[width=.8\linewidth]{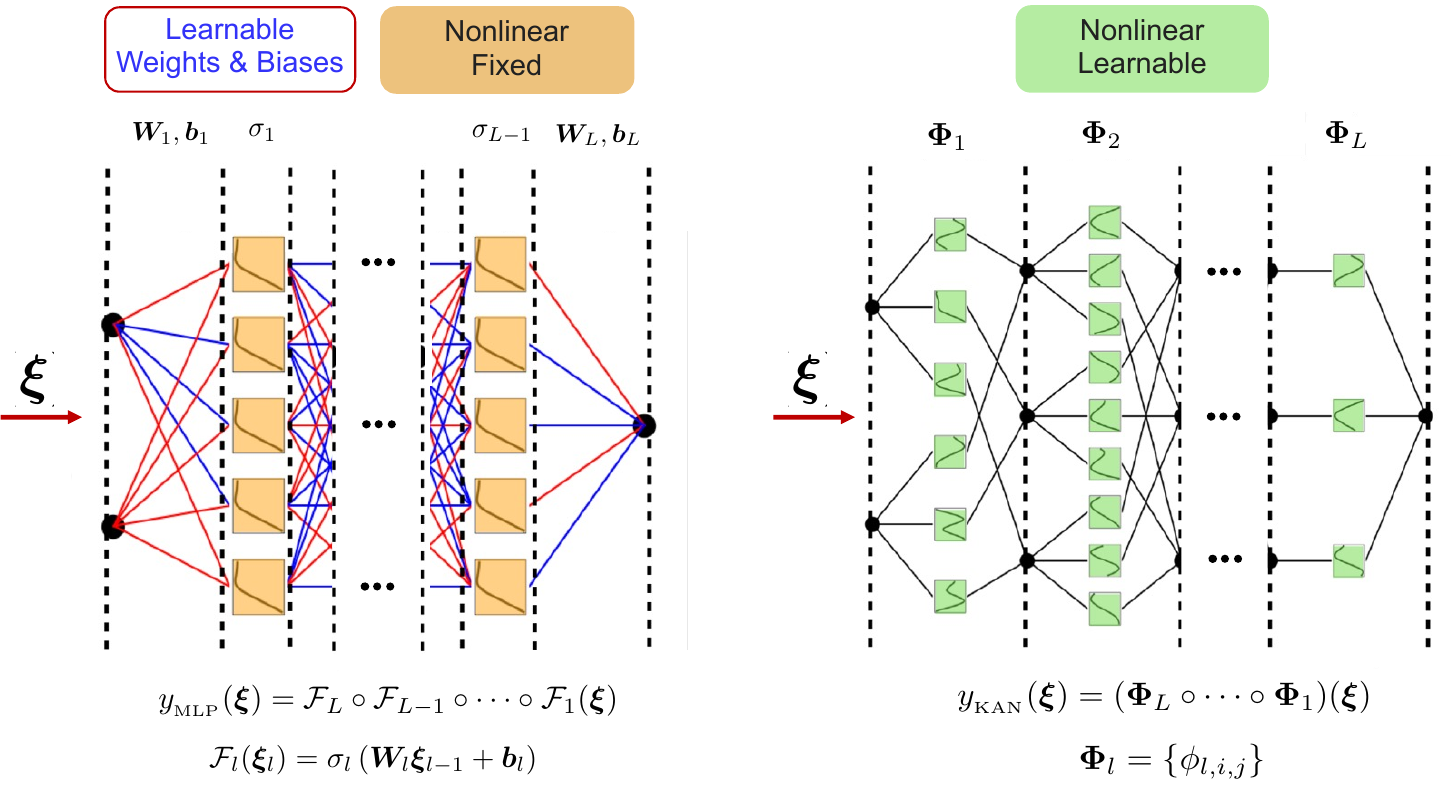}
    \caption{A conceptual illustration of how MLPs and KANs work. MLPs use a fully interconnected structure that mixes all input dimensions in each layer; 
    in contrast, KANs process each dimension with a univariate function, then combine the results via summation. This dimension-wise approach can offer a more direct path to representing high-dimensional functions. Here,  $\boldsymbol{\xi}$ represents the input vector, \(\boldsymbol{W}_{(l)}\) and \(\boldsymbol{b}_{(l)}\) refer to the weights and biases of the \(l\)-th layer, respectively, \(\sigma\) is a fixed activation function that can take different forms, e.g., ReLU, sigmoid, sin, tanh,~\cite{faroughi2023physics}, and  \(\boldsymbol{\Phi}_l\) is a KAN layer defined as a matrix of 1-dimensional functions.}
    \label{mlpvskan}
\end{figure}

\subsection{KANs vs.~MLPs}

A conceptual illustration of how MLPs and KANs work is shown in Fig.~\ref{mlpvskan}. To understand the key differences of KANs over traditional MLPs, we compare the two architectures across four dimensions: theoretical guarantees, architectural structure, representation power, and interpretability. (i) Both KANs and MLPs are universal approximators~\cite{cybenko1989approximation, hornik1989multilayer,  hornik1991approximation,  liu2024kan}, capable of representing any continuous function on a compact domain. However, the rate of approximation and the  compactness of representation may differ. KANs are grounded in the Kolmogorov–Arnold representation theorem~\cite{kolmogorov1957representations, arnold1959representation, sprecher2002space, koppen2002training}, which ensures exact representation of continuous multivariate functions via nested compositions and summations of univariate functions. 
Let each network be parameterized by a vector of trainable parameters \(\boldsymbol{\theta}\), with \(|\boldsymbol{\theta}|\) denoting the total number of parameters. Moreover, consider a target function $y \colon \mathbb{R}^n \to \mathbb{R}$, which is approximated by the networks with outputs \( y_{_\text{KAN}} \) and \( y_{_\text{MLP}} \).
Then, under suitable smoothness assumptions~\cite{barron2002universal, devore1998nonlinear}, we may write,
\begin{equation}
    \bigl\| y - y_{_\text{KAN}} \bigr\|_\infty \leq C_{_\text{KAN}}(y,n) \cdot |\boldsymbol{\theta}|^{-\alpha}, \quad
    \bigl\| y - y_{_\text{MLP}} \bigr\|_\infty \leq C_{_\text{MLP}}(y,n) \cdot |\boldsymbol{\theta}|^{-\beta},
\end{equation}
for constants \( \alpha > \beta > 0 \). In other words, the exponent $\alpha$ that governs the approximation rate in a KAN can exceed the MLP's exponent $\beta$, leading to more efficient parameter usage, especially in problems where each dimension has substantial structure that can be exploited by dimension-wise univariate transformations~\cite{sharma2020neural, michaud2023precision, poggio2020theoretical}. (ii) MLPs rely on layers of linear transformations followed by fixed pointwise nonlinearities (e.g., ReLU, sigmoid), mixing all input dimensions at each layer~\cite{hornik1991approximation, karlik2011performance}. In contrast, KANs decompose multivariate functions into combinations of learnable univariate transformations along edges (Fig.~\ref{mlpvskan})~\cite{liu2024kan, liu2024kan2}. This leads to a fundamentally different parameterization strategy that emphasizes function composition over matrix multiplication. In KANs, each edge applies a univariate transformation (e.g., radial basis functions or Chebyshev polynomials, see Section~\ref{subsec:univariate_basis_functions})~\cite{rivlin2020chebyshev, li2024kolmogorov, bozorgasl2405wav}, making the architecture more aligned with the structure of the Kolmogorov–Arnold theory. (iii) The representational strength of MLPs heavily depends on their width and depth, with no explicit structural guidance for decomposing input dimensions. KANs, by design, embed a strong inductive bias based on univariate superpositions. This bias can be especially beneficial in high-dimensional settings where individual input dimensions follow structured, learnable transformations~\cite{liu2024kan2}. (vi) KANs provide clearer interpretability by isolating the influence of each input dimension through dedicated univariate modules. This facilitates more transparent analyses of input–output relationships, a feature less accessible in traditional MLPs, where variable mixing via weight matrices obscures individual contributions~\cite{kilani2024kolmogorov, cruz2025state}.

\subsection{Choice of Univariate Basis Functions}\label{subsec:univariate_basis_functions}

A central design consideration in constructing KANs is the choice of univariate basis functions used in Eq.~\eqref{eq:mlayerKAN}. A variety of parametric forms can be used, but many studies have focused on established approximation families, such as polynomials, wavelets, or radial kernels, for their well-known theoretical and practical benefits~\cite{trefethen2019approximation, bozorgasl2405wav, li2024kolmogorov}. Traditional KAN architectures often utilize B-spline functions~\cite{de1978practical, bohra2020learning, vaca2024kolmogorov, samadi2024smooth}, though recent investigations have proposed alternative bases to improve flexibility and efficiency. For example, the Chebyshev KANs~\cite{ss2024chebyshev} leveraged Chebyshev polynomials to handle functions with high oscillations, yielding better performance on tasks such as image classification and complex function approximation. FastKAN~\cite{li2024kolmogorov} replaced B-splines with Gaussian radial basis functions for significant speedups while preserving accuracy. Similarly, Wav-KAN~\cite{bozorgasl2405wav} employed wavelet expansions to capture multi-resolution patterns, leading to better interpretability for certain datasets.

Jacobi polynomials, including Legendre, Chebyshev, and others, are a well-studied class for polynomial expansions. Within this class, B-splines and Chebyshev polynomials are frequently favored in KAN implementations~\cite{liu2024kan, vaca2024kolmogorov, samadi2024smooth}. KANs originally employ basis functions in the form of splines,
\begin{equation}
\phi(\xi) = w_b b(\xi) + w_s \text{spline}(\xi), \quad b(\xi) = \frac{\xi}{1 + e^{-\xi}}, \quad \text{spline}(\xi) = \sum_i c_i B_i(\xi),
\end{equation}
where \(B_i(\xi)\) are spline basis functions defined by the grid size \(g\) and polynomial degree \(k\) tending to provide robust local control over the function shape, and $c_i$, $w_b$, and $w_s$ are coefficients learned during training, thus $\boldsymbol{\theta} = \{c_i,\;w_b,\;w_s\}$. B-splines are often robust for modeling structured data, since local changes in the knot placement or coefficients can affect only a limited portion of the domain. However, increasing \(g\) to fit complex data can greatly increase the number of trainable parameters $|\boldsymbol{\theta}|_\text{KAN} \sim O(N_l N_n^2 (k + g))$,
where \(N_l\) is the number of layers, and \(N_n\) is the number of neurons per layer. A potential solution to the computational challenge involves utilizing Chebyshev polynomials, which serve as a foundational tool in approximation theory and numerical analysis, by employing them as an orthogonal basis~\cite{rivlin2020chebyshev, schmidt2021kolmogorov}. Chebyshev polynomials are defined as,
\begin{equation}\label{eq:cheb_series}
   \phi(\xi) = \sum_{i=0} c_i\, T_{i}(\xi),
\end{equation}
where $T_{i}(\xi)$ denotes the $i$th Chebyshev polynomial of degree $k$, and the learnable parameter set is $\boldsymbol{\theta} = \{c_i \}$ reducing the number of trainable parameters to 
$|\boldsymbol{\theta}|_\text{cKAN} \sim O(N_l N_n^2 k)$. Such expansions often yield rapid convergence for smooth functions. Chebyshev polynomials can be evaluated in two main ways. A recursive formulation is more stable at larger polynomial degrees but may be slower,
\begin{equation}\label{eq:cheb_recursive}
   T_{0}(\xi) = 1,\quad T_{1}(\xi) = \xi,\quad 
   T_{i}(\xi) = 2\,\xi\,T_{i-1}(\xi)\;-\;T_{i-2}(\xi),
\end{equation}
while, by contrast, a trigonometric formulation can be faster but may risk numerical instability if $\xi$ approaches the bounds of $[-1,1]$ during training,
\begin{equation}\label{eq:cheb_trig}
   T_{i}(\xi) = \cos\bigl(i\,\arccos(\xi)\bigr),
\end{equation}
which can trigger undefined values and NaN losses~\cite{boyd2001chebyshev, mason2002chebyshev, trefethen2019approximation}. Recent works in KANs have proposed careful input normalization or additional regularization to mitigate this issue~\cite{mostajeran2025scaled}.  To improve stability while training, Chebyshev polynomials restrict their inputs in the range \([-1, 1]\). That is accomplished by applying the  \(\tanh\)  after each layer~\cite{guo2024physics, hu2024tackling, ss2024chebyshev}, changing the corresponding forward pass, described in Fig.~\ref{mlpvskan} as,
\begin{equation}  
y_{_\text{cKAN}}(\boldsymbol{\xi}) = (\boldsymbol{\Phi}_L \circ \tanh \circ \cdots \circ \tanh \circ ~ \boldsymbol{\Phi}_1 \circ \tanh)(\boldsymbol{\xi}), 
\end{equation}

Wavelet expansions offer a powerful and flexible framework for capturing localized or piecewise-regular structures in data~\cite{bozorgasl2405wav}. A univariate function can be represented as,  
\begin{equation}\label{eq:wavelet_expansion}
   \phi(\xi) = \sum_{\ell}\sum_{r} d_{\ell,r}\,\zeta_{\ell,r}(\xi),
\end{equation}  
where \(d_{\ell,r}\) denotes the wavelet coefficient, representing the contribution of the basis function at scale \(\ell\) and location \(r\), and \(\zeta_{\ell,r}(\xi)\) is a scaled and shifted version of the mother wavelet. Specifically, the wavelet basis is defined as \(\zeta_{\ell,r}(\xi) = \phi\left( (\xi - c_{\ell,r})/ \sigma_{\ell,r}\right)\), where \(c_{\ell,r}\) controls the translation and \(\sigma_{\ell,r}\) governs the dilation (or spread) of the wavelet function~\cite{bozorgasl2405wav, mostajeran2023novel}. In the context of Wav-KAN~\cite{bozorgasl2405wav}, this parameterization enables adaptive learning of localized features in each input dimension. The complete set of trainable parameters is given by  
\(
\boldsymbol{\theta} = \left\{c_{\ell,r},\ \sigma_{\ell,r},\ d_{\ell,r}\right\},
\)
and the total number of parameters in Wav-KAN scales as \(|\boldsymbol{\theta}|_\text{Wav-KAN} \sim O(3 N_l N_n^2)\), where \(N_l\) is the number of layers and \(N_n\) is the number of wavelet nodes per layer~\cite{bozorgasl2405wav}. Common choices for the mother wavelet \(\phi\) include the Morlet~\cite{bhat2024novel}, Mexican-Hat~\cite{ilyas2024design}, Haar~\cite{ramalakshmi2024identity}, and Daubechies wavelets~\cite{ali2024short}, each offering distinct localization properties in time and frequency domains~\cite{mallat1999wavelet, daubechies1992ten, torrence1998practical}. Recent developments have introduced novel mother wavelet designs optimized for specific applications. For instance, Levie et al.~\cite{levie2021wavelet} proposed a variational approach to design mother wavelets with optimally localized ambiguity functions, enhancing sparsity and reducing redundancy in continuous wavelet transforms. Additionally, Gauthier et al.~\cite{gauthier2022parametric} introduced Parametric Scattering Networks, wherein wavelet parameters such as scales and orientations are learned from data, leading to improved performance in small-sample classification tasks. The wavelet-based formulation allows KANs to model fine-grained, local variations across dimensions, which is essential for learning from data with abrupt transitions, discontinuities, or highly non-uniform patterns~\cite{bruna2013invariant, bozorgasl2405wav}.

Radial basis functions (RBFs) provide yet another avenue for univariate approximation~\cite{li2024kolmogorov}. A typical representation of a univariate function using RBFs is,  
\begin{equation}
    \phi(\xi) = \sum_{r=1}^{N_n} w_r\,\kappa\left(\frac{\xi - c_r}{\sigma_r}\right),
\end{equation}  
where \(w_r\) are the trainable weights, \(c_r\) are the centers, \(\sigma_r\) are the shape (or scale) parameters, and \(\kappa(\cdot)\) is a chosen radial kernel, for instance, the Gaussian kernel \(\kappa(\xi) = \exp(-\gamma\,\xi^2)\), where \(\gamma = 1/\sigma^2\). The set of all trainable parameters is  
\(
\boldsymbol{\theta} = \left\{w_r,\, c_r,\, \sigma_r \right\}, 
\)
and the total number of parameters scales as \(|\boldsymbol{\theta}|_\text{RBF} \sim O(3N_l N_n^2)\). RBF expansions are known for smooth interpolation properties and ease of application to certain regression tasks~\cite{mostajeran2023radial, liu2025solving, muthusamy2025economic, mirzaei2024parallel}. Several types of RBFs have been developed to address diverse approximation tasks. Gaussian RBFs are ideal for smooth function modeling~\cite{bouzidi2025gaussian, buhmann2003radial, stenkin2024mathematical}, multiquadric functions perform well on scattered data~\cite{liu2025efficient}, inverse multiquadrics offer smooth interpolation for PDEs~\cite{ku2024deep}, and Wendland functions enable sparse, meshfree approximations due to their compact support~\cite{wendland2004scattered}.

Two other alternative approaches for implementing univariate modules in KANs are: (i) using shallow fully connected neural networks, and (ii) combining multiple basis functions. In the first approach, each univariate transformation is represented by a small multilayer perceptron. This retains the univariate nature of the architecture while leveraging GPU-accelerated training and flexible function approximation. Recent studies have shown that this method can perform competitively on high-dimensional tasks with proper tuning~\cite{lai2021kolmogorov, zhang2021multiscale}. The second approach, exemplified by FC-KAN~\cite{ta2024fc}, enhances representation power by combining fundamental functions such as B-splines, wavelets, and radial basis functions. Instead of relying on a single function type, FC-KAN employs strategies like element-wise operations, polynomial expansions, and concatenation to enrich the expressiveness of univariate mappings.
A related line of work by Fakhoury \& Speleers~\cite{fakhoury2025expressivity} analyzed the ExSpliNet model, a KAN-inspired architecture based on multivariate B-spline representations, and provided constructive approximation results demonstrating its strong expressivity and ability to mitigate the curse of dimensionality.

Each family of basis functions comes with distinct advantages and challenges. In Table~\ref{Tab:Orders}, we summarize the key characteristics of commonly used basis functions in KANs, including their parameterization, computational complexity, and typical application scenarios. Polynomials are often fast for smooth functions but can require careful scaling or recursive evaluation. Wavelets work well for localized behaviors but demand extra decisions about wavelet types and decomposition levels. RBFs excel in smooth interpolation but require a suitable kernel choice. Shallow neural networks benefit from mature deep-learning knowledge stacks, though controlling their capacity may require significant hyperparameter searches. In practice, the choice of basis functions is guided by the problem's structure, computational efficiency, and interpretability requirements. Ongoing research continues to refine these choices to enhance KAN performance across diverse applications~\cite{ss2024chebyshev,li2024kolmogorov,bozorgasl2405wav}.

\begin{table}[h!]
\centering
\caption {Comparison of basis functions used in KANs with \(N_l\) layers, \(N_n\) neurons per layer, grid size \(g\), and polynomial degree \(k\). The table outlines their parameterization, computational order, and suitability for function approximation.}

\label{Tab:Orders}
\footnotesize
\begin{tabular}{p{1.9cm} p{2cm} p{2.2cm} p{6.5cm} p{2.0cm}}
\toprule  
  {Basis Function} &  {Parameters \(\boldsymbol{\theta}\)}  &  {Order}  & {Applications} & {Ref}\\ 
\midrule
Splines
&\(\{w_s, w_b, c_i\}\)
&\(O(N_l N_n^2 (k + g))\)
&
Robust for structured data with local control; high parameter count for complex functions&
\cite{bohra2020learning, vaca2024kolmogorov, samadi2024smooth,li2024kolmogorov}\\
\addlinespace
\addlinespace
Chebyshev
&\(\{c_i\}\)
&\(O(N_l N_n^2 k)\)
& 
Efficient for smooth functions; fewer parameters and fast convergence; requires input normalization
&
\cite{boyd2001chebyshev, mason2002chebyshev, mostajeran2025scaled, guo2024physics, hu2024tackling, ss2024chebyshev}\\
\addlinespace
\addlinespace
Wavelets
&\(\{d_{\ell,r}, c_{\ell,r}, \sigma_{\ell,r}\}\)
&\(O(3 N_l N_n^2)\)
& Suitable for functions with sharp, high-amplitude oscillations;
 ideal for multi-resolution data
&
\cite{bozorgasl2405wav, mostajeran2023novel, bhat2024novel, ilyas2024design, ramalakshmi2024identity, ali2024short, mallat1999wavelet, daubechies1992ten, torrence1998practical,levie2021wavelet, gauthier2022parametric}\\
\addlinespace
\addlinespace
RBFs
&\(\{w_{r}, c_{r}, \sigma_{r}\}\)
&\(O(3 N_l N_n^2)\)
& 
Effective for smooth interpolation; compact, adaptive, and easy to tune with kernel variations
&
\cite{mostajeran2023radial, liu2025solving, bouzidi2025gaussian, buhmann2003radial, liu2025efficient, ku2024deep, wendland2004scattered, muthusamy2025economic, mirzaei2024parallel, stenkin2024mathematical}\\ 
\bottomrule
\end{tabular}
\end{table}

\section{Data-driven KANs}

This section presents a comprehensive overview of data-driven Kolmogorov–Arnold Networks, organized into three main parts. We first describe the core architecture and training strategies used to construct KAN models from input–output datasets, with emphasis on optimization methods and recent structural variations. The second part focuses on applications, highlighting how KANs have been employed across regression, classification, and segmentation tasks in both structured and unstructured domains. Finally, we provide a comparative analysis between KAN and MLP models along three dimensions: predictive accuracy, training convergence, and spectral behavior. Together, these components aim to provide a practical and task-oriented understanding of KANs in data-driven learning scenarios.

\begin{figure}[!ht]
    \centering
    \includegraphics[width=1\linewidth]{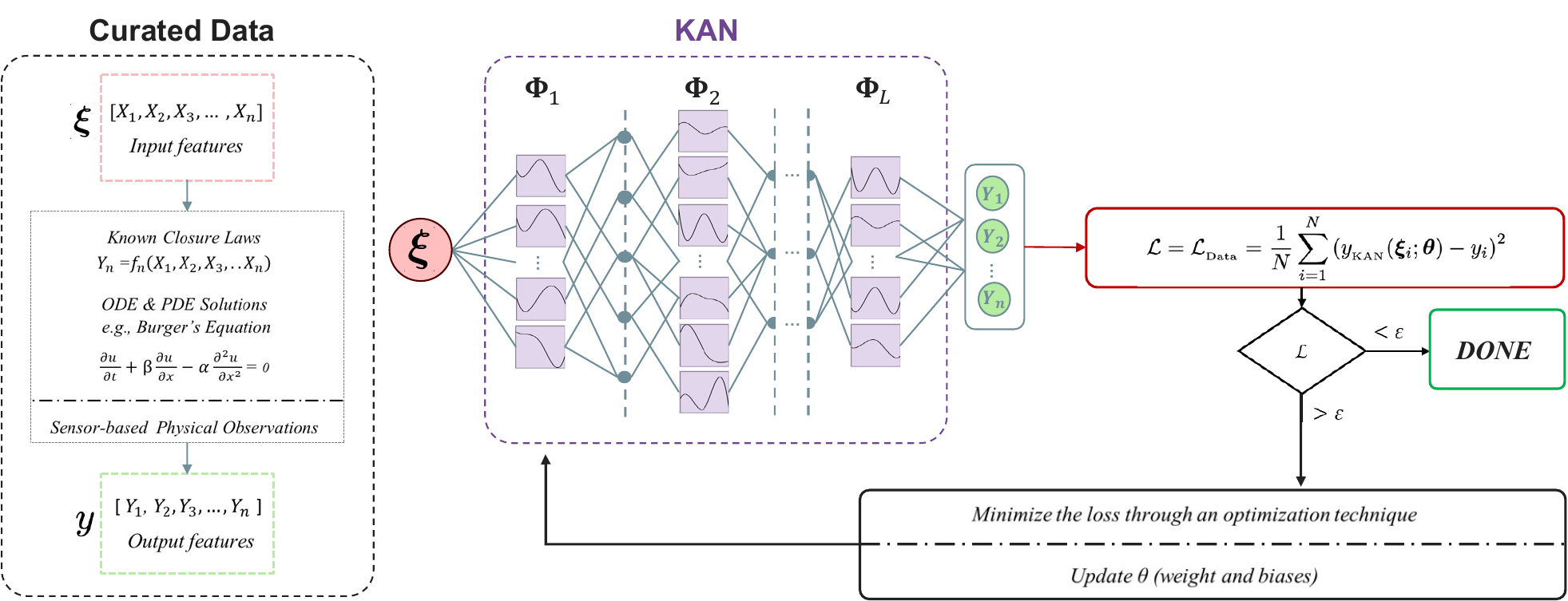}
    \caption{
    Schematic architecture of a data-driven KAN model. The network receives input features and directly approximates the target output by learning a parameterized mapping $y_{_\text{KAN}}(\boldsymbol{\xi}; \boldsymbol{\theta})$ from data, without using physical governing equations. It employs a modified feed-forward architecture with learnable activation functions. The parameters $\boldsymbol{\theta}$ are optimized by minimizing a supervised loss, typically the mean squared error, using gradient-based methods, enabling the network to capture the underlying structure in the observed data.
    }
   \label{fig:DataDriven}
\end{figure}

\subsection{Architecture and Training Strategies}\label{Sec.DataDArchi}

In the data-driven setting, the KAN-based model is trained purely from input-output data pairs without incorporating any knowledge of the governing physical laws. As illustrated in Fig.~\ref{fig:DataDriven}, the training process aims to approximate an unknown functional relationship between input features and target outputs using a KAN. Let the training dataset consist of $N$ samples, denoted by the input-output pairs $\{(\boldsymbol{\xi}_i, y_i)\}_{i=1}^N$, where each input vector $\boldsymbol{\xi}_i \in \mathbb{R}^d$ lies in a $d$-dimensional domain, and $y_i$ is the corresponding observed output. The network prediction at the $i$-th input is represented by $y_{_\text{KAN}}(\boldsymbol{\xi}_i; \boldsymbol{\theta})$, where $\boldsymbol{\theta}$ is the set of trainable parameters of the network. The goal of training is to learn a parameterized function that accurately maps inputs $\boldsymbol{\xi}_i$ to outputs $y_i$, thereby capturing the underlying structure of the data. To guide the learning process, a supervised loss function is employed to quantify the mismatch between the network’s predictions and the observed data. A commonly used choice is the mean squared error (MSE), which measures the average squared difference between predicted and true values. This loss function is defined as,
\begin{equation}\label{Eq.LossData}
\mathcal{L} = \mathcal{L}_{_\text{Data}} = \frac{1}{N} \sum_{i=1}^{N} \left( y_{_\text{KAN}}(\boldsymbol{\xi}_i; \boldsymbol{\theta}) - y_i \right)^2,
\end{equation}
where $N$ is the total number of training samples, $y_i$ is the observed output corresponding to the input $\boldsymbol{\xi}_i$, and $y_{_\text{KAN}}(\boldsymbol{\xi}_i; \boldsymbol{\theta})$ denotes the network’s prediction parameterized by $\boldsymbol{\theta}$. This loss formulation encourages the model to produce predictions that closely match the empirical data. The training objective is to determine the optimal parameters $\boldsymbol{\theta}^*$ that minimize the data-driven loss function,
\begin{equation}
    \boldsymbol{\theta}^* = \arg\min_{\boldsymbol{\theta}} \mathcal{L},
\end{equation}
thereby enabling the network to learn the underlying mapping from inputs to outputs based solely on empirical observations. This framework is widely applicable to general regression problems, where the goal is to approximate an unknown function from observed data samples.

To achieve this objective, the network parameters are iteratively updated using backpropagation~\cite{wright2022deep, dampfhoffer2023backpropagation, ji2024comprehensive}; a procedure that computes the gradients of the loss function with respect to all trainable parameters. These gradients provide the necessary direction for optimization algorithms to adjust the network parameters toward minimizing the loss. Commonly used first-order optimization methods include stochastic gradient descent (SGD)~\cite{newton2018stochastic, tian2023recent} and the Adam optimizer~\cite{kingma2014adam}, both of which rely on gradient information to efficiently update the model during training. In contrast, second-order or quasi-Newton methods, such as the Broyden–Fletcher–Goldfarb–Shanno (BFGS) algorithm~\cite{nawi2006improved} and its limited-memory variant (L-BFGS)~\cite{saputro2017limited}, utilize curvature information to accelerate convergence. More recent advancements, such as the Self-Scaled Broyden (SSB) algorithms~\cite{al2014broyden, kiyani2025optimizer}, further enhance these quasi-Newton approaches by employing self-scaling strategies. These quasi-Newton methods iteratively construct approximations to the inverse Hessian matrix, allowing for more informed parameter updates by incorporating curvature information. This typically results in faster and more stable convergence compared to first-order methods, especially in high-dimensional or ill-conditioned optimization problems~\cite{wright2006numerical, rafati2020quasi}. To address the optimization challenges in data-driven KANs, it is worth considering hybrid strategies that combine the strengths of different optimizers. For instance, combining adaptive first-order methods like Adam~\cite{kingma2014adam}, which are effective for fast initial convergence, with second-order methods such as L-BFGS~\cite{liu1989limited, saputro2017limited}, which offer precise updates near local optima, may significantly improve training efficiency. This approach can be particularly beneficial for KANs, where the optimization involves a large number of parameters and the joint tuning of both the shape and location of spline-based transformations~\cite{de1978practical, bohra2020learning, fakhoury2022exsplinet}. In this context, Adam can help navigate the early stages of training with noisy gradients, while L-BFGS can refine the solution more effectively in the later stages.

Recent advances in the design of data-driven KAN architectures have centered on structural innovations aimed at improving expressivity, interpretability, and efficiency. One direction involves hybrid basis function integration, as demonstrated in BSRBF-KAN, which combines B-splines with radial basis functions to enhance local approximation accuracy~\cite{ta2024bsrbf}. This hybrid formulation is implemented through a combined transformation layer applied to normalized inputs. The authors highlighted that, while both basis types contribute to performance, ablation studies suggest that B-splines play a more dominant role in preserving generalization. Additionally, Ta~\cite{ta2024bsrbf} provided insights into model stability and convergence behavior, indicating BSRBF-KAN’s potential as a flexible template for further exploration of basis function combinations in KANs. Other works, such as FC-KAN~\cite{ta2024fc}, propose function composition strategies by combining multiple elementary functions, such as wavelets, B-splines, and Gaussians, using element-wise operations or polynomial mappings to boost learning capacity. These methods show improved accuracy at the cost of increased training time, highlighting a trade-off between representational richness and computational load. Another important line of work focuses on initialization strategies for spline-based KANs. Rigas et al.~\cite{rigas2025initialization} systematically evaluated both theory-driven schemes, inspired by LeCun~\cite{lecun2002efficient} and Glorot~\cite{glorot2010understanding} initializations, and an empirical family of power-law strategies, showing that initialization plays a decisive role in convergence speed and predictive accuracy across regression and PDE benchmarks. In a different direction, Polar \& Poluektov~\cite{polar2021probabilistic} proposed a probabilistic extension of the KAN by combining it with the divisive data re-sorting (DDR) method. While conventional regression models focus on expected values or confidence intervals, their method enables KANs to represent full probability distributions of outputs under aleatoric uncertainty, efficiently capturing input-dependent and multimodal behaviors with relatively low computational cost. Moreover, Mostajeran \& Faroughi~\cite{mostajeran2024epi} proposed physics-guided parallel and serial KAN architectures with Chebyshev basis functions to evaluate their effectiveness in accurately modeling complex datasets that represent elasto-plastic material behavior. These structural adaptations reflect the growing interest in tailoring KAN frameworks to specific task demands, offering pathways toward more flexible and application-aware neural architectures.

Alongside structural enrichment, another key research direction involves simplifying data-driven KAN architectures to reduce computational overhead without compromising performance. The HKAN model~\cite{dudek2025hkan} introduces a novel hierarchical design that replaces iterative backpropagation with a non-iterative, layer-wise least squares fitting procedure, eliminating gradient computation. This architecture adopts a stacking mechanism in which each layer incrementally refines the previous output, enhancing model stability and interpretability while remaining computationally efficient. Meanwhile, the PRKAN architecture~\cite{ta2025prkan} targets parameter reduction by combining convolutional layers, pooling, attention mechanisms, and dimension summation strategies to drastically lower the number of learnable parameters. Despite reduced complexity, PRKAN maintains competitive performance in image classification, leveraging normalization and hybrid basis functions like GRBFs. In parallel, the Multifidelity KAN (MFKAN) framework~\cite{howard2025multifidelity} provides a method for combining low- and high-fidelity datasets, thereby reducing the need for costly high-fidelity data. By using both linear and nonlinear relationships between datasets, MFKANs become more reliable when data is limited. This reflects a shift toward lightweight, scalable KAN architectures, balancing structural simplicity with functional robustness.
Warin~\cite{warin2024p1} also  introduced the P1-KAN architecture, which employs compact piecewise-linear basis functions with bounded support. This formulation simplifies the representation while maintaining high approximation accuracy, particularly for irregular or non-smooth functions in high-dimensional spaces.

\subsection{Applications}

The adoption of data-driven KANs has been on the rise for a diverse array of tasks, including forecasting, regression, classification, and segmentation applications in both structured tabular domains and high-dimensional vision tasks. These models leverage the flexible function approximation capabilities to learn complex mappings directly from high-dimensional data. A summary of recent application-oriented advances in data-driven KANs is provided in Table~\ref{tab:KAN_Applications}.

\begin{table}[h]
\centering
\small
\renewcommand{\arraystretch}{1.05}
\caption{A non-exhaustive list of recent studies on application-oriented advances in data-driven KANs, highlighting their architectures, basis functions, and targeted real-world tasks across scientific and engineering domains.}
\label{tab:KAN_Applications}
\footnotesize
\begin{tabularx}{\textwidth}{>{\hsize=0.2\hsize}X >{\hsize=0.75\hsize}X >{\hsize=0.15\hsize}r}
\toprule  
\textbf{Method} & \textbf{Application Focus} & \textbf{Ref.} \\ 
\midrule
EPi-cKAN & 
Developed  physics-guided serial and parallel architectures to model nonlinear elasto-plastic behavior of granular materials;
Used Chebyshev polynomials as basis functions;
Applied to predict stress–strain responses.
& \cite{mostajeran2024epi} \\
\addlinespace
\addlinespace
Time-Series KAN &
Developed a KAN framework for time series forecasting;
Used B-spline basis functions (degree 3 and grid size 5);
Applied to forecast real-world satellite traffic using normalized hourly data from the 5G-STARDUST project.
& \cite{vaca2024kolmogorov} \\
\addlinespace
\addlinespace
Geotechnical KAN & 
Developed a KAN model to predict settlement and residual strength after liquefaction;
Used B-spline basis functions;
Applied to field and lab data from case studies, including risk assessment in Patna and railway embankment analysis.
& \cite{karakacs2024novel} \\
\addlinespace
\addlinespace
COEFF-KANs & Developed a two-stage modeling strategy to predict Coulombic Efficiency (CE) of lithium-metal battery electrolytes;
Applied to a real-world Li/Cu half-cell dataset to predict log-transformed CE values.
& \cite{li2024coeff} \\
\addlinespace
\addlinespace
SCKansformer &
Developed a fine-grained classification model to improve diagnostic accuracy in hematological malignancies by addressing limitations of existing microimage analysis methods; Applied to over 10,000 cell images across three datasets (BMCD-FGCD, PBC, ALL-IDB).
& \cite{chen2024sckansformer} \\
\addlinespace
\addlinespace
U-KAN & 
Developed a KAN-enhanced convolutional architecture to improve nonlinear representation and interpretability in image segmentation and generative tasks; Used tokenized KAN layers; Applied to diverse imaging datasets (BUSI, GlaS, and others).
& \cite{li2025u} \\
\addlinespace
\addlinespace
KACQ-DCNN &
Developed a classical–quantum hybrid neural architecture;
Used B-spline basis functions within a dual-channel design to model nonlinear activation patterns and capture complex feature relationships;
Applied to structured data for binary classification. 
& \cite{jahin2024kacq} \\
\addlinespace
\addlinespace
KAN-IDIR &
Developed an implicit neural representation framework for deformable image registration with KANs; Utilized adapted Chebyshev KANs ; Applied to three medical imaging benchmarks: DIR-Lab lung CT, OASIS-1 brain MRI, ACDC cardiac MRI.
& \cite{drozdov2025learning} \\
\bottomrule
\end{tabularx}
\end{table}

\begin{figure} [h]
    \centering
    \includegraphics[width=1\linewidth]{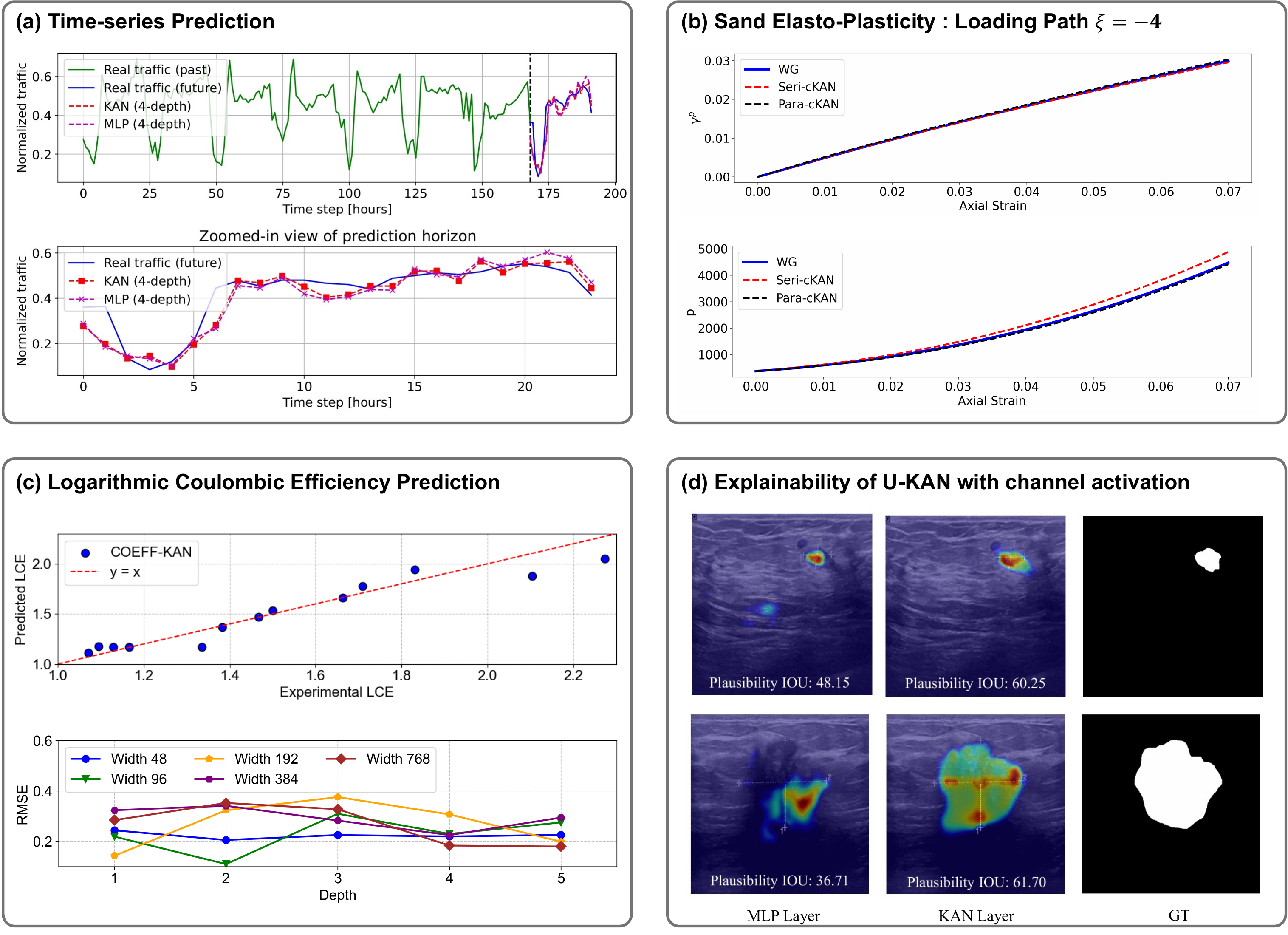}
    \caption{\textbf{(a)} KAN with learnable B-spline activations for time series forecasting. The forecast is based on real GEO satellite traffic data (hourly resolution, 168h context / 24h prediction) provided by the 5G-STARDUST project, highlighting KAN's applicability to AI-driven satellite resource management. This panel is adopted from~\cite{vaca2024kolmogorov}. \textbf{(b)} Prediction of plastic shear strain $\gamma^p$ and mean effective stress $p$ using parallel and serial KAN architectures under an undrained loading path with $\xi = -4$, $p_{\text{in}} = 375$~kPa, and $e_{\text{in}} = 0.64$. The results are reproduced based on the method presented in~\cite{mostajeran2024epi}, along with the ground truth obtained from numerical integration. \textbf{(c)} Application of COEFF-KAN to electrolyte property prediction. \textbf{Top:} Parity plot comparing predicted and ground-truth values of logarithmic Coulombic Efficiency (LCE) across test samples, illustrating strong agreement and model generalization. \textbf{Bottom:} RMSE performance of KAN versus MLP across varying network depths and widths. This panel is reproduced from~\cite{li2024coeff}. \textbf{(d)} Channel-wise activation maps for explainability evaluation in tumor segmentation. Integrating KAN layers enhances spatial alignment between model attention and ground truth, improving plausibility in localization tasks. This panel is adopted from~\cite{li2025u}.}
    \label{fig:DataDrivenKANs}
\end{figure}

In the context of regression tasks, Vaca-Rubio et al.~\cite{vaca2024kolmogorov} proposed a data-driven KAN architecture as a parameter-efficient and interpretable alternative to traditional MLPs for time series forecasting, with a specific focus on satellite traffic prediction. Inspired by  KART, their model replaces conventional weight-activation pairs with spline-parametrized univariate transformations, enabling greater modeling flexibility. As shown in Fig.~\ref{fig:DataDrivenKANs}(a), the KAN model closely follows the true traffic trajectory, particularly during rapid fluctuations. This demonstrates KAN’s ability to capture high-frequency temporal features and long-range dependencies. Further analysis demonstrates the impact of KAN-specific hyperparameters, such as node count and grid size, in balancing model accuracy and computational cost. These findings establish KANs as a promising, resource-efficient approach for complex time series forecasting tasks. Beyond temporal forecasting, KANs have also been successfully applied to structured regression problems. For example,  Karakaş~\cite{karakacs2024novel} employed a data-driven KAN model to predict liquefaction-induced settlements using a dataset derived from the Pohang earthquake region in South Korea, previously compiled by Park et al.~\cite{park2020simple}. The input features included depth, unit weight, cyclic stress ratio, and corrected SPT blow count, with settlement magnitude as the target variable. The study demonstrated that KANs effectively capture the nonlinear dependencies between geotechnical parameters and settlement outcomes, achieving high predictive accuracy across standard evaluation metrics such as $R^2$, Mean Absolute Error (MAE), and Root Mean Squared Error (RMSE). Notably, the model identified cyclic stress ratio and SPT blow count as the most influential predictors, aligning well with domain-specific understanding. These results emphasize the potential of KANs as reliable and interpretable tools for modeling complex phenomena in data-limited environments. Further extending the applicability of KANs to granular material modeling, Mostajeran \& Faroughi~\cite{mostajeran2024epi} developed Chebyshev-based KAN variants to predict nonlinear elasto-plastic behavior under complex loading. Using parallel and serial KAN architectures, the models achieved accurate predictions of plastic shear strain $\gamma^p$ and mean effective stress $p$, even under unseen undrained loading paths such as $\xi = -4$, as shown in Fig.~\ref{fig:DataDrivenKANs}(b). These results demonstrate the effectiveness of KANs in capturing stress–strain responses in elasto-plastic materials with compact and interpretable architectures. Further broadening the landscape of KAN applications, Li et al.~\cite{li2024coeff} introduced \textit{COEFF-KAN}, a two-stage framework leveraging Kolmogorov--Arnold Networks to predict Coulombic Efficiency (CE) in lithium battery electrolytes. In this structured regression task, the model inputs consisted of SMILES-based molecular embeddings weighted by their molar ratios, derived from a pre-trained MoLFormer transformer. These formulation-level descriptors were then passed to a downstream KAN model, enabling the prediction of logarithmic CE (LCE) with state-of-the-art accuracy. As depicted in Fig.~\ref{fig:DataDrivenKANs}(c), the model exhibits robust generalization, with parity plots (top) demonstrating tight alignment between predicted and actual LCE values across held-out samples, and ablation studies (bottom) confirming KAN’s superior performance. The study underscores KAN’s strengths in capturing complex, high-dimensional structure--property relationships from limited chemical data, while preserving interpretability through its modular, spline-based transformations.

In the context of classification tasks, Ta~\cite{ta2024bsrbf} introduced BSRBF-KAN to evaluate two standard classification benchmarks, MNIST and Fashion-MNIST, using a consistent network structure across all experiments. BSRBF-KAN achieved stable and competitive performance, reporting average accuracies of 97.55\% on MNIST and 89.33\% on Fashion-MNIST, while also exhibiting faster convergence and smoother loss decay across training epochs. This study revealed the importance of architectural components such as layer normalization and base output terms, while highlighting that B-spline components contributed more significantly to performance than RBFs. These results emphasize the potential of hybrid basis-function strategies within the KAN framework for enhancing expressivity and robustness in classification tasks. In the same line of KAN-based classification work, Chen et al.~\cite{chen2024sckansformer} introduced SCKansformer, a hybrid architecture tailored for fine-grained classification of bone marrow cells in microscopic images. Central to the model is a Kansformer encoder, which replaces the traditional MLP layers in Transformer blocks with KAN components to improve nonlinear feature extraction and model interpretability. This design is complemented by a spatial-channel convolutional encoder and a global-local attention encoder, together enhancing both redundancy reduction and context sensitivity in microimage analysis. The model was validated on a newly constructed Bone Marrow Cell Fine-Grained Classification Dataset, as well as publicly available benchmarks, demonstrating consistently superior accuracy and robustness across class-imbalanced and high-variability image sets. Ablation studies confirmed the critical contribution of the KAN-based module within the overall architecture, highlighting  its value in handling morphological subtleties and long-tailed class distributions typical of clinical image data. These results together show the versatility of KANs in  image classification tasks.

In the domain of image segmentation, Li et al.~\cite{li2025u} proposed U-KAN, a novel U-shaped architecture that integrates KANs into the encoder-decoder backbone of traditional convolutional segmentation models. Designed to address the challenges of low-contrast and high-variability tissue boundaries in medical imaging, U-KAN replaces MLP blocks with spline-parameterized KAN modules at the bottleneck and skip-connection layers. This hybrid structure enhances local adaptability and long-range dependency modeling without inflating parameter count. Applied to liver and tumor segmentation tasks from the LiTS dataset, U-KAN achieved state-of-the-art performance, surpassing Swin-UNet and TransUNet baselines in both Dice and HD95 metrics. As shown in Fig.~\ref{fig:DataDrivenKANs}(d), the model preserves sharper boundary details and finer anatomical structures, particularly in complex tumor regions. These results illustrate how KAN-enhanced modules can significantly improve segmentation quality in clinically relevant contexts, reinforcing their potential in data-driven biomedical image analysis.

\subsection{Comparison with Data-driven MLPs}

\begin{figure}[h]
    \centering
    \includegraphics[width=1\linewidth]{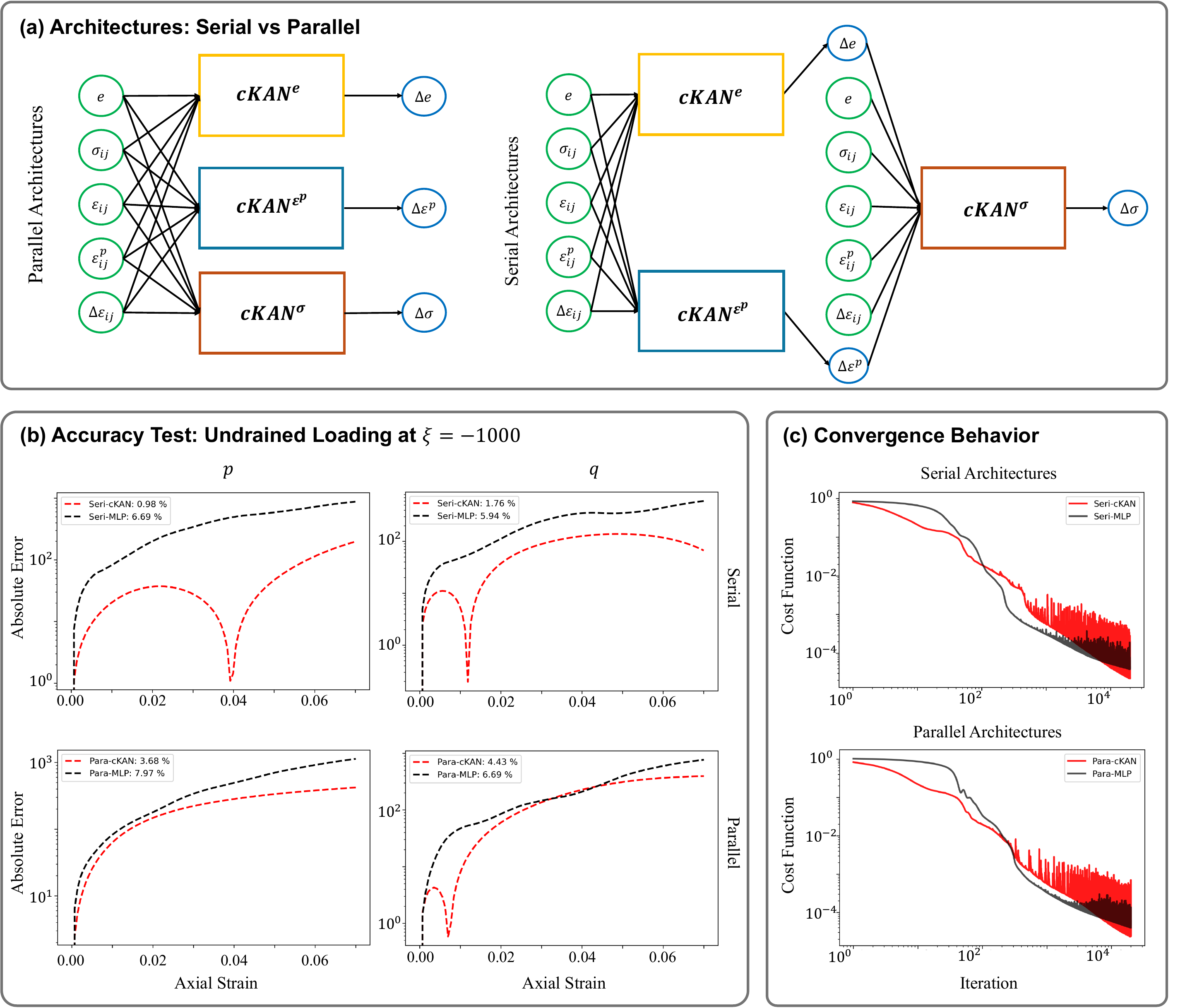}
    \caption{Performance comparison of cKAN and MLP architectures for data-driven modeling of sand elasto-plasticity.\textbf{(a)} Network architectures (serial and parallel). The models take as input a vector of stress $\boldsymbol{\sigma}$, strain $\boldsymbol{\varepsilon}$, plastic strain $\boldsymbol{\varepsilon}^{\text{p}}$, void ratio $e$, and the incremental strain $\Delta \boldsymbol{\varepsilon}\}$, and the output captures the corresponding changes in material response, including changes in void ratio $\Delta e$, incremental stress $\Delta \boldsymbol{\sigma}$, and incremental plastic strain $\Delta \boldsymbol{\varepsilon}^{\text{p}}\}$.\textbf{(b)} Prediction accuracy under a triaxial loading path defined by $\xi = -1000$, $p_{\text{in}} = 375$~kPa, and $e_{\text{in}} = 0.64$. The absolute errors in predicting the mean effective stress ($p$) and deviatoric stress ($q$) are assessed by comparing model outputs to reference results obtained from high-accuracy numerical integration.\textbf{(c)} Training convergence trends for serial and parallel architectures. The cKAN sub-networks use relatively compact architectures, with 2 layers of 20 neurons (degree 3) for the void ratio, and 3 layers of 20 neurons (degree 4) for both stress and plastic strain outputs. Moreover, the MLP counterparts employ deeper and wider networks, with 2 layers of 45 neurons for the void ratio and 5 layers of 35 neurons for stress and plastic strain, without polynomial basis functions. All panels present new analysis based on \cite{mostajeran2024epi}.}
    \label{fig:DataDrivenKANscomparision}
\end{figure}

\begin{figure}
    \centering
    \includegraphics[width=1\linewidth]{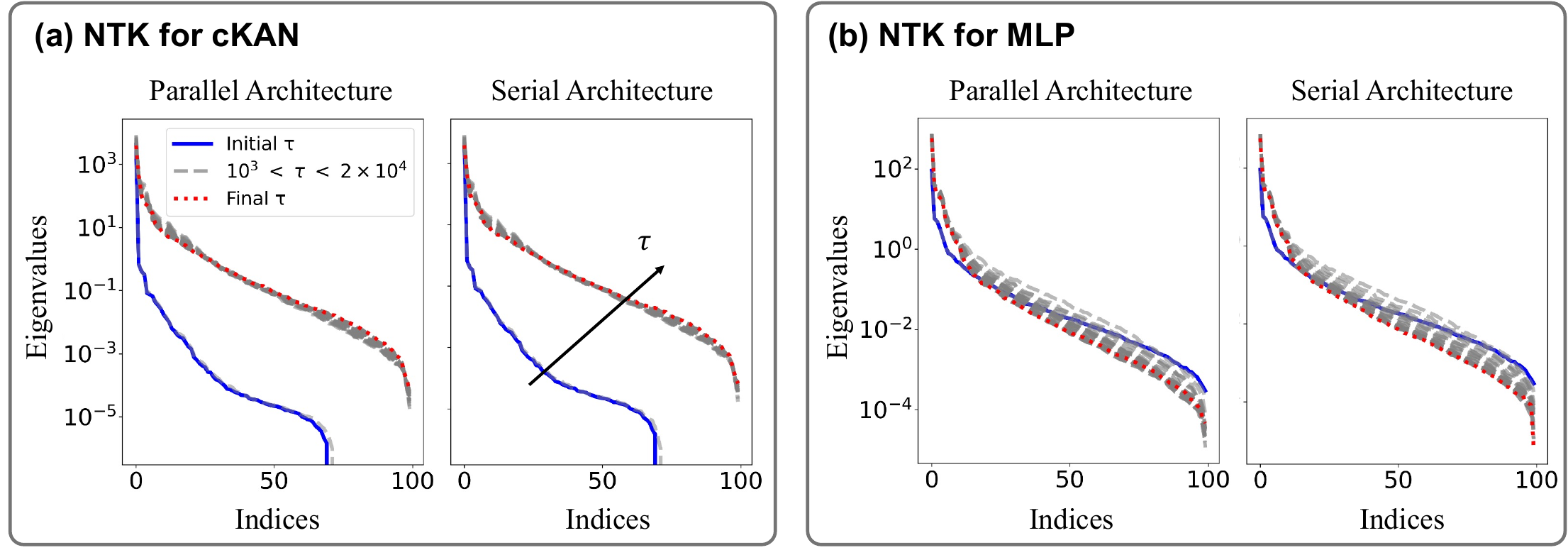}
    \caption{Evolution of NTK eigenvalues during training of the void ratio subnetwork $e$, using parallel and serial architectures introduced in~\cite{mostajeran2024epi}.
    \textbf{(a)} cKAN architectures. They exhibit more stable spectral distribution. 
    \textbf{(b)} MLP architectures. They show less structured and more dispersed eigenvalue spectra.
    All panels present new analysis based on \cite{mostajeran2024epi}.}
    \label{fig:EpicKAN_NTK}
\end{figure}

To better understand the advantages of KANs in data-driven learning tasks, it is essential to examine how they perform compared to vanilla MLPs. Such a comparison is significant given the widespread use of MLPs as standard function approximators and the growing interest in KANs as interpretable and efficient alternatives. In this section, we compare the two network types based on three fundamental aspects: accuracy, convergence behavior, and training dynamics analyzed through the Neural Tangent Kernel (NTK)~\cite{jacot2018neural, faroughi2025neural} framework. Accuracy evaluates how well each model captures the underlying data distribution, while convergence analysis reflects training efficiency and stability. NTK analysis provides further insight into the models' training behavior by examining the spectral properties of their kernel matrices. These spectral metrics help explain differences in learning dynamics and generalization capacity. Together, these three criteria provide valuable insights into the practical performance of KANs versus MLPs in data-driven settings. Fig.~\ref{fig:DataDrivenKANscomparision} and Fig.~\ref{fig:EpicKAN_NTK} illustrate the comparative results, highlighting the relative strengths and weaknesses of the KAN and MLP models across all three criteria. Fig.~\ref{fig:DataDrivenKANscomparision}(a) presents the architectural layouts used in~\cite{mostajeran2024epi}, where parallel and serial configurations were applied to both KAN and MLP models to predict the elasto-plastic response of sand under unseen loading paths.

\subsubsection{Accuracy Comparison}

Assessing model accuracy is fundamental to understanding how effectively learning frameworks replicate complex behaviors in data-driven tasks. 
Several studies have compared the accuracy of KAN models across different application domains. In comparative experiments, the P1-KAN~\cite{warin2024p1} achieved higher accuracy than multilayer perceptrons and other KAN variants for both smooth and highly irregular functions, demonstrating strong robustness across different regularity regimes. In time-series forecasting tasks~\cite{vaca2024kolmogorov}, KANs tracked rapid changes in traffic volume more accurately than MLPs, especially during sudden shifts. In image segmentation, U-KAN~\cite{li2025u} achieved a better balance between accuracy and efficiency compared to U-Net.
In addition, FC-KAN~\cite{ta2024fc} outperformed MLP models by combining basis functions such as Difference of Gaussians (DoG) and B-splines, with the FC-KAN (DoG+BS) variant achieving the highest accuracy using a quadratic output representation. The study in~\cite{mostajeran2024epi} also examined accuracy by evaluating model predictions during the second stage, where trained networks forecast material responses under unseen loading paths. Fig.~\ref{fig:DataDrivenKANscomparision}(b) illustrates the absolute error profiles for key stress components, highlighting differences between cKAN and MLP models in both serial and parallel architectures. Across all configurations, cKAN models consistently demonstrate notably lower errors than their MLP counterparts, indicating superior fidelity in representing nonlinear elasto-plastic behavior. Specifically, serial cKAN architectures reduce prediction errors by a substantial margin compared to serial MLPs, while parallel cKANs also show meaningful improvements over parallel MLPs. These findings suggest that the structured functional form employed by cKANs enhances their ability to accurately model complex stress-strain relationships, supporting their promise for data-driven constitutive modeling in geomechanics and related fields.

\subsubsection{Convergence Analysis}

Analyzing convergence patterns offers valuable insight into how different models utilize training iterations to refine their predictions.  
Several studies have examined convergence behavior in KAN-based models. For example, BSRBF-KAN~\cite{ta2024bsrbf}, which combines B-splines and radial basis functions, showed faster convergence compared to MLPs in various learning tasks. Similarly, FC-KAN~\cite{ta2024fc}, based on function combinations in KANs, consistently achieved lower training losses than MLPs, attributed to its fast convergence properties. The P1-KAN~\cite{warin2024p1} also exhibited stable and consistent convergence behavior, with smooth error reduction during training, particularly in stochastic optimization problems such as hydraulic valley modeling. Convergence was also analyzed in~\cite{mostajeran2024epi}, where parallel and serial models were trained as part of a two-stage approach. Fig.~\ref{fig:DataDrivenKANscomparision}(c) shows the evolution of the cost function during the training stage. The results indicate that cKAN models, in both configurations, exhibit a gradual and sustained decrease in the cost function, reflecting stable and ongoing learning. In contrast, MLP models tend to converge more quickly and often plateau early, particularly in the serial setup, suggesting premature stagnation. While cKANs display some variability near convergence, especially in the parallel case, these fluctuations occur around a consistently declining trend, indicating active adaptation rather than instability. MLPs, however, show abrupt loss reduction followed by flat regions, which implies limited further learning. These convergence dynamics suggest that cKANs are better suited for extended training and offer more reliable long-term improvement, which may ultimately lead to stronger generalization capabilities.

\subsubsection{Spectral Behavior}

To compare the spectral behavior of cKAN and MLP, we use the NTK framework to track the evolution of the eigenvalue spectrum during training. NTK theory, introduced by Jacot et al.~\cite{jacot2018neural}, describes how infinitely wide neural networks evolve under gradient descent, showing that in this limit, the NTK remains constant throughout training. This allows learning to be studied in function space, where convergence rates are directly determined by the NTK’s eigenvalues, directions associated with larger eigenvalues converge more rapidly. Building on this foundation, Seleznova et al.~\cite{seleznova2022analyzing, seleznova2022neural} extended NTK analysis to practical, finite-width networks and demonstrated that the NTK can vary significantly during training, particularly in deep networks or under poor initialization. Their results showed that NTK variability increases with network depth, especially in regimes characterized as chaotic or at the edge of chaos. Most NTK-related studies in purely data-driven settings have focused on analyzing MLPs, while limited attention has been given to KANs. Building on the ideas proposed by Mostajeran \& Faroughi~\cite{mostajeran2024epi}, Fig.~\ref{fig:EpicKAN_NTK} presents the evolution of the NTK eigenvalue spectrum for both cKAN and MLP models during the training of networks tasked with predicting the void ratio $e$, across parallel and serial architectural configurations. 
In all experiments, the networks operate in the finite-width regime, where each layer contains a limited number of neurons and the weights evolve throughout training. As a result, the NTK is not constant but evolves dynamically over time, capturing feature learning effects rather than purely kernel regression behavior.
The cKAN demonstrates a consistently stable and compact spectrum, characterized by low dispersion and smooth progression over time. In contrast, the MLP exhibits irregular spectral dynamics, with final eigenvalues in some cases falling below their initial values, indicating inefficient propagation of learning networks. Rigas et al.~\cite{rigas2025initialization} also emphasized the link between initialization and NTK dynamics in spline-based KANs. They showed that baseline schemes often produce spectra that collapse during training, leading to poor conditioning, whereas Glorot- and power-law-based strategies yield eigenvalue spectra that stabilize early and remain well-conditioned. These insights reinforce the importance of initialization in maintaining stable NTK behavior, in line with the spectral stability observed for KANs. These findings underscore the superior spectral conditioning and training stability of cKANs, which contribute to their enhanced convergence and generalization in data-driven modeling tasks.

\section{Physics-informed KANs}\label{Sec:PIKANs}

Physics-informed Kolmogorov–Arnold Networks extend data-driven KANs by embedding domain knowledge, such as partial differential equations, conservation laws, and boundary conditions, directly into the network architecture and training process~\cite{shukla2024comprehensive, wang2024kolmogorov}. This section provides a structured overview of PIKANs, emphasizing their integration of physical laws into learning-based models. The first part details the architectural components and training strategies that enable PIKANs to incorporate governing equations, boundary conditions (e.g., PDEs), and initial constraints into the learning process. The second part highlights a range of applications in solid mechanics, fluid dynamics, and inverse modeling, demonstrating how PIKANs can resolve complex field behaviors and capture sharp gradients or nonlinear material responses with high fidelity. The third part presents a comparative analysis between PIKANs and conventional PINN frameworks, focusing on accuracy, convergence behavior, and spectral characteristics to evaluate their relative performance in solving physics-based problems.

\begin{figure}[!ht]
    \centering
    \includegraphics[width=1\linewidth]{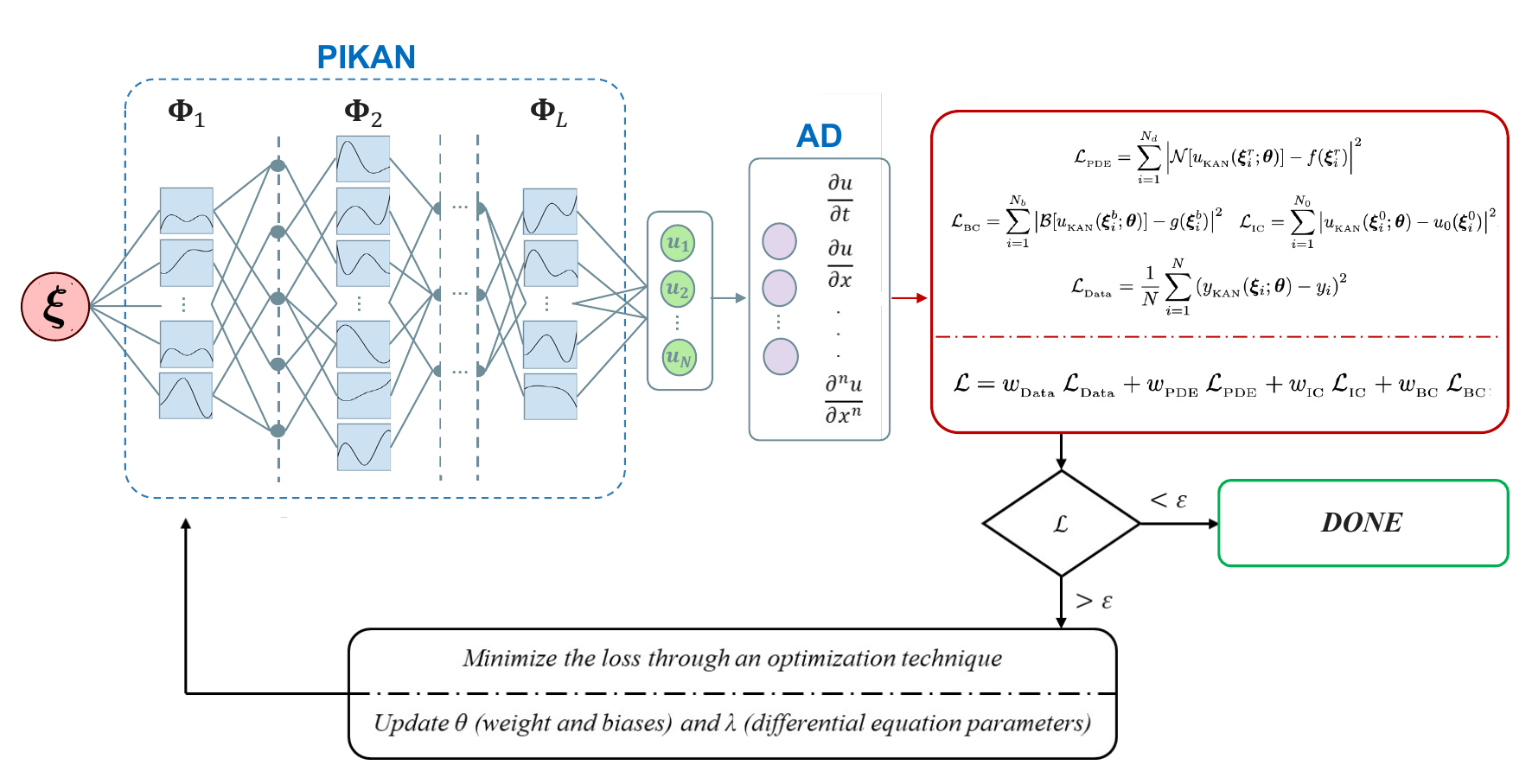}
    \caption{Schematic architecture of PIKANs. The network receives spatiotemporal coordinates as input and approximates the multiphysics solution $u$ (forward modeling) as well as unknown parameters in the governing equation (inverse modeling). The final layer computes derivatives of $u$ with respect to input variables, which are used to formulate the residuals of the strong form of governing equations and data loss terms, each weighted by distinct coefficients. The learnable parameters $\boldsymbol{\theta}$ (neural weights and biases) and $\boldsymbol{\lambda}$ (unknown parameters) are optimized simultaneously through loss minimization.}
    \label{fig:PIKAN}
\end{figure}

\subsection{Architecture and Training Strategies}

Physics-informed KANs integrate physical constraints into their architecture and training process to ensure physically consistent predictions~\cite{wang2024kolmogorov}. Unlike purely data-driven approaches, PIKANs leverage the strong form of the governing equations in the loss to enhance accuracy and generalization~\cite{shukla2024comprehensive}. As illustrated in Fig.~\ref{fig:PIKAN}, the PIKAN architecture is designed to approximate the solutions of physical systems governed by PDEs/ODEs or other closure models by mapping spatiotemporal input samples to physically meaningful output quantities. The network takes as input a set of spatiotemporal domain samples, denoted by $\boldsymbol{\xi} = (\boldsymbol{x}, t)$, where $\boldsymbol{x} \in \Omega \subset \mathbb{R}^d$ represents the spatial coordinates and $t \in [0, T]$ denotes time. The corresponding output represents physically relevant quantities, such as pressure, velocity, displacement, or concentration, depending on the specific application. Let the behavior of the system be governed by a PDE of the general form,
\begin{equation}\label{Eq.PDEpikan}
\mathcal{N}[u(\boldsymbol{x}, t)] = f(\boldsymbol{x}, t), \quad \text{in\: } \Omega \times (0, T],
\end{equation}
subject to the initial condition,
\begin{equation}\label{Eq.PDEpikanIC}
u(\boldsymbol{x}, 0) = u_0(\boldsymbol{x}), \quad \text{for\: } \boldsymbol{x} \in \Omega,
\end{equation}
and the boundary condition,
\begin{equation}\label{Eq.PDEpikanBC}
\mathcal{B}[u(\boldsymbol{x}, t)] = g(\boldsymbol{x}, t), \quad \text{on\: } \partial\Omega \times (0, T],
\end{equation}
where $u$ is the solution to be approximated by the network, $\mathcal{N}$ is a (possibly nonlinear) differential operator that encodes the governing physical laws, $f$ is a source or forcing term, $u_0$ defines the initial state of the system, and $\mathcal{B}$ represents a boundary operator with associated boundary values $g$. To train PIKANs, a composite loss function is used to enforce consistency with observed data and compliance with physical constraints. This loss function is defined as~\cite{raissi2019physics, shukla2024comprehensive},
\begin{equation}\label{Eq.LossPIKAN}
    \mathcal{L} =  w_{_\text{Data}}\: \mathcal{L}_{_\text{Data}} + w_{_\text{PDE}}\: \mathcal{L}_{_\text{PDE}} + w_{_\text{IC}} \:\mathcal{L}_{_\text{IC}} + w_{_\text{BC}}\: \mathcal{L}_{_\text{BC}},
\end{equation}
where the weighting factors \( w_{_\text{Data}}, w_{_\text{PDE}}, w_{_\text{IC}}, w_{_\text{BC}} \) balance the influence of each term during optimization. The individual components of the composite loss function in Eq.~\eqref{Eq.LossPIKAN} are defined as follows. The data loss $\mathcal{L}_{_\text{Data}}$ quantifies the discrepancy between the network prediction $u_{_\text{KAN}}(\boldsymbol{\xi}^d_i; \boldsymbol{\theta})$, which depends on the model parameters $\boldsymbol{\theta}$, and the ground-truth measurements $u(\boldsymbol{\xi}^d_i)$, as defined in Eq.~\eqref{Eq.LossData}. This loss is evaluated over the set of spatiotemporal data points $\{\boldsymbol{\xi}^d_i \in \Omega \times (0,T]\}_{i=1}^{N_d}$. The set of data points ${\boldsymbol{\xi}^d_i}$ and their corresponding ground-truth values $u(\boldsymbol{\xi}^d_i)$ are (only) required in inverse problems, where certain components of the PDE formulation, such as coefficients, initial conditions, or boundary conditions, are partially unknown. In purely forward problems, where the full PDE and associated conditions are known, this dataset and the related data loss term are not required. The PDE loss $\mathcal{L}_{_\text{PDE}}$ enforces the governing physics by penalizing the residuals of the PDE at a set of residual points $\{\boldsymbol{\xi}_i^r \in \Omega \times (0,T]\}_{i=1}^{N_r}$,
\begin{equation}
    \mathcal{L}_{_\text{PDE}} = \sum_{i=1}^{N_d} \Big{\vert} \mathcal{N}[u_{_\text{KAN}}(\boldsymbol{\xi}_i^r; \boldsymbol{\theta})] - f(\boldsymbol{\xi}_i^r) \Big{\vert}^2.
\end{equation}

The initial condition loss $\mathcal{L}_{_\text{IC}}$ and boundary condition loss $\mathcal{L}_{_\text{BC}}$ ensure that the network satisfies the initial and boundary constraints, respectively, that are defined as,
\begin{equation}  
    \mathcal{L}_{_\text{IC}} = \sum_{i=1}^{N_{0}} \big\vert u_{_\text{KAN}}(\boldsymbol{\xi}_i^0; \boldsymbol{\theta}) - u_0(\boldsymbol{\xi}_i^0) \big\vert^2, \quad  
    \mathcal{L}_{_\text{BC}} = \sum_{i=1}^{N_{b}} \big\vert \mathcal{B}[u_{_\text{KAN}}(\boldsymbol{\xi}_i^b; \boldsymbol{\theta})] - g(\boldsymbol{\xi}_i^b) \big\vert^2,  
\end{equation}  
where $\{\boldsymbol{\xi}_i^0 \in \Omega \times \{t = 0\}\}_{i=1}^{N_0}$ and $\{\boldsymbol{\xi}_i^b \in \partial\Omega \times (0,T]\}_{i=1}^{N_b}$ denote the sets of initial and boundary points, respectively.

These constraints may be imposed as either soft or hard. The key difference between soft and hard constraints involves the integration of physical conditions within the learning framework. For soft constraints~\cite{shukla2024comprehensive}, physical laws and conditions are enforced indirectly by adding penalty terms to the loss function akin to the approach outlined in Eq.~\eqref{Eq.LossPIKAN}. This approach encourages the network to approximate the desired behavior but does not guarantee strict satisfaction of the constraints. In contrast, hard constraints are embedded directly into the network architecture~\cite{lu2021physics, lai2025hard}, ensuring that the output inherently satisfies the initial and boundary conditions. This is typically achieved by re-parameterizing the network output as,
\begin{equation}
u_{\boldsymbol{\theta}}(\boldsymbol{\xi}) = g(\boldsymbol{\xi}) + h(\boldsymbol{\xi})\: u_{_\text{KAN}}(\boldsymbol{\xi}; \boldsymbol{\theta}),
\end{equation}
where $g$ is a known function that exactly satisfies the boundary (and possibly initial) conditions, and $h$ is a smooth function that vanishes on the boundaries. The term $u_{_\text{KAN}}(\cdot ; \boldsymbol{\theta})$ is the output of the trainable KAN-based network. This formulation guarantees that the resulting solution respects the hard constraints by design.

Training PIKANs also involves optimizing the complex, multi-term loss function (Eq.~\eqref{Eq.LossPIKAN}), which requires careful selection of optimization strategies. Although first-order methods such as Adam are widely used, recent research has shown that second-order optimizers, particularly self-scaled Broyden (SSBroyden)~\cite{al2014broyden} and BFGS with Wolfe line search~\cite{nawi2006improved, saputro2017limited, liu1989limited}, can significantly accelerate convergence and improve accuracy in training PIKANs~\cite{kiyani2025optimizer}. These quasi-Newton methods exploit curvature information from the loss landscape, helping to avoid local minima and saddle points that often trap first-order optimizers. A promising approach is to use hybrid optimization strategies, where the network is first trained with adaptive first-order methods like RAdam to quickly reduce the loss, followed by second-order fine-tuning with BFGS variants. This combination has been shown to achieve high accuracy and stable convergence across various problems~\cite{daryakenari2025representation}. The effectiveness of these optimizers also depends on architectural and numerical factors. For example, the use of Chebyshev polynomial-based activation in tanh-cPIKANs~\cite{daryakenari2025representation} enhances optimization by producing smoother gradients, while double-precision arithmetic provides better numerical stability and optimization performance. Additionally, adaptively adjusting the weights of different loss components helps to balance learning signals and avoid domination by any single term during training~\cite{liu2022physics, gao2025physics, heydari2019softadapt, barron2019general}.

\begin{table}[h]
\centering
\small
\renewcommand{\arraystretch}{1.05}
\caption{A non-exhaustive list of recent studies on structural and methodological developments in PIKANs, showcasing various architectures, basis functions, and application domains. These advances highlight the adaptability of KAN-based models for solving diverse forward and inverse PDE problems.}
\label{Tab:PIKAN_Advances_Table}
\footnotesize
\begin{tabularx}{\textwidth}{>{\hsize=0.2\hsize}X >{\hsize=0.75\hsize}X >{\hsize=0.15\hsize}r}
\toprule  
\textbf{Method} & \textbf{Objective and Application Domain} & \textbf{Ref.} \\ 
\midrule
PIKAN &
Used different variants of KAN (B-spline, Chebyshev, Legendre, Hermit, Jacoiy, Bayesian-PIKAN); Applied to  Helmholtz, Navier–Stokes, Allen–Cahn, and Reaction–diffusion equations. 
& \cite{shukla2024comprehensive} \\
\addlinespace
\addlinespace
Scaled-cPIKAN & 
Combined Chebyshev-based KANs with a scaling approach for spatial variables and governing equations; Applied to PDEs with oscillatory dynamics over large domains, such as diffusion, Helmholtz, Allen–Cahn, and reaction–diffusion equations. 
& \cite{mostajeran2025scaled}\\
\addlinespace
\addlinespace
KAN-ODE 
& Proposed an interpretable and modular framework without prior knowledge by integrating KANs into a Neural ODE structure;
Used Gaussian RBFs  basis functions;
Applied to Lotka–Volterra, Fisher-KPP, Burgers’, Schr\"{o}dinger, and Allen-Cahn equations.
& \cite{koenig2024kan} \\
\addlinespace
\addlinespace
HRKANs & 
Enhanced the approximation capability of KANs by introducing higher-order ReLU basis functions in place of linear splines; Applied to the linear Poisson equation and the nonlinear Burgers’ equation with viscosity.& \cite{so2024higher}\\
\addlinespace
\addlinespace
SPIKAN & 
Introduced a separable architecture for PIKANs by applying the principle of variable separation; Used B-splines as basis functions; 
Applied to 2D Helmholtz, 2D steady lid-driven cavity flow, 1D+1 Allen–Cahn, and 2D+1 Klein–Gordon equations& \cite{jacob2024spikans}\\
\addlinespace
\addlinespace
KKAN &
Designed a two-block architecture with MLP-based inner functions and basis-function-based outer functions to form a universal approximator;
Used RBFs as outer blocks;
Applied to the Allen–Cahn equation. 
&\cite{toscano2025kkans}\\
\addlinespace
\addlinespace
 Bayesian PIKANs & 
 Proposed a gradient-free method combining DTEKI with Chebyshev KANs, with incorporation of active subspaces method; 
 Applied to transport, 1D diffusion, 1D nonlinear and 2D Darcy flow equations.
 & \cite{gao2025scalable}\\	
\addlinespace
\addlinespace
 LeanKAN &
Proposed a direct and modular alternative to MultKAN and AddKAN layers to reduce parameters and hyperparameters;
Used RBF expansions combined with a base activation function;
Applied to the classical Lotka–Volterra predator–prey model.
& \cite{koenig2025}\\
\addlinespace
\addlinespace
asKAN & 
Proposed active subspace embedded KAN to circumvent the inflexibility and organized KANs in a hierarchical framework;
Used a third-order spline function, with 5 grid points;
Applied to the Poisson equation and scattering sound field reconstruction.	
& \cite{zhou2025}\\
\bottomrule
\end{tabularx}
\vspace{-10pt}
\end{table}

Research on architectural advancements in PIKANs has led to various structural modifications tailored to specific tasks. A non-exhaustive overview of these developments is provided in Table~\ref{Tab:PIKAN_Advances_Table}. A key line of work is presented in~\cite{shukla2024comprehensive}, where extensive evaluations of different basis functions were conducted. In this study, the B-spline activations, commonly used in early KAN architectures, were replaced with orthogonal polynomial functions such as Chebyshev, Jacobi, and Legendre. These replacements were further improved through the incorporation of residual-based attention and entropy viscosity methods. To improve scalability, new variants such as DF-PIKANs and DD-PIKANs~\cite{patra2024physics} were introduced. These models leverage wavelet and B-spline activations to reduce model complexity while preserving accuracy across a wide range of PDEs, particularly those dominated by convection.
Building on this, SPIKAN~\cite{jacob2024spikans} employed a separation-of-variables strategy, effectively reducing both training cost and memory usage in high-dimensional scenarios. Furthermore, HRKANs~\cite{so2024higher} adopted Higher-order-ReLU activations, aiming to streamline implementation and enhance solver performance for problems such as Poisson and Burgers’ equations.

\subsection{Applications}

\begin{figure}[h]
    \centering
    \includegraphics[width=1\linewidth]{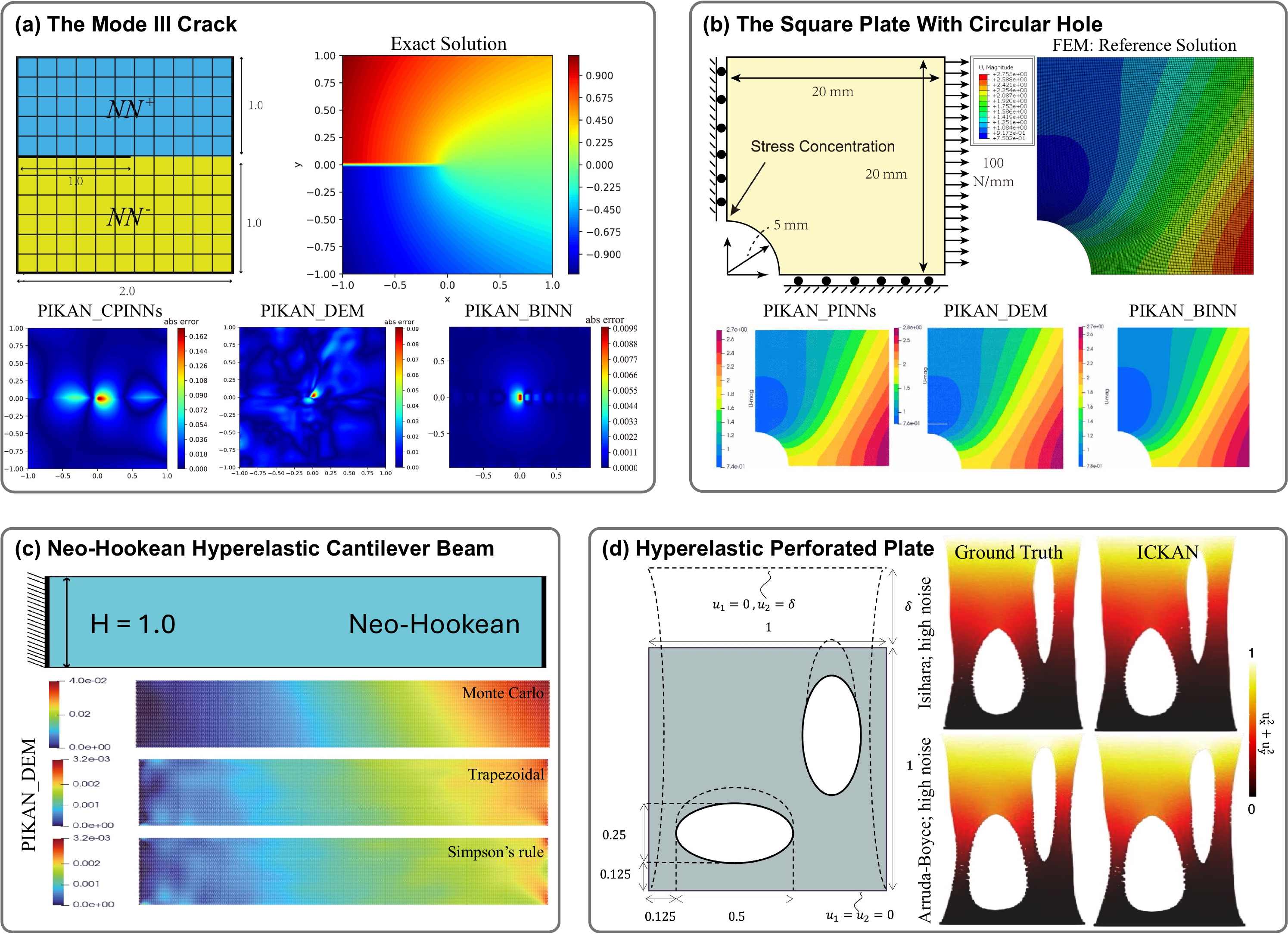}
    \caption{\textbf{(a)} The mode III crack problem is defined over a square domain of $[-1,1]^2$, with the angular coordinate $\theta$ ranging from $[-\pi, \pi]$. The analytical displacement solution is given by $u = {r^{1/2}} \sin(\theta/2)$. This panel is adopted from~\cite{wang2024kolmogorov}. \textbf{(b)} The model consists of a square plate with a 20 mm edge length and a centrally located circular hole of 5 mm radius, with predicted displacement solutions shown for PIKAN versions. This panel is adopted from~\cite{wang2024kolmogorov}. \textbf{(c)} Description of the Neo-Hookean hyperelastic cantilever beam problem with dimensions of: height $H = 1.0$, length $L = 4.0$, $E = 1000$, and $\nu = 0.3$. The left end $x = 0.0$ is fixed, and a uniformly distributed downward load $\bar{t} = 5$ is applied at the right end. Distributions of absolute errors for the Neo-Hookean hyperelastic cantilever beam obtained from the DEM using the KAN approach. This panel is adopted from~\cite{wang2024kolmogorov}. \textbf{(d)} A square plate with two asymmetrically positioned elliptical holes is subjected to uniaxial tension under displacement control. The resulting deformed configuration, with displacement magnitudes represented by color contours, is shown for both the ground-truth data and the ICKAN-based simulation results. This panel is adopted from~\cite{thakolkaran2025can}.}
    \label{fig:PIKANS-Solids}
\end{figure}

PIKAN variants have been applied to address specific computational challenges, such as resolving high-dimensional, non-linear PDEs in both forward and inverse problems.  A non-exhaustive list of recent applications of PIKANs is provided in Table~\ref{Tab:PIKAN_Advances_Table_Part2}. For example, in the solid mechanics domain, PIKAN has been used to model fracture mechanics problems characterized by displacement singularities and sharp stress gradients. In this context, Wang et al.~\cite{wang2024kolmogorov} investigated the Mode III crack problem, which involves intense singularities near the crack tip, and demonstrated that the PIKAN framework successfully captured the steep stress gradients and singular fields with notable accuracy. The results indicate that PIKAN effectively resolves near-tip singularities while maintaining stable convergence and efficient training. Unlike conventional architectures that often struggle with the spectral bias when learning high-frequency features near singular points, PIKAN’s ability to adaptively learn univariate activation functions allows for a more precise representation of the steep gradients inherent to fracture mechanics. Quantitative evaluations against analytical solutions and finite element results show that PIKAN accurately predicts both displacement and stress fields near the crack front (see Fig.~\ref{fig:PIKANS-Solids} (a)).

\begin{table}[h]
\centering
\small
\renewcommand{\arraystretch}{1.05}
\caption{A non-exhaustive list of recent studies highlighting application-oriented advances in PIKANs. These methods demonstrate the flexibility of KAN-based architectures for solving complex PDEs across diverse scientific and engineering domains.}
\label{Tab:PIKAN_Advances_Table_Part2}
\footnotesize
\begin{tabularx}{\textwidth}{>{\hsize=0.2\hsize}X >{\hsize=0.75\hsize}X >{\hsize=0.15\hsize}r}
\toprule  
\textbf{Method} & \textbf{Objective and Application Domain} & \textbf{Ref.} \\ 
\midrule
KAN-PointNet & 
Combined KANs with PointNet to predict incompressible steady-state fluid flow over irregular domains;
Utilized Jacobi polynomials in shared KANs;
Applied to incompressible laminar steady-state flow over a cylinder with varying cross-sectional geometries.
&\cite{KASHEFI2025117888}\\
\addlinespace
\addlinespace
PI-KAN-PointNet	&
Proposed  physics-informed KAN-PointNet for simultaneous inverse solutions over multiple geometries;
Employed Jacobi polynomials;
Applied to incompressible steady-state fluid flow predicting velocity, pressure, and temperature across 135 domains.
&\cite{kashefi2025physics}\\
\addlinespace
\addlinespace
ChebPIKAN & 
Combined KANs with Chebyshev polynomial basis functions and physics-informed loss to solve fluid dynamics PDEs efficiently; Applied to Allen–Cahn, nonlinear Burgers', 2D Helmholtz, 2D Kovasznay flow, and 2D Navier–Stokes equations.
& \cite{guo2024physics}\\
\addlinespace
\addlinespace
Legendre-KAN & Proposed a neural network framework using Legendre polynomial basis; Applied to high-dimensional and singular fully nonlinear Monge–Amp\'{e}re equations with Dirichlet boundary conditions and the transport problem (Brenier Theorem). 
&\cite{hu2025s}\\
\addlinespace
\addlinespace
tanh-cPIKAN	&  
Introduced a Chebyshev-based PIKAN variant; Applied to inverse problems in pharmacology systems, including pharmacokinetics (single-dose compartmental model) and pharmacodynamics (chemotherapy drug-response modeling).	 
& \cite{daryakenari2025representation}\\
\addlinespace
\addlinespace
Anant-Net &
Proposed a KAN-based surrogate framework tailored for solving high-dimensional PDEs on hypercubic domains; 
Employed  spline based basis functions;
Applied to Poisson, Sine-Gordon, and Allen–Cahn equations up to 300 dimensions.
&\cite{menon2025}\\
\addlinespace
\addlinespace
AIVT &
Introduced artificial intelligence velocimetry-thermometry for reconstructing temperature and velocity fields; 
Used Chebyshev KANs;
Applied to thermal turbulence in Rayleigh–B\'{e}nard convection using experimental 3D velocity data.	&\cite{toscano2025aivt}\\
\addlinespace
\addlinespace
ICKANs & Proposed monotonic input-convex KANs for learning polyconvex constitutive models in hyperelasticity; 
Used uniform B-splines;
Applied to Neo-Hookean, Isihara, Haines-Wilson, Gent-Thomas, Arruda-Boyce and Ogden models.& \cite{thakolkaran2025can}\\

\bottomrule
\end{tabularx}
\vspace{-10pt}
\end{table}

In another benchmark by Wang et al.~\cite{wang2024kolmogorov}, a plate with a central hole was subjected to uniaxial tension to assess the model's ability to resolve stress concentrations, a crucial test for evaluating the fidelity of surrogate models in capturing localized field amplification. PIKAN demonstrated superior performance in reconstructing both displacement and stress fields in the vicinity of the hole, where stress gradients become highly localized due to geometric discontinuity. PIKAN achieved an accurate approximation of steep solution features with efficient convergence, requiring fewer training epochs. Fig.~\ref{fig:PIKANS-Solids}(b) illustrates the initial configuration along with the corresponding solutions obtained using the FEM and PIKAN frameworks. This efficiency stems from the KAN structure’s ability to represent sharp variations in stress distribution without excessive network depth or tuning.

Beyond linear elasticity, the method was extended to simulate nonlinear hyperelastic responses using a Neo-Hookean material model by Wang et al.~\cite{wang2024kolmogorov}. Under large deformations and boundary-driven displacement conditions, PIKAN approximated the energy functional and displacement field with high fidelity, yielding results consistent with reference finite element solutions. The variational formulation adopted in this case allowed PIKAN to learn the equilibrium configuration by minimizing the stored strain energy, a strategy that proved effective in capturing nonlinear stress–strain behavior without explicit supervision of the stress field. One notable advantage observed was PIKAN’s robustness in handling geometric and material nonlinearity without the need for architectural adjustments or retraining instability, which are common issues in traditional PINNs under large-deformation regimes. Moreover, the ability of the network to preserve physical constraints such as incompressibility and boundary condition adherence was markedly improved, owing to the localized adaptability of KAN-based activations. Fig.~\ref{fig:PIKANS-Solids}(c) presents the Neo-Hookean hyperelastic cantilever beam configuration along with the corresponding absolute error contours obtained from the DEM using the KAN framework.

To show the potential of PIKANs in solid mechanics, Thakolkaran et al.~\cite{thakolkaran2025can} proposed a monotonic Input-Convex KAN (ICKAN) architecture tailored to identify hyperelastic functionals that adhere to polyconvexity constraints. Unlike traditional supervised training on stress–strain data, their model was trained in an unsupervised manner using full-field strain data and global reaction forces, mimicking digital image correlation (DIC) experiments. The ICKAN model enforced convexity with respect to input invariants and monotonicity in the output through structured B-spline activation layers, ensuring thermodynamic consistency and stable optimization. Results demonstrated that the learned energy models accurately captured the underlying physical behavior and generalized well beyond the training distribution, including to unseen geometries and deformation modes. Fig.~\ref{fig:PIKANS-Solids}(d) shows finite element solutions for the Isihara and Arruda-Boyce ground-truth models, which closely match the predictions produced by the ICKAN-based models.

\begin{figure}[h!]
    \centering
    \includegraphics[width=1.0\linewidth]{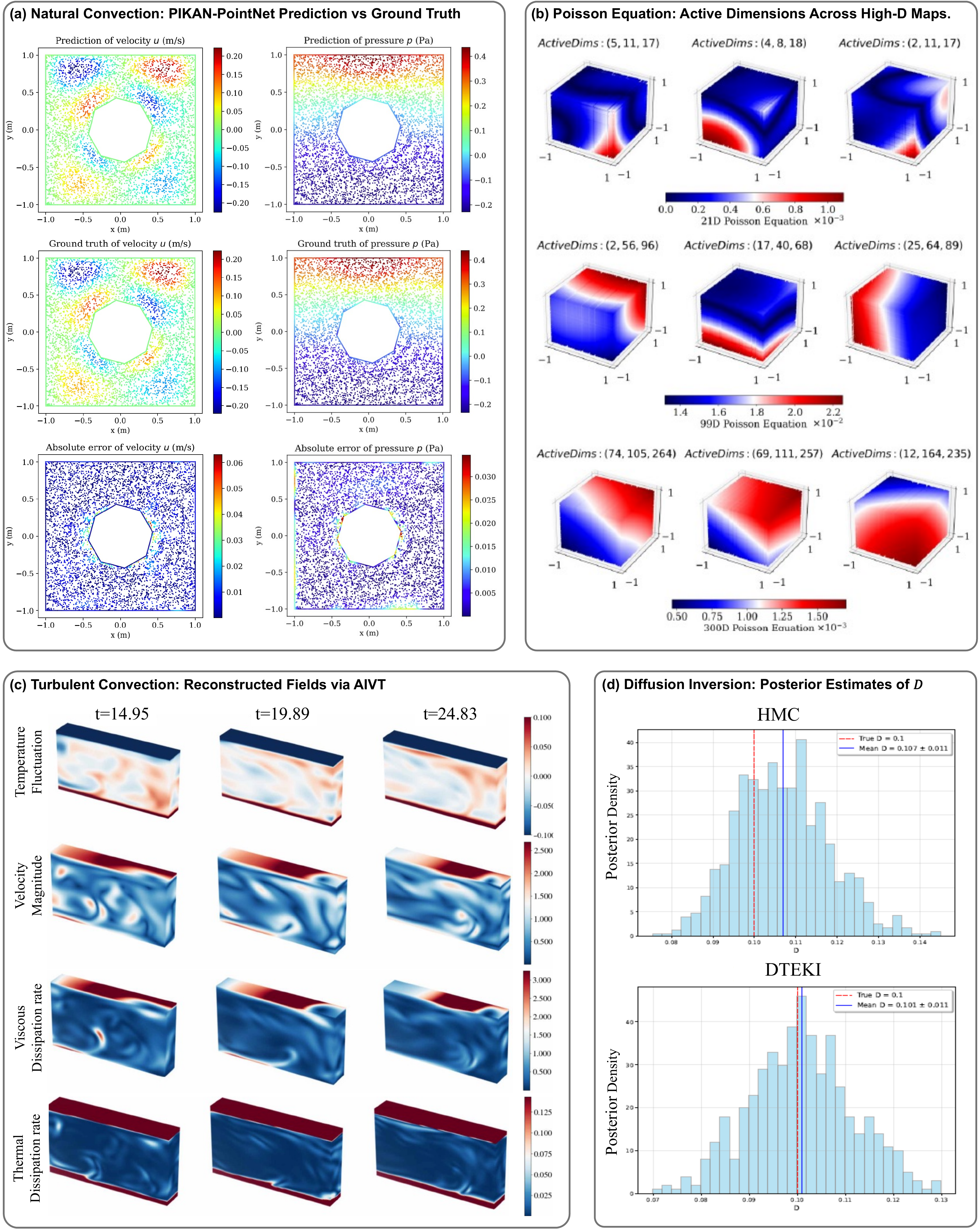}
    \caption{\textbf{(a)} Comparison between the reference solution and the predictions generated by the physics-informed KAN PointNet for the velocity and pressure fields. The model employs Jacobi polynomials of degree 2, with parameters $\alpha = \beta = -0.5$, and uses a sample density of $n_s = 0.5$. This panel is adopted from~\cite{kashefi2025physics}. \textbf{(b)} Absolute point-wise testing error evaluated on randomly selected 3D subspaces extracted from the 21D, 99D, and 300D Poisson equation benchmarks. The notation ``\textbf{ActiveDims:} \{p, q, r\}'' refers to the dimensions in the $d$-dimensional space where the gradient is non-zero, while $\text{D} \setminus \{p, q, r\}$ indicates the remaining inactive dimensions. Here, $\text{D} = \{1, 2, \dots, d\}$ represents the complete set of dimensions. This panel is adopted from~\cite{menon2025}. \textbf{(c)} Snapshots illustrate the evolution of the flow over 10 free-fall time units. This panel is adopted from~\cite{toscano2025aivt}. \textbf{(d)} Diffusion equation: Histogram of predicted values for $D$ obtained using Hamiltonian Monte Carlo (HMC) and Dropout Tikhonov Ensemble Kalman Inversion (DTEKI) with Chebyshev-based KANs. This panel is adopted from~\cite{gao2025scalable}.
}
    \label{fig:fluid mechanics}
\end{figure}

In addition to solid mechanics, PIKAN-based frameworks have also been extended to model complex fluid flow phenomena. In this context, Kashefi et al.~\cite{kashefi2025physics} proposed the PI-KAN-PointNet framework for reconstructing steady-state velocity, pressure, and temperature fields in natural convection problems across 135 two-dimensional domains with internal non-circular cylinders. The network was tasked with inferring flow fields from sparse sensor data and partial boundary conditions, particularly estimating the unknown temperature distribution along the surface of the inner cylinder. Results showed that PI-KAN-PointNet achieved high prediction accuracy, with relative errors below 3\% for pressure and temperature fields, and under 11\% for velocity components, even in the presence of sharp gradients induced by geometric discontinuities. Fig.~\ref{fig:fluid mechanics}(a) shows that the physics-informed KAN PointNet accurately reconstructs the velocity and pressure fields across complex geometries, achieving close agreement with the reference solution. The model’s ability to adaptively learn activation functions via Jacobi polynomial expansions proved especially effective near irregular boundaries. Visual comparisons with finite element ground truth confirmed the framework’s strength in capturing localized flow features, especially around re-entrant corners and high-curvature regions of the internal obstacles. Furthermore, the hybrid architecture using MLP encoders and KAN decoders enhanced convergence stability and reduced training epochs, showcasing the benefit of combining geometric feature extraction with physics-informed functional adaptability.

Another notable application of PIKAN-based architectures in fluid modeling is Anant-Net, introduced by Menon \& Jagtap~\cite{menon2025}, which integrates MLPs with KANs to solve high-dimensional nonlinear PDEs efficiently. As a benchmark task, the authors applied Anant-Net to the high-dimensional Poisson equation, a canonical elliptic PDE widely used to evaluate the scalability and numerical stability of surrogate models. The model was tested on domains up to 100 dimensions with manufactured solutions, allowing for quantitative error evaluation. Fig.~\ref{fig:fluid mechanics}(b) presents the absolute point-wise testing error computed over randomly selected 3D subspaces from the 21D, 99D, and 300D Poisson equation benchmarks. The results demonstrate that Anant-Net maintains low prediction error in the active dimensions, even as the ambient dimensionality increases, showing the scalability and robustness of the proposed architecture in high-dimensional settings.

In another benchmark, Toscano et al.~\cite{toscano2025aivt} proposed an artificial intelligence velocimetry-thermometry (AIVT) framework based on PIKANs, designed to reconstruct full 3D flow fields in turbulent Rayleigh–Bénard convection from sparse experimental measurements. The AIVT model was trained using sparse velocity measurements obtained via Lagrangian particle tracking, combined with partial temperature data from particle image thermometry. By incorporating the governing equations of momentum and energy, as well as boundary constraints, into the loss function, AIVT inferred smooth, instantaneous flow fields consistent with physical laws. Fig.~\ref{fig:fluid mechanics}(c) illustrates that the framework accurately reconstructs temperature fluctuation, velocity magnitude, viscosity dissipation, and temperature dissipation rate across space and time, capturing key features such as plume detachment, vorticity intensification, and dissipation zones. Compared to raw measurements, the reconstructed fields exhibited improved spatial coherence and better gradient resolution, particularly in regions lacking direct measurements. The use of KANs enabled localized functional adaptation, mitigating the spectral bias and enhancing the model’s ability to resolve fine-scale turbulent structures.

Expanding the scope of fluid dynamics applications, Gao \& Karniadakis~\cite{gao2025scalable} introduced a scalable gradient-free PIKAN framework tailored to infer spatially varying diffusion coefficients in elliptic PDEs under sparse and noisy observations. The authors introduced a scaled variant of the method, referred to as SDTEKI, which improves numerical stability and convergence, particularly under data scarcity and high-noise conditions. As a representative benchmark, the model was applied to a two-dimensional steady-state diffusion problem, where sparse and noisy pressure data served as training observations for inferring the underlying diffusion field. The results, drawn from the 1D diffusion equation benchmark, indicate that while all methods successfully recover the true parameter range, SDTEKI yields a more concentrated and stable posterior distribution compared to Hamiltonian Monte Carlo (HMC) and DTEKI, demonstrating improved robustness and inference efficiency under sparse and noisy data conditions. Fig.~\ref{fig:fluid mechanics}(d) presents histograms of the posterior estimates for the scalar diffusion parameter \( D \) obtained using HMC, standard DTEKI, and SDTEKI.

\subsection{Comparison with PINNs}

In this section, we aim to systematically compare the PIKAN with the widely used PINN framework. This comparison is structured around three key aspects that are essential for evaluating any physics-informed model: accuracy, convergence rate, and spectral behavior. 
This discussion is intended not only to present numerical improvements but also to explain the underlying reasons for these improvements, offering both intuition and evidence for when and why PIKAN may be a more suitable modeling choice.

\subsubsection{Accuracy Comparison}

\begin{table}[h]
\centering
\small
\renewcommand{\arraystretch}{1.05}
\caption{Comparison of various physics-informed models based on basis functions, optimization strategies, model complexity (number of trainable parameters $|\boldsymbol{\theta}|$), and relative $\mathcal{L}^2$ error in solving the Allen–Cahn equation. Results are reported across multiple studies with different spatial-temporal domains: $x \in [-1, 1],\ t \in [0, 1]$ for~\cite{shukla2024comprehensive, toscano2025kkans}, and $x \in [-6, 6],\ t \in [0, 1]$ for~\cite{mostajeran2025scaled}}
\label{Tab:ComparisonPIKAN}
\footnotesize
\begin{tabular}{p{2.5cm} p{2.5cm} p{3.5cm} p{2.5cm} p{2.5cm} p{1cm}}
\toprule  
Method & Basis & Optimization  & $|\boldsymbol{\theta}|$  & Rel. $\mathcal{L}^2$ Error & Ref \\ 
\midrule
PIKAN & B-spline & Adam  & $6721$  & $5.83 \times 10^{-1}$ & \cite{shukla2024comprehensive} \\ 
PIKAN & Chebyshev & Adam  & $6720$  & $5.15 \times 10^{-2}$&  \\ 
PIKAN+RBA & Chebyshev & Adam & $6720$ & $5.65 \times 10^{-2}$ &  \\ 
PINN+RBA & tanh & Adam & $50049$  & $1.51 \times 10^{-2}$&  \\
\addlinespace
\addlinespace
Scaled-PIKAN & Chebyshev & Adam & $7560$ & $5.90 \times 10^{-2}$ &  \cite{mostajeran2025scaled} \\
Scaled-PINN & tanh & Adam  & $7850$  & $5.10 \times 10^{-1}$ & \\
PIKAN & Chebyshev & Adam  & $7560$ & $6.00 \times 10^{-1}$ &   \\
PINN & tanh & Adam  & $7850$ & $5.10 \times 10^{-1}$ & \\
\addlinespace
\addlinespace
PIKAN & K\.{u}rkov\'a (KKAN) & Adam & $19572$  & $4.41 \times 10^{-3}$ & \cite{toscano2025kkans} \\ 
PIKAN+ssRBA & K\.{u}rkov\'a (KKAN) & Adam & $118491$   & $3.07 \times 10^{-5}$ &  \\ 
PIKAN & Chebyshev & Adam  & $20361$  & $3.55 \times 10^{-3}$ &  \\ 
PIKAN+ssRBA & Chebyshev &  Adam & $104136$ &   $2.91 \times 10^{-4}$ &  \\ 
PINN & tanh & Adam  & $21318$  & $1.36 \times 10^{-2}$ &  \\ 
PINN+ssRBA & tanh &  Adam & $91521$ &   $3.52 \times 10^{-5}$ &  \\ 
\addlinespace
\addlinespace
PIKAN &  B-spline & Adam  + BFGS  & $5540$  & $1.17 \times 10^{-2}$ &  \cite{kiyani2025optimizer} \\
PIKAN &  B-spline & Adam + SSBroyden  & $5540$  & $4.52 \times 10^{-4}$ &   \\
PIKAN &   Chebyshev &Adam + BFGS & $1368$  & $1.23 \times 10^{-3}$ &   \\
PIKAN &  Chebyshev & Adam  + SSBroyden  & $1368$  & $9.01 \times 10^{-6}$ &   \\
PINN &  tanh & Adam + BFGS & $2019$  & $7.59 \times 10^{-4}$ &   \\
PINN &  tanh &Adam  + SSBroyden & $2019$  & $1.15 \times 10^{-6}$ &   \\
PINN &  tanh & ssBroyden  & $2019$& $1.28 \times 10^{-6}$ &   \\ 
\bottomrule
\end{tabular}
\vspace{-10pt}
\end{table}

Evaluating the accuracy of physics-informed models is essential for understanding their ability to approximate complex solutions to PDEs. In the case of PIKAN, a key advantage lies in its use of basis functions, which can offer higher accuracy with fewer trainable parameters compared to vanilla PINNs. Table~\ref{Tab:ComparisonPIKAN} presents a non-exhaustive overview of studies comparing PIKAN and PINN across various configurations, including different basis types, optimization strategies, and problem settings for solving the Allen-Cahn equation as an example problem. The effect of hyperparameter selection, such as polynomial degree, network configuration, and the number of training samples, on the accuracy of the PIKAN model is also discussed. These comparisons not only highlight the superior accuracy of PIKANs in many cases but also demonstrate how architectural and algorithmic choices influence its effectiveness.

For instance, the choice of optimization strategy plays a critical role in determining the accuracy of physics-informed models. In the work of Kiyani et al.~\cite{kiyani2025optimizer}, the performance of different optimizers was systematically evaluated for both PIKAN and PINN architectures. The results show that combining the Adam optimizer with SSBroyden (Self-Scaled Broyden) significantly improves the accuracy of PIKAN. This improvement is primarily due to the self-scaling property of SSBroyden, which enables the optimizer to adapt to the local geometry of the loss landscape, thereby enhancing stability and robustness, especially for ill-conditioned problems such as nonlinear PDEs.
Specifically, as shown in Table~\ref{Tab:ComparisonPIKAN}, Chebyshev-based PIKAN with hybrid Adam/SSBroyden achieved a 99.3\% error reduction compared to the same architecture optimized with hybrid Adam/BFGS. 
However, achieving this level of accuracy in PINNs, on the order of $10^{-6}$ relative error, required an even greater number of trainable parameters than in Chebyshev-based PIKAN.
These findings confirm that PIKAN architectures are more capable of achieving high-accuracy solutions with fewer parameters, particularly when combined with suitable basis functions such as Chebyshev and advanced optimizers like SSBroyden, making them a more practical and efficient choice for solving complex PDEs, such as the Allen–Cahn equation.

\begin{figure}
    \centering
    \includegraphics[width=1.0\linewidth]{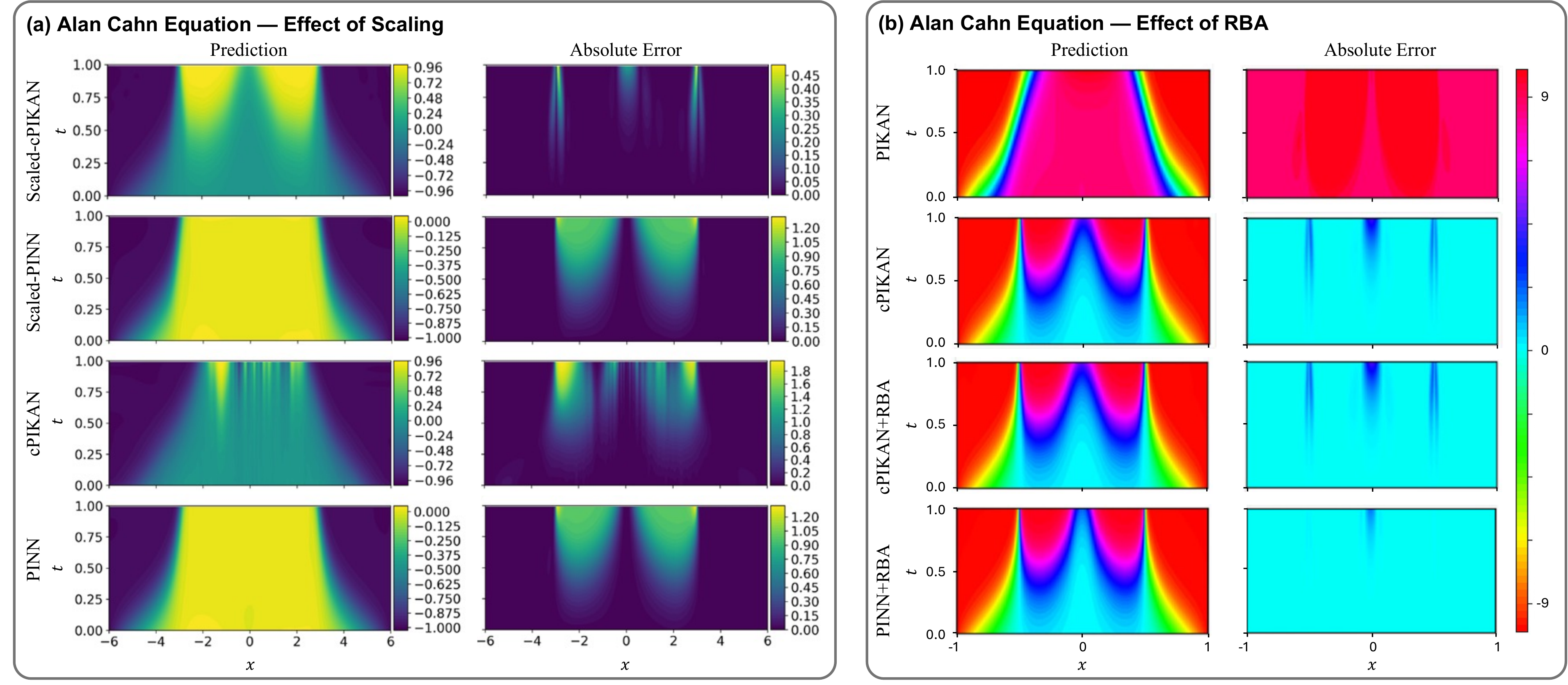}
    \caption{Comparison of predicted solutions for the Allen–Cahn equation with diffusion coefficient $D = 1 \times 10^{-4}$. \textbf{(a)} Solution over the spatio-temporal domain $[-6, 6] \times [0, 1]$. The model was trained using 50,000 residual points, with architectures containing 7,560 CKAN parameters and an MLP with 7,850 parameters. This panel is adapted from~\cite{mostajeran2025scaled}. \textbf{(b)} Solution over the spatio-temporal domain $[-1, 1] \times [0, 1]$. The network used 50000 residual points, with 6720 KAN parameters and an MLP with 50049 parameters. This panel is adopted from~\cite{shukla2024comprehensive}.
}
    \label{fig:Comparison_AC}
\end{figure}

A major challenge for achieving high accuracy in both PIKANs and PINNs arises when these models are applied to large or wide spatial domains, where numerical instability often occur. To address these challenges, researchers have proposed several strategies, including domain decomposition into overlapping subdomains~\cite{howard2024finite} and regularization techniques such as dropout Tikhonov ensemble Kalman inversion~\cite{gao2025scalable}. A particularly effective and conceptually simple alternative is the application of scaling strategies to both the input domain and the PDE residual, mapping them onto a range more compatible with the basis functions, such as the natural interval $[-1, 1]$ for Chebyshev polynomials. Mostajeran \& Faroughi~\cite{mostajeran2025scaled} implemented such a method by scaling both the input coordinates and residual terms in the loss function, resulting in the Scaled-cPIKAN model. This type of scaling helps normalize gradients and stabilize the optimization process, which is especially beneficial in wide spatial domains. They applied this approach to solve the Allen–Cahn equation over the extended domain $[-6, 6]$, which is significantly more challenging than the $[-1, 1]$ interval. As shown in Table~\ref{Tab:ComparisonPIKAN}, under identical training conditions, Scaled-cPIKAN achieved an 88\% reduction in relative error compared to Scaled-PINN and a 90\% improvement over the vanilla cPIKAN. Fig.~\ref{fig:Comparison_AC}(a) illustrates the performance of Scaled-cPIKAN and Scaled-PINN in predicting the Allen–Cahn solution. These results demonstrate that domain and residual scaling not only improves learning efficiency but also enables high accuracy over large spatial domains without increasing model complexity or relying on auxiliary techniques.

Incorporating structured basis functions and architectural enhancements has significantly improved the accuracy of PIKANs. Among the most common basis functions, B-splines have traditionally been used due to their local support and smoothness properties. However, other studies show that structured basis functions such as Chebyshev polynomials~\cite{guo2024physics, mostajeran2025scaled}, wavelets~\cite{patra2024physics, mostajeran2023novel}, and higher-order ReLU~\cite{so2024higher} can provide significantly better approximation accuracy than B-splines.
Beyond these classical bases, more advanced hybrid kernel-based architectures have been proposed to enhance model accuracy. For example, in~\cite{toscano2025kkans}, K\.{u}rkov\'a-based KANs (KKANs) replace the MLP layers with structured basis expansions, combining the nonlinear expressiveness of neural networks with the robustness of linear approximation techniques. This design leads to substantial improvements in accuracy without requiring additional techniques such as attention mechanisms. As shown in Table~\ref{Tab:ComparisonPIKAN}, KKAN achieves approximately 67\% greater accuracy than the standard PINN, while using only about 8\% fewer parameters. Likewise, the Chebyshev-based PIKAN model improves accuracy by roughly 74\% compared to PINN, with a comparable parameter count. KKAN demonstrates an accuracy improvement of over 99\% compared to the B-spline-based PIKAN, albeit with a parameter count nearly three times higher.
These findings highlight the limitations of B-spline bases in complex domains and emphasize the advantages of using structured expansions like Chebyshev polynomials and hybrid architectures such as KKAN. By combining structured linear components with globally expressive basis functions, these models achieve significantly higher accuracy than traditional PINNs, often with comparable or fewer parameters. These results demonstrate the critical role of selecting appropriate basis functions and model architectures in developing accurate and efficient physics-informed neural networks.

Attention mechanisms like Residual-Based Attention (RBA) and its self-scaled variant (ssRBA) can also improve the accuracy of physics-informed models when applied appropriately. RBA works by adaptively weighting regions of the domain using the exponentially weighted moving average of point-wise residuals~\cite{anagnostopoulos2024residual}, allowing the model to focus more on areas with larger errors. This attention-like effect enhances accuracy with little computational overhead~\cite{toscano2024inferring,rigas2024adaptive,chen2024self}. The self-scaled version, ssRBA, further improves training stability by dynamically adjusting its internal scaling to preserve a high signal-to-noise ratio during learning~\cite{mcclenny2023self}. Shukla et al.~\cite{shukla2024comprehensive} evaluated RBA in both PINN and PIKAN models. Their findings show that switching from a B-spline to a Chebyshev basis in PIKAN reduces error by over 90\% (from $5.83 \times 10^{-1}$ to $5.15 \times 10^{-2}$) without changing the number of parameters. However, adding RBA to this configuration did not lead to further improvement, indicating that Chebyshev-based PIKAN already achieves high accuracy without requiring attention mechanisms. On the other hand, the PINN model required RBA to achieve a similar level of accuracy, but this improvement came at the cost of using significantly more parameters. The RBA-enhanced PINN used 50049 parameters, which is approximately 645\% more than the 6720 parameters used in the Chebyshev-based PIKAN. Fig.~\ref{fig:Comparison_AC}(b) compares the accuracy of PIKAN and PINN, along with their RBA-invariant counterparts, in predicting the solution of the Allen–Cahn equation over the domain $x \in [-1, 1]$ and $t \in [0, 1]$. In more advanced architectures like KKAN and Chebyshev-based PIKAN~\cite{toscano2025kkans}, incorporating ssRBA led to very high accuracy gains, up to 99\% and 90\% error reduction, respectively, but these benefits were achieved by increasing the model size considerably. These results show that PIKAN, especially when using Chebyshev basis functions, can achieve high accuracy with fewer parameters and without relying on attention mechanisms. In comparison, traditional PINNs depend more heavily on RBA or ssRBA to reach similar accuracy, which often requires much larger models.

\begin{figure}
    \centering
    \includegraphics[width=1\linewidth]{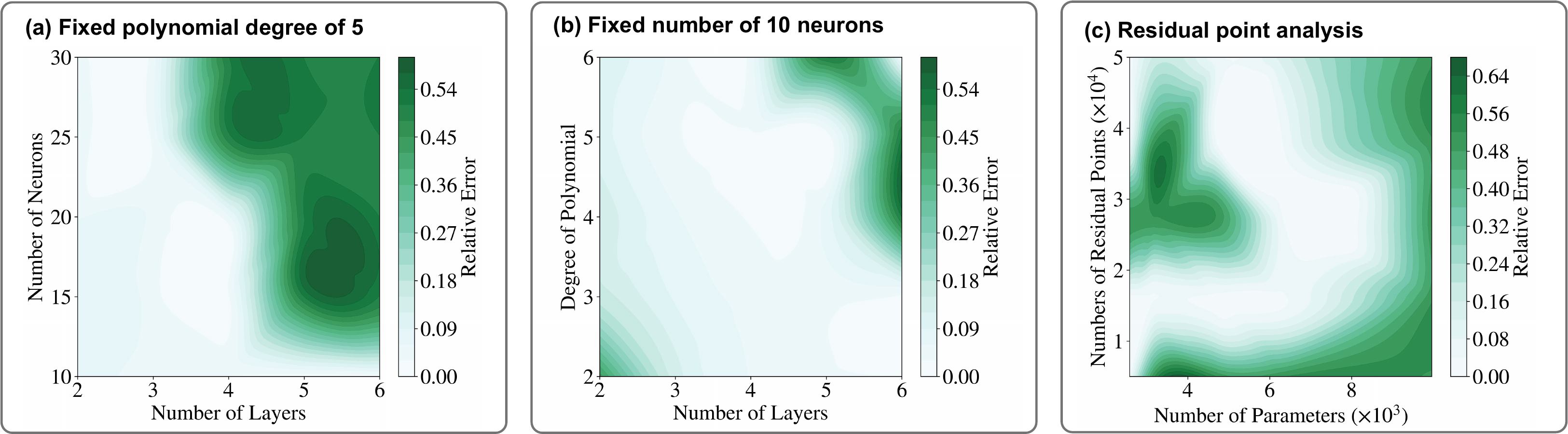}
    \caption{Influence of hyperparameter selection on the relative error of the PIKAN model for the Allen–Cahn example.
\textbf{(a)} Layer and neuron configuration. The minimum relative error \((10^{-2})\) occurs in the fifth layer with 10 neurons. 
\textbf{(b)} Degree and layer configuration. The minimum relative error \((10^{-2})\) is achieved in the fifth layer with a polynomial degree of 5. 
\textbf{(c)} Number of parameters and residual points configuration. The minimum relative error \((10^{-2})\) is obtained with 2,580 parameters and 50,000 residual points.
All panels present new analysis based on \cite{mostajeran2025scaled}.
}
    \label{fig:hyperparameter}
\end{figure}

Hyperparameter configuration plays a key role in determining both the accuracy and generalization performance of KAN-based models. Despite its importance, only a few studies have systematically examined how these factors influence learning behavior, including~\cite{zeng2024kan, pourkamali2024kolmogorov}. For example,~\cite{pourkamali2024kolmogorov} presented an empirical investigation within a data-driven framework, analyzing the sensitivity of KANs to key hyperparameters such as network depth and spline order. Their findings revealed that, unlike MLPs whose accuracy remained relatively stable with increasing depth, KAN performance tended to degrade as additional layers were introduced. Moreover, the polynomial order of the B-spline activation functions was identified as a crucial factor: higher spline orders generally led to reduced accuracy across real-world datasets. 
To further illustrate the practical implications of hyperparameter sensitivity, a quantitative analysis is provided for the Allen–Cahn example using the cPIKAN model.
Fig.~\ref{fig:hyperparameter} provides a standardized illustration of how variations in key hyperparameters, such as network configuration and the number of residual points, affect the relative $\mathcal{L}^2$ error in the PIKAN model. This example demonstrates that accuracy is sensitive to architectural and sampling choices, highlighting the needs  for  incorporating standardized datasets, metrics, and tuning procedures across studies to ensure comparability, reproducibility, and meaningful benchmarking within the KAN research community.

Overall, the comparative studies consistently demonstrate that  PIKAN outperforms PINN in terms of accuracy across a wide range of settings. Whether using standard domains or extended spatial intervals, and regardless of the optimizer or training scheme, PIKAN models achieve lower relative errors, often by margins exceeding  85–99\%  compared to equivalent PINN configurations. Importantly, PIKAN reaches this level of accuracy with significantly fewer trainable parameters, highlighting its architectural efficiency. While techniques like RBA and ssRBA can help improve PINN accuracy, they require much larger models and more complex training setups to match the accuracy of PIKAN. Furthermore, the use of Chebyshev basis functions and advanced optimizers such as SSBroyden consistently enhances PIKAN’s accuracy, whereas PINN relies heavily on these additions to stay competitive. Taken together, these findings confirm that  PIKAN offers a more accurate and resource-efficient framework for solving PDEs, making it a highly practical alternative to vanilla PINN architectures.

\subsubsection{Convergence Analysis}

\begin{figure}
    \centering
    \includegraphics[width=1.0\linewidth]{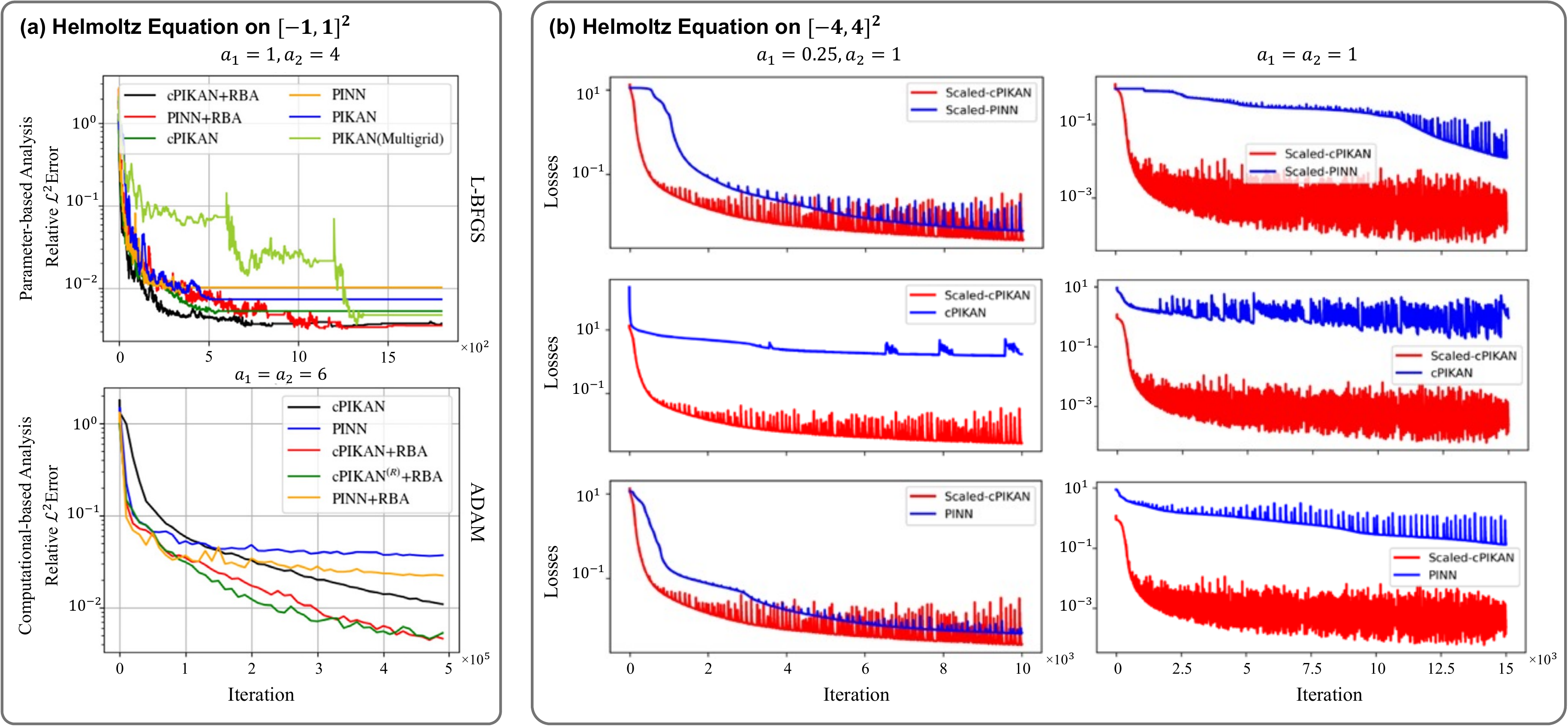}
    \caption{Convergence behavior of PIKAN and PINN for solving the Helmholtz equation. \textbf{(a)} Performance comparison of PINN, PIKAN, and cPIKAN variants with and without RBA, trained using L-BFGS and ADAM optimizers on the domain $[-1,1]^2$. This panel is adopted from~\cite{shukla2024comprehensive}. \textbf{(b)} Training loss evolution of scaled-cPIKAN compared to PINN and unscaled models across different frequency settings on the extended domain $[-4,4]^2$. This panel is adopted from~\cite{mostajeran2025scaled}.}
    \label{Fig:ConvergenceBehavior}
\end{figure}

Understanding the convergence behavior of physics-informed neural models is essential for evaluating their training efficiency and overall performance. Convergence typically refers to how quickly and consistently a model reduces its prediction error during training, and it can be assessed through three key aspects: the rate of loss decay, the stability of the optimization process, and the occurrence of early flattening in the loss curve. In this section, we provide a comparative analysis of convergence characteristics between PINN and PIKAN-based methods, using benchmark results from the Helmholtz equation with different wave numbers. The convergence behavior is summarized in Fig.~\ref{Fig:ConvergenceBehavior}, which displays relative error and training loss curves under various model configurations and solvers.

To evaluate convergence behavior, we focus on the loss decay rate, which reflects how quickly the training error decreases over iterations. A faster decay indicates more efficient learning and a more accurate approximation of the solution. Fig.~\ref{Fig:ConvergenceBehavior}(a), adapted from~\cite{shukla2024comprehensive}, shows the relative $\mathcal{L}^2$ error plotted over training iterations and compares the convergence behavior of PIKAN and PINN under both BFGS and Adam optimizers in both parameter-based and computation-based analyses. While PIKAN typically converges more slowly than vanilla PINN, the integration of the RBA strategy substantially improves convergence speed in both architectures. Notably, PINN+RBA and cPIKAN+RBA exhibit the steepest error decay, highlighting the effectiveness of adaptive weighting in accelerating the training process. Fig.~\ref{Fig:ConvergenceBehavior}(b), also adapted from~\cite{mostajeran2025scaled}, presents loss curves over training iterations to analyze the effects of scaling and recurrence. Scaled-cPIKAN demonstrates the most substantial gains, reducing the loss much faster than its unscaled counterparts. As the wave number increases and the problem complexity grows, Scaled-cPIKAN consistently maintains its superior decay rate, significantly outperforming both vanilla PINN and cPIKAN.

Stability in convergence refers to a model’s ability to maintain consistent progress during training without abrupt oscillations, spikes in error, or divergence. Stable convergence is particularly critical when solving PDEs like the Helmholtz equation, where oscillatory solutions and complex gradients can amplify numerical instability. In Fig.~\ref{Fig:ConvergenceBehavior}(a), PIKAN, though slower in raw convergence speed, demonstrates smoother and more consistent error curves compared to vanilla PINN, especially under BFGS. The integration of RBA weighting further enhances stability in both models. PINN+RBA reduces the fluctuations present in standard PINN, while cPIKAN+RBA maintains a steady descent throughout training, with minimal noise in the loss trajectory. Moreover, adding recurrence (as in cPIKAN$^{(R)}$) provides marginal improvements by dampening early-stage oscillations. In Fig.~\ref{Fig:ConvergenceBehavior}(b), Scaled-cPIKAN, in particular, exhibits stable convergence patterns across both low and high wave number regimes. As the wave number increases, the optimization landscape becomes more rugged, often destabilizing standard models, while Scaled-cPIKAN maintains smooth loss curves, on average.

Early flattening refers to the point where the training loss quickly reaches a plateau, and further iterations no longer lead to meaningful improvements. This phenomenon is important to monitor, as it often signals optimization challenges such as poor gradient flow, limited model capacity, or inflexible learning dynamics. As shown in Fig.~\ref{Fig:ConvergenceBehavior}(a), PINN consistently exhibits early flattening under both Adam and BFGS optimizers. In contrast, PIKAN maintains a more gradual reduction in error, especially under BFGS, demonstrating a better ability to avoid early stagnation. The addition of RBA effectively addresses this issue: both PINN+RBA and cPIKAN+RBA avoid early flattening entirely, showing sustained and prolonged convergence. While the recurrent version, cPIKAN$^{(R)}$, slightly delays the onset of flattening compared to its standard counterpart, the improvement is modest. In Fig.~\ref{Fig:ConvergenceBehavior}(b), Scaled-cPIKAN shows strong robustness by maintaining a steady decrease in loss throughout training. The use of scaling significantly reduces early flattening, with Scaled-cPIKAN continuing to converge effectively even as the problem complexity increases.

The convergence characteristics of PINN and PIKAN models show that convergence efficiency is best evaluated using three interrelated metrics: the loss decay rate, training stability, and the presence of early flattening. Comparative analyses indicate that PIKAN models generally offer improvements in at least one of these aspects compared to standard PINNs. While PIKAN typically exhibits smoother and more stable convergence curves, its decay rate can be further improved by incorporating RBA. Recurrence, when applied to Chebyshev-based PIKAN architectures, provides moderate benefits by reducing early-stage instability. In contrast, scaling techniques consistently lead to faster and more sustained convergence, and are especially effective in suppressing early flattening, particularly in high-frequency problem settings. These findings highlight the critical role of architectural choices and adaptive strategies in achieving fast, stable, and robust convergence in physics-informed neural models.

\subsubsection{Spectral Behavior}

\begin{figure} [h]
    \centering
    \includegraphics[width=1.0\linewidth]{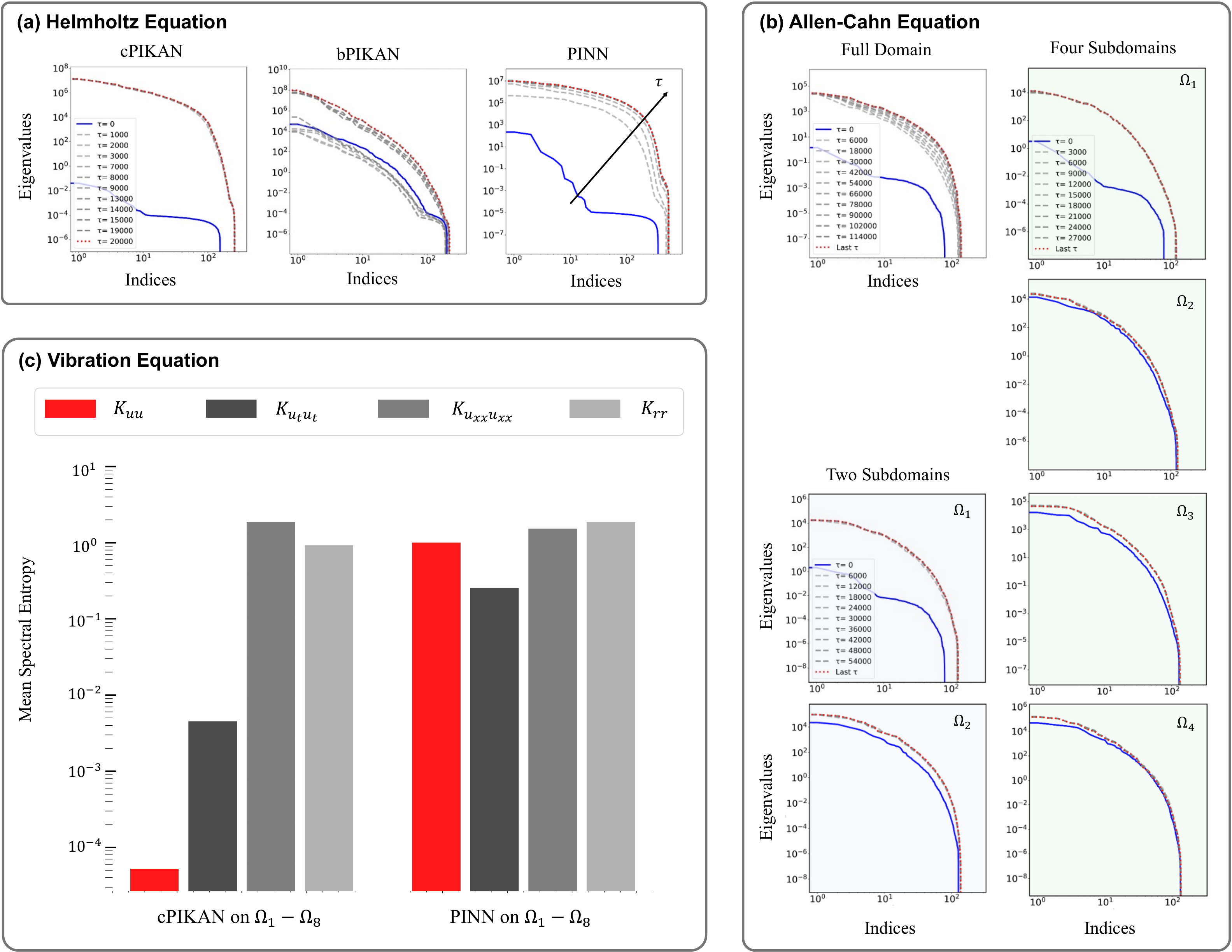}
    \caption{Comparison of spectral behavior of PIKAN and PINN during training. \textbf{(a)} Evolution of the NTK eigenvalue spectra for different models. cPIKAN shows stable and converging spectra, reflecting well-conditioned training, while PINN and bPIKAN exhibit delayed or unstable spectral dynamics linked to slower or less reliable convergence. \textbf{(b)} Spectral evolution of cPIKAN with varying numbers of temporal subdomains. More subdomains lead to faster spectral convergence and improved conditioning, indicating more efficient and stable training. \textbf{(c)} Mean spectral entropy comparison between cPIKAN and PINN using 8 temporal subdomains. cPIKAN yields consistently lower spectral entropy, suggesting reduced kernel complexity and improved representation of solution features. 
    All panels are adopted from~\cite{mostajeran2025scaled}.
    }
    \label{fig:CompPIKANNTK}
\end{figure}

While the previous sections provided a comparative analysis of PIKAN and PINN regarding accuracy and convergence behavior, this section focuses on a different but equally important aspect of model performance: the spectral characteristics of their training dynamics. Understanding these spectral properties is crucial for explaining the observed convergence trends and for improving the theoretical foundations of physics-informed models. The Neural Tangent Kernel~\cite{jacot2018neural, seleznova2022analyzing, seleznova2022neural} is a particularly valuable tool in this context. Prior work has highlighted the sensitivity of NTK dynamics to network structure and training regimes, especially in the context of PINNs~\cite{Wang2020NTKPinns, saadat2022neural}. 
However, these analyses often rely solely on eigenvalue distributions, which may not capture the full complexity of NTK behavior in KAN models. To address this limitation, Faroughi \& Mostajeran~\cite{faroughi2025neural} have introduced more informative spectral metrics, spectral entropy, a concept derived from Shannon’s information theory~\cite{shannon1948mathematical}. Spectral entropy quantifies the uncertainty or dispersion in the NTK eigenvalue spectrum and is defined as,
\begin{equation}\label{Eq.shanon}
    \text{Spectral Entropy} = -\sum_i \frac{|\lambda_i|}{\sum_j |\lambda_j|} \log \left( \frac{|\lambda_i|}{\sum_j |\lambda_j|} \right),
\end{equation}
where $\lambda_i$ are the eigenvalues of the NTK matrix. This metric serves as a proxy for how learning signals are distributed, offering insights into the alignment between model structure and the frequency content of the solution. Fig.~\ref{fig:CompPIKANNTK} summarizes the spectral behavior of the NTK matrix under various settings, complementing empirical observations and providing a theoretical foundation for understanding when and why PIKAN outperforms traditional PINNs.
The architectures operate in the finite-width regime, as the parameter initialization follows standard scaling, and the NTK evolves during training, capturing feature-learning effects characteristic of practical networks.

Analyzing the eigenvalue spectrum of the NTK matrices during training provides critical insight into the learning dynamics and numerical stability of physics-informed based methods. In foundational work, the NTK of PINNs was derived under the infinite-width assumption~\cite{Wang2020NTKPinns}, revealing that the kernel converges to a nearly constant form during training. However, the eigenvalue distribution showed a pronounced spectral bias, causing imbalanced convergence across different components of the loss. This limitation was further explored in~\cite{saadat2022neural}, where PINNs were shown to struggle with learning initial and boundary conditions in advection- or diffusion-dominated regimes, prompting the use of adaptive loss weighting and periodic activations. More recently, NTK analysis has been extended to Kolmogorov–Arnold-type networks. A simplified study~\cite{mostajeran2025scaled} examined the effect of Chebyshev bases on the NTK spectrum using non-nested approximations, while a more comprehensive approach in~\cite{faroughi2025neural} derived the full NTK for cPIKANs, incorporating nested structures and physics-informed constraints to reflect realistic training setups. Fig.~\ref{fig:CompPIKANNTK}(a) presents the evolution of NTK eigenvalues during training for the Helmholtz equation, under a setup that includes domain and residual scaling. The cPIKAN model exhibits stable and smoothly converging eigenvalue spectra across training iterations, indicating consistent kernel behavior and well-conditioned learning dynamics. In contrast, the bPIKAN model displays highly irregular and disordered spectra, with no clear convergence pattern throughout training. This chaotic spectral behavior supports previous findings that bPIKAN fails to accurately learn the Helmholtz solution under the same configuration. For the PINN model, the NTK eigenvalues remain unstable during early training and only begin to stabilize after several thousand iterations. Convergence in the spectrum becomes apparent mainly in the later stages of training, suggesting delayed and uneven learning across different modes. These observations underscore the importance of both the chosen basis and architectural design in shaping NTK dynamics and the general learning behavior during training, particularly for stiff or oscillatory PDEs such as the Helmholtz equation.

An important strategy for improving the spectral behavior and training performance of PINN and PIKAN models is domain decomposition, a technique that partitions the input space, e.g., time or space-time domain, into smaller subdomains to simplify the learning task. Previous studies have highlighted the impact of this approach on spectral bias and convergence dynamics. In particular, Saadat et al.~\cite{saadat2022neural} showed that restricting the training domain to a smaller subregion significantly improved the performance of PINNs. Their findings confirmed that spectral bias, caused by the imbalance in learning different solution modes, can be mitigated by reducing the complexity of the target function. This principle underlies several sequential learning strategies, such as time-marching or sequence-to-sequence PINNs, where the model is trained incrementally over smaller time windows. However, such methods come with added algorithmic and computational overhead. More recently, Faroughi \& Mostajeran~\cite{faroughi2025neural} conducted a detailed NTK analysis of cPIKAN models under temporal domain decomposition. Their results demonstrated that as the number of subdomains increases, the NTK matrices become better conditioned and converge more rapidly, leading to more stable learning and improved accuracy. This effect was especially pronounced in diffusion-dominated problems like the Allen–Cahn equation. As illustrated in Fig.~\ref{fig:CompPIKANNTK}(b), splitting the time domain into 1, 2, or 4 subdomains results in progressively smoother NTK eigenvalue spectra, confirming the benefits of decomposition in reducing spectral complexity. These findings suggest that domain decomposition is a powerful yet computationally intensive tool for addressing spectral challenges in physics-informed models.

Given the increasing complexity of NTK matrices in advanced physics-informed models, there is a growing need for such advanced spectral metrics that go beyond conventional eigenvalue analysis. Traditional approaches may fail to capture key aspects of training behavior, especially in models involving higher-order derivatives or multi-component loss functions. Wang et al.~\cite{Wang2020NTKPinns} analyzed the failure of standard PINNs in solving a 1D wave equation and found a severe imbalance in the convergence rates of different loss components, leading to instability during training. By tracking the dynamic evolution of NTK-weighted loss terms, they demonstrated that customized optimization strategies could restore balance and ensure stable learning. In a more challenging case, Faroughi \& Mostajeran~\cite{faroughi2025neural} studied the forced vibration equation, which includes higher-order terms such as $u_{xxxx}$ and $u_{tt}$, along with complex initial and boundary conditions. Instead of relying solely on NTK eigenvalue trajectories, they used spectral entropy, Eq.~\eqref{Eq.shanon}, to assess how concentrated or dispersed the NTK spectrum is during training. As shown in Fig.~\ref{fig:CompPIKANNTK}(c), both PINN and cPIKAN are applied to this problem using 8 temporal subdomains. The figure reports the mean spectral entropy of NTK blocks like $K_{uu}$ (solution interactions) and $K_{u_t u_t}$ (time derivative interactions), showing that cPIKAN consistently achieves lower entropy, indicating better alignment with high-frequency solution features. These results underscore the value of spectral entropy as a diagnostic tool for revealing the deeper dynamics of learning in complex physics-informed models.

\section{Deep-operator KANs}

This section introduces Deep-operator Kolmogorov–Arnold Networks (DeepOKANs), an emerging class of KANs paradigm designed to model entire operators, rather than pointwise input-output relationships. Instead of processing a finite-dimensional input vector, these architectures map an input function (e.g., boundary or initial conditions) to an output function (e.g., a PDE solution), thereby addressing the repeated-simulation bottleneck in parametric PDE settings, by learning a single surrogate that applies to \emph{any} admissible input condition within a prescribed family.
The section is organized into three parts.  
The first part describes the architecture and training process of DeepOKAN, focusing on its branch–trunk structure and the use of KANs to represent operators. It also explains how operator-level decomposition enables integration of physics-based constraints across the solution domain.
The second part presents applications in areas such as solid mechanics, heat transfer, and electromagnetics, highlighting the model’s ability to approximate solution fields under varying conditions. The third part provides a comparative analysis between DeepOKAN and other operator learning methods, focusing on predictive accuracy and convergence behavior.

\begin{figure}[h!]
    \centering
\includegraphics[width=.9\linewidth]{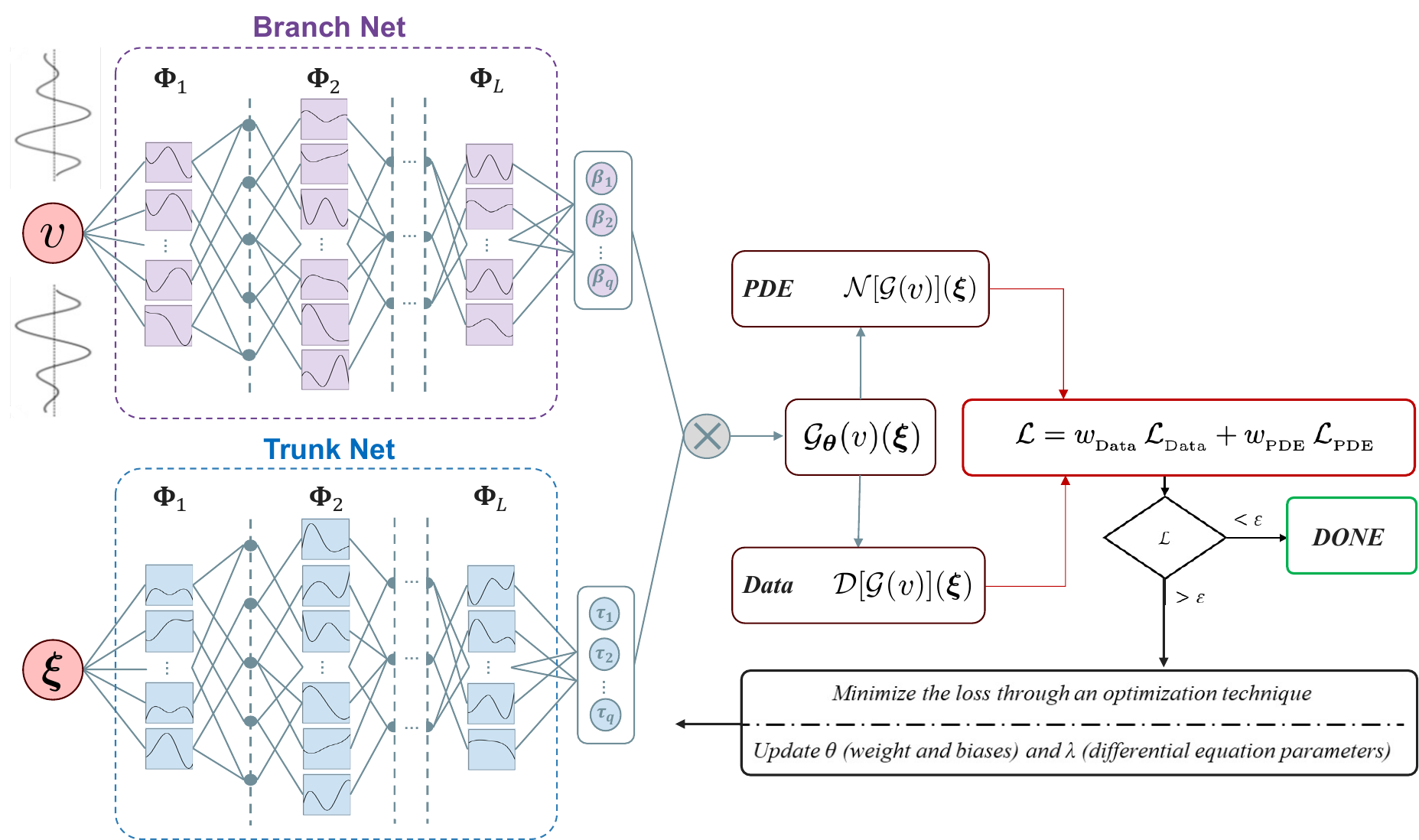}
    \caption{A schematic architecture of DeepOKANs composed of a branch network that encodes samples of the input functions, extracting their latent representations, and a trunk network that extracts latent representations of spatiotemporal points in the solution domain at which the output functions are evaluated. In this setup, a dot product is then used to merge the latent representations extracted by each subnetwork and obtain a continuous and differentiable representation of the output functions. In a pure data-driven format $\lambda =0$. In a physics-informed format, the data loss also contains the initial and boundary condition loss terms, and automatic differentiation is used to formulate appropriate terms in both data and PDE loss expressions.   }
    \label{fig:deeppokan}
\end{figure}

\subsection{Architecture and Training Strategies}

Similar to the DeepONet framework of Lu et al. (2021)~\cite{lu2021learning}, the DeepOKAN architecture is composed of two neural subnetworks 
as shown in Fig.~\ref{fig:deeppokan}: (i)~a \textit{branch network} that encodes the input function, represented via its evaluations at a set of fixed sensor locations $\boldsymbol{s}$, into a low-dimensional latent feature vector, and (ii)~a \textit{trunk network} that maps arbitrary spatiotemporal query points $\boldsymbol{\xi} = (\boldsymbol{x}, t)$ to a corresponding set of latent basis functions.
Unlike DeepONet, which typically employs MLPs in both subnetworks, DeepOKAN replaces them with  KANs. These perform learnable univariate transformations along each input dimension, offering greater expressivity and flexibility in capturing localized, non-smooth, or highly nonlinear behaviors. Such a structure helps alleviate spectral bias commonly observed in standard operator learning~\cite{abueidda2025deepokan,Wang2020NTKPinns}.
Formally, given an input function $v \in \mathcal{V}$, its discretized evaluation at sensor points is passed through the branch network with parameters $\boldsymbol{\theta}_b$, producing a feature vector $\boldsymbol{\beta}(v; \boldsymbol{\theta}_b) = [\beta_1, \beta_2, \dots, \beta_q]^\top \in \mathbb{R}^q$.
Simultaneously, for a query point $\boldsymbol{\xi} \in \Omega \times [0,T]$, the trunk network with parameters $\boldsymbol{\theta}_t$ yields $\boldsymbol{\tau}(\boldsymbol{\xi}; \boldsymbol{\theta}_t) = [\tau_1, \tau_2, \dots, \tau_q]^\top \in \mathbb{R}^q$.
The model prediction is obtained through an inner product,
\begin{equation}\label{Eq.OKAN1}
    \mathcal{G}_{\boldsymbol{\theta}}(v)(\boldsymbol{\xi}) = \sum_{k=1}^{q} \beta_k(v; \boldsymbol{\theta}_b) \cdot \tau_k(\boldsymbol{\xi}; \boldsymbol{\theta}_t),
\end{equation}
where $\boldsymbol{\theta}_b$ and $\boldsymbol{\theta}_t$ denote the parameter vectors of the branch and trunk networks, respectively, and the full parameter set is $\boldsymbol{\theta} = \{\boldsymbol{\theta}_b, \boldsymbol{\theta}_t\}$.
Equation~\eqref{Eq.OKAN1} defines a nonlinear operator  $\mathcal{G}_{\boldsymbol{\theta}}\: :\:  \mathcal{V} \to \mathcal{A}$ that maps input functions to output solution functions over the spatiotemporal domain.
To guide the training of DeepOKAN, a composite loss function is used that combines both data-driven and physics-based terms. Let $\{v^{(j)}\}_{j=1}^N$ denote a set of training input functions sampled from the input space $\mathcal{V}$, each associated with an output governed by the underlying PDE. 

For each training input function $v^{(j)}$, we define a set of output evaluation points $\{\boldsymbol{\xi}_i^d\}_{i=1}^{N_d} \subset \Omega \times [0,T]$, at which the model is expected to predict the corresponding output values. These points may vary across different training samples. In contrast, the input function $v^{(j)}$ is represented through its evaluations at a fixed set of sensor locations $\{\boldsymbol{s}_i\}_{i=1}^{m} \subset \Omega_s \subset \Omega$, where the values $v^{(j)}(\boldsymbol{s}_i)$ are known. This setup enables the model to learn a continuous mapping from input functions, sampled at fixed sensors, to output functions evaluated over flexible spatiotemporal locations.
 Similarly, we define residual points $\{\boldsymbol{\xi}_i^r\}_{i=1}^{N_r} \subset \Omega \times [0,T]$, where the PDE residuals are evaluated using the trained model predictions.
The total loss function is defined as,
\begin{equation}\label{Eq.lossDeepOKAN}
\mathcal{L} = w_{_\text{Data}} \, \mathcal{L}_{_\text{Data}} + w_{_\text{PDE}} \, \mathcal{L}_{_\text{PDE}},
\end{equation}
where $w_{_\text{Data}}$ and $w_{_\text{PDE}}$ are weighting coefficients for the data and PDE components of the loss, respectively.
In a purely data-driven setting, the PDE component is omitted by setting $w_{_\text{PDE}} = 0$. Conversely, in a physics-informed framework, the data loss includes contributions from both initial and boundary conditions. In this context, automatic differentiation is employed to compute the necessary derivatives for both the data and PDE loss terms, as discussed in Section~\ref{Sec:PIKANs}.
In Eq.~\eqref{Eq.lossDeepOKAN}, the data loss term ensures agreement between the predicted and ground-truth data,
\begin{equation}
\mathcal{L}_{_\text{Data}} = \frac{1}{N N_d} \sum_{j=1}^{N} \sum_{i=1}^{N_d} \left| \mathcal{G}_{\boldsymbol{\theta}}(v^{(j)})(\boldsymbol{\xi}_i^d) - v^{(j)}(\boldsymbol{s}_i) \right|^2,
\end{equation}
and the PDE loss term enforces adherence to the governing physical laws,
\begin{equation}\label{Eq.lossPDEokan}
\mathcal{L}_{_\text{PDE}} = \frac{1}{N N_r} \sum_{j=1}^{N} \sum_{i=1}^{N_r} \left| \mathcal{N}[\mathcal{G}_{\boldsymbol{\theta}}(v^{(j)})](\boldsymbol{\xi}_i^r) - f^{(j)}(\boldsymbol{\xi}_i^r) \right|^2.
\end{equation}

In Eq.~\eqref{Eq.lossPDEokan}, $\mathcal{N}[\cdot]$ is the differential operator associated with the underlying partial differential equation, and $f^{(j)}$ is the known forcing term for the $j$-th input function. This formulation enables DeepOKAN to learn a nonlinear operator that maps each input function $v^{(j)} \in \mathcal{V}$ to its corresponding solution over space and time.

Once trained, DeepOKAN generalizes across a wide class of input functions and provides continuous predictions over the spatiotemporal domain without requiring re-solving the PDE for new inputs. Its use of dimension-wise KAN layers reduces the complexity of learning high-dimensional operators and improves training stability, particularly in challenging physical domains such as fluid dynamics, wave propagation, and solid mechanics~\cite{abueidda2025deepokan}. Furthermore, normalization of inputs and dynamic loss balancing strategies can be employed to further enhance numerical stability and convergence during training.

To better address the challenges of operator learning, various architectural variants have been proposed by modifying the network components within the branch and trunk subnetworks. For example, Kiyani et al.~\cite{kiyani2025predicting} introduced a simplified architecture in which the trunk network is replaced by a KAN, trained without incorporating any physics-based loss. In contrast, Abueidda et al.~\cite{abueidda2025deepokan} presented DeepOKAN, a neural operator that utilizes Gaussian RBF KANs in both subnetworks, demonstrating enhanced performance and interpretability. Their formulation serves as a performance benchmark and opens avenues for exploring other KAN variants, such as multi-quadratic RBFs and B-splines. Wang et al.~\cite{wang2025efkan} proposed a hybrid architecture combining a FNO in the branch network with a KAN-based trunk, leveraging the complementary strengths of spectral and localized representations. Additionally, Yu et al.~\cite{yu2025deepoheat} developed DeepOHeat-v1, which integrates adaptive KANs with learnable activation functions in the trunk to capture complex multi-scale thermal patterns. This model also incorporates a separable training strategy and GMRES-based (Generalized Minimal Residual Method) refinement to improve computational efficiency and reliability. Expanding on the DeepONet formulation, Toscano et al.~\cite{toscano2025kkans} proposed DeepOKKAN, which incorporates KKANs characterized by MLP-based inner blocks and radial basis function outer blocks, offering a structured and interpretable architecture for operator learning. Collectively, these variants underscore the flexibility of KAN-based operator networks and their increasing significance in physics-informed machine learning frameworks.

\subsection{Applications}
\label{sec:deepokan_applications}

The development of DeepOKANs marks a significant advancement in extending KAN-based frameworks to operator learning tasks.  This operator-centric approach enables efficient approximation of solution fields across varying input conditions without the need for retraining, making DeepOKANs particularly attractive for rapid surrogate modeling and real-time prediction. A non-exhaustive overview of recent developments, along with the associated methods and their application domains, is presented in Table~\ref{Tab:DeepOKAN_Advances_Table}.

\begin{table}[h]
\centering
\small
\renewcommand{\arraystretch}{1.05}
\caption{A non-exhaustive list of recent studies on DeepOKANs, illustrating methodological innovations in network design and operator learning.
}
\label{Tab:DeepOKAN_Advances_Table}
\footnotesize
\begin{tabularx}{\textwidth}{>{\hsize=0.15\hsize}X >{\hsize=0.75\hsize}X >{\hsize=0.18\hsize}r}
\toprule  
\textbf{Method} & \textbf{Objective and Application Domain} & \textbf{Ref.} \\ 
\midrule
DeepOKAN &
Introduced neural operators to model intricate relationships between input parameters and output fields;
Used Gaussian RBFs for both branch and trunk networks;
Applied to 1D sinusoidal waves, 2D orthotropic elasticity, and the transient Poisson’s problem.
& \cite{abueidda2025deepokan} \\
\addlinespace
\addlinespace
 Hybrid Operators & 
Employed an MLP branch and Chebyshev KAN trunk; 
Trained concurrently;
Applied to predicting displacement and phase fields from crack size, boundary displacements in 1D bars, and notched structures under tensile/shear loading.
& \cite{kiyani2025predicting} \\
\addlinespace
\addlinespace
Operator KANs
& 
Analyzed the performance of KANs and MLPs in both shallow and deep network models; 
Used spline-based KANs;
Applied to Burgers’ equation, 1D and 2D Darcy Flow, and an Elastic Plate
problem.  
& \cite{pant2025mlps} \\
\addlinespace
\addlinespace
 EFKAN & 
 Employed FNO as a branch and KAN (B-spline, degree 3, grid 5) as a trunk network;
Mapped resistivity models to apparent resistivity and phase at target locations and frequencies;
Applied to MT forward modeling in the frequency domain.
 & \cite{wang2025efkan} \\
\addlinespace
DeepOHeat-v1 & 
Expanded DeepOHeat with Chebyshev KAN trunk; Used separable training and hybrid GMRES-based refinement for efficiency and reliability;
Applied to 2D power maps, 3D volumetric power distribution, and thermal optimization of floorplans.
& \cite{yu2025deepoheat}\\

\addlinespace
DeepOKKAN & 
Employed cKAN and KKAN variants with RBF-based trunk blocks within the DeepOKKAN framework;
Enhanced with QR-DeepONet reparameterization for improved stability and accuracy;
Applied to solving the 1D Burgers' equation.
&
\cite{toscano2025kkans}\\

\addlinespace
PO-CKAN & 
Proposed a physics-informed deep-operator framework combining DeepONet with Chunk-rational KAN (CKAN) modules;
Used CKAN for both branch and trunk nets;
Applied to Burgers’, Eikonal, and diffusion-reaction equations.
&
\cite{wu2025po}\\
\bottomrule
\end{tabularx}
\vspace{-10pt}
\end{table}

\begin{figure}[h!]
    \centering
    \includegraphics[width=1.0\linewidth]{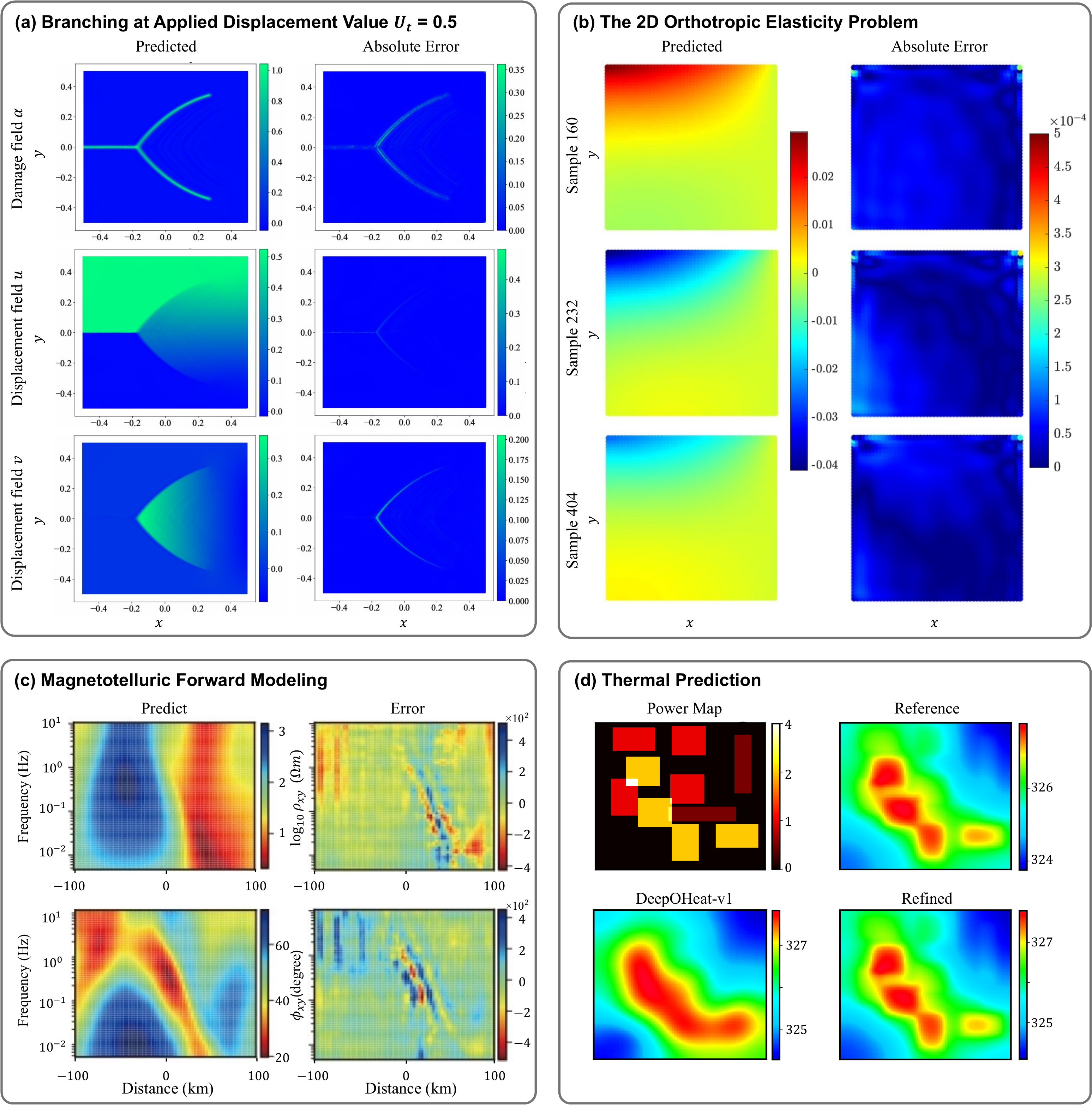}
    \caption{\textbf{(a)} Predicted crack branching and corresponding absolute error obtained using DeepOKAN. The model is trained using phase-field data from 45 randomly selected initial notches and tested on 5 unseen samples; the branch network processes the phase-field inputs, while the trunk network encodes spatial coordinates; each subnetwork contains 7 hidden layers with 100 nodes per layer, with 300 neurons in the final branch layer and 900 in the trunk output layer. This panel is adopted from~\cite{kiyani2025predicting}. 
    \textbf{(b)} Predicted solution and absolute error for the 2D orthotropic elasticity problem using DeepOKAN. The model, with a medium architectural complexity with approximately 7,200 trainable parameters. This panel is adopted from~\cite{abueidda2025deepokan}.
    \textbf{(c)} Predicted and absolute error of apparent resistivity and phase by EFKAN for a smooth resistivity model. The network is trained on data from 22 frequencies and 22 spatial locations, and evaluated on a denser grid of 64 frequencies and 64 coordinates. This panel is adopted from~\cite{wang2025efkan}.\textbf{(d)} Evaluation of predictive accuracy and refinement capability in operator learning. DeepOHeat-v1 employs a separable architecture with a shared branch network and three independent trunk networks based on Chebyshev KANs, while the hybrid method refines the neural predictions by correcting residuals using GMRES. This panel is adopted from~\cite{yu2025deepoheat}.
    }
    \label{fig:DeepOKAN1}
\end{figure}

In the world of solid mechanics, Kiyani et al.~\cite{kiyani2025predicting} used DeepOKAN to predict displacement and phase fields directly from initial crack geometry and boundary conditions, achieving high accuracy in 1D and 2D settings. The model bypasses traditional solver limitations by learning the operator mapping, resulting in reduced computational cost and data requirements. As shown in Fig.~\ref{fig:DeepOKAN1}(a), DeepOKAN captures steep gradients in damage and displacement fields with close agreement to reference solutions. Its ability to represent nonlinear multiscale features makes it a promising tool for efficient and accurate fracture prediction.

In the context of fast surrogate modeling for complex mechanical systems, Abueidda et al.~\cite{abueidda2025deepokan} proposed DeepOKAN, a surrogate model based on KANs with Gaussian RBFs, enabling efficient operator learning across diverse mechanical applications. The model has been applied in both data-driven and physics-informed settings, including 1D wave propagation, 2D linear elasticity, and time-dependent Poisson equations. These examples cover a range of static and dynamic problems. As shown in Fig.~\ref{fig:DeepOKAN1}(b), the predicted fields closely match reference solutions, indicating accurate performance across varying temporal and spatial domains. For geophysical electromagnetics,    Wang et al.~\cite{wang2025efkan} introduced the EFKAN framework, a neural operator framework that incorporates KANs into the projection layers. This design replaces standard MLP components to improve flexibility and representation in complex geophysical settings. The model has been tested on smooth and heterogeneous resistivity distributions generated via spectral methods, showing accurate predictions of apparent resistivity and phase responses (see Fig.~\ref{fig:DeepOKAN1}(c)). EFKAN offers an efficient surrogate for MT simulations governed by Maxwell’s equations, suitable for integration into data-driven electromagnetic inversion workflows. In the area of electronics and heat management,  Yu et al.~\cite{yu2025deepoheat} adaptively capture multiscale thermal patterns using the DeepOHeat-v1 framework. DeepOHeat-v1 is highly applicable to thermally aware floor planning and optimization in solid-state electronic systems. Fig.~\ref{fig:DeepOKAN1}(d) illustrates the refinement process for thermal field predictions, where initial operator network outputs exhibit notable discrepancies. The refinement corrects these inaccuracies by leveraging the error feedback, producing thermally consistent solutions aligned with the ground truth.

\subsection{Comparison with Other Operators}
\label{sec:deepokan_comparison}

We compare DeepOKAN and DeepONet to evaluate their performance in operator learning tasks. The comparison focuses on two key criteria: prediction accuracy and training convergence. Accuracy measures how well each model approximates the true operator across different input functions. Convergence analysis tracks how quickly and stably each model minimizes training error over time. These metrics reveal differences in generalization behavior and optimization efficiency, providing insight into the practical trade-offs between the two frameworks.

\begin{table}[h]
\centering
\small
\renewcommand{\arraystretch}{1.05}
\caption{Performance comparison between DeepONet and DeepOKAN models for solving the Darcy Flow problem in 1D and 2D settings. The table reports the number of trainable parameters ($|\boldsymbol{\theta}|$), basis functions used, and relative $\mathcal{L}^2$ errors under varying noise levels (0\%, 5\%, 10\%).}
\label{Tab:ComparisonDeepOKAN}
\footnotesize
\begin{tabular}{p{1.5cm} p{2.5cm} p{2cm} p{2cm} p{1cm} p{1cm} p{1cm} p{1cm}}
\toprule  
Dimension & Method & Basis & $|\boldsymbol{\theta}|$ & \multicolumn{3}{c}{Relative $\mathcal{L}^2$ Error} & Ref \\
\cmidrule(lr){5-7}
& & & & 0\% & 5\% & 10\% & \\

\midrule
1~D & DeepONet & SiLU & Shallow & 2.89 \%  & 5.58 \% & 9.42 \%&\cite{pant2025mlps} \\ 
 & DeepOKAN & B-spline& Shallow & 0.39 \%  & 4.34 \% & 8.61 \%& \\ 
 & DeepONet &SiLU & Deep & 1.26 \%  & 4.67 \% & 8.80 \%& \\ 
 & DeepOKAN & B-spline & Deep & 0.42 \%  & 4.35 \% & 8.61 \%& \\ 
\addlinespace
\addlinespace
2~D & DeepONet&tanh & 147000 & 1.62 \%  & 2.25 \% & 3.47 \%&\cite{shukla2024comprehensive} \\
 & DeepOKAN&Chebyshev & 585200 & 2.18 \%  & 2.20 \% & 2.30 \%& \\
\bottomrule
\end{tabular}
\vspace{-10pt}
\end{table}

\begin{figure}[h]
    \centering
    \includegraphics[width=1.0\linewidth]{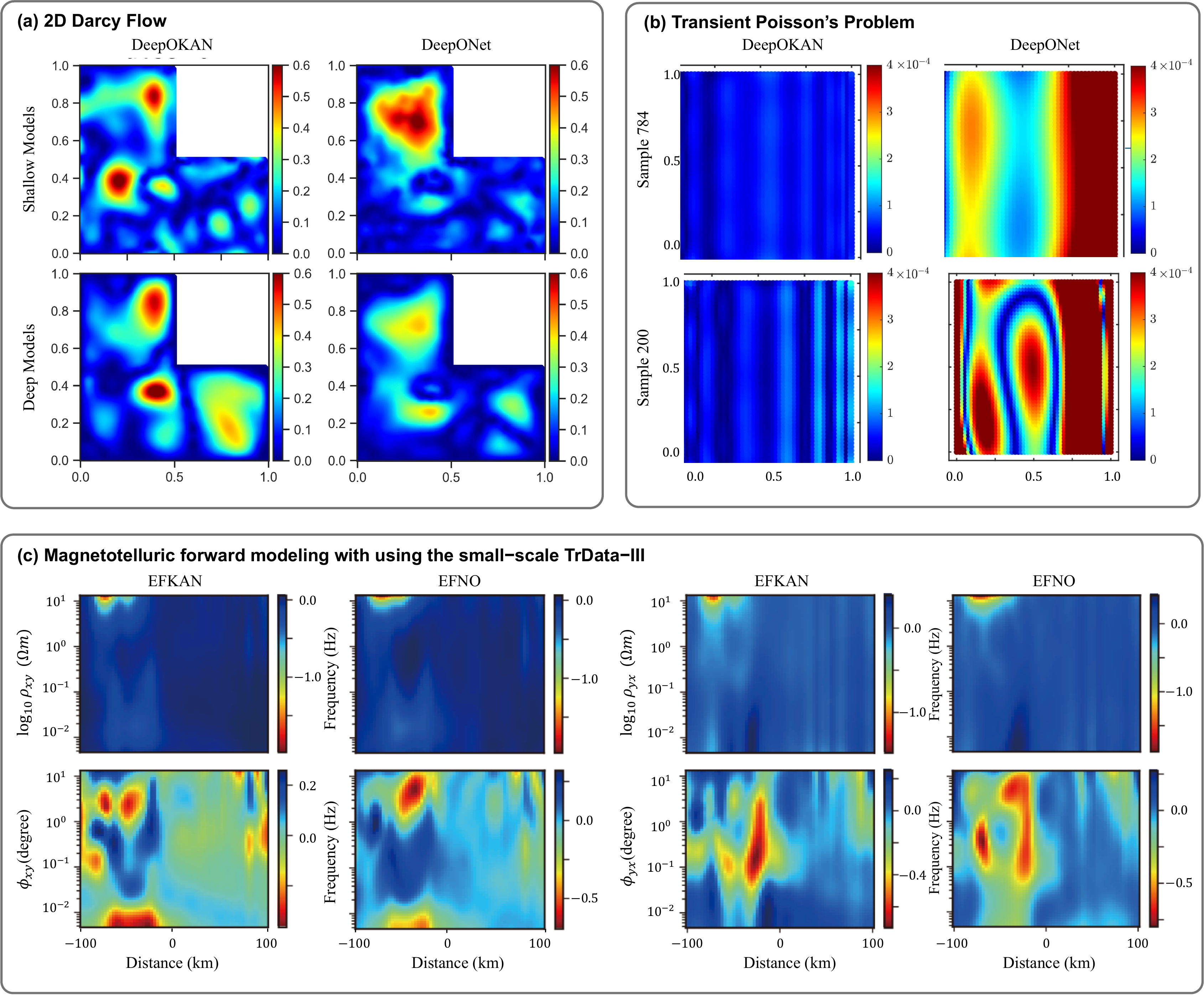}
    \caption{Comparison of prediction accuracy between different operator learning frameworks, focusing on error behavior under varying architectural depths, complexity levels, and physical settings.\textbf{(a)} Absolute error maps highlighting the different network architectures on prediction accuracy in heterogeneous flow fields. This panel is adapted from~\cite{pant2025mlps}.
    \textbf{(b)} Absolute error compared to the FEA ground-truth in a high-complexity case with 50600 learnable parameters for DeepOKAN and 51204 learnable parameters for DeepONet. This panel is adapted from~\cite{abueidda2025deepokan}.
    \textbf{(c)} Error using TeData-III containing 100 resistivity profiles with rectangular anomalies, simulated at 64 frequencies (0.005–12.589 Hz). This panel is adapted from~\cite{wang2025efkan}.}
    \label{fig:DeepOKANaccuracy}
\end{figure}

\subsubsection{Accuracy Comparison}

Evaluating the accuracy of operator learning models requires a careful examination of their behavior under diverse conditions. In this section, we analyze the predictive accuracy of DeepOKAN and DeepONet across a range of representative scenarios to understand how architectural and data-related factors influence their performance. Specifically, we consider both shallow and deep model configurations to assess how network depth impacts solution fidelity. We also examine how the number of samples influences the accuracy of DeepOKAN and DeepONet, highlighting the impact of data availability on model generalization across operator learning tasks. In addition, the effect of noisy input-output data is investigated to evaluate model robustness under realistic conditions. We also explore how different architectural designs, such as incorporating Fourier neural operators, affect the models' ability to capture complex functional relationships. A summary of key results is provided in Table~\ref{Tab:ComparisonDeepOKAN} and Fig.~\ref{fig:DeepOKANaccuracy}. This comparative analysis aims to offer insight into the conditions under which each model architecture maintains high accuracy and what factors contribute most significantly to performance variation.

Understanding how operator learning models behave across varying network depths is crucial for evaluating their expressive power and computational efficiency. 
Pant et al.~\cite{pant2025mlps} systematically studied the behavior of shallow and deep architectures using KAN and MLP-based models within operator learning frameworks. Table~\ref{Tab:ComparisonDeepOKAN} presents a summary of their results on the 1D Darcy flow problem. In shallow configurations, the KAN-based DeepOKAN shows a significant reduction in error compared to the MLP-based DeepONet, improving accuracy by over 85\%. In deeper settings, DeepOKAN also performs better, with an error reduction of approximately 65\%, although the performance gap becomes narrower as both models benefit from increased depth.
Furthermore, Fig.~\ref{fig:DeepOKANaccuracy}(a) illustrates the model outputs for the more complex 2D Darcy flow scenario. The visualizations reveal that DeepOKAN retains strong performance in shallow architectures, capturing key spatial patterns with fewer parameters. This supports the conclusion that shallow KAN models offer high expressiveness and efficiency, making them well-suited for physics-based learning tasks that demand accuracy without high computational overhead.

One important factor influencing the accuracy of operator learning methods is the number of available data samples. In many practical scenarios, obtaining large amounts of data may not be feasible due to cost, time, or physical limitations. Therefore, a desirable model should maintain high predictive accuracy even when the number of samples is limited. In particular, some works such as~\cite{wang2025efkan} and~\cite{abueidda2025deepokan} have reported the effect of sample size on model performance in specific examples. Abueidda et al.~\cite{abueidda2025deepokan} reported the behavior of DeepOKAN and DeepONet under different numbers of data samples, providing useful insights into how sample count affects model performance. Fig.~\ref{fig:DeepOKANaccuracy}(b) presents the prediction errors of DeepONet and DeepOKAN for the Transient Poisson’s problem under different sample settings. The results show that while the performance of DeepONet degrades significantly with fewer samples, DeepOKAN maintains relatively stable accuracy.

In real-world physics-based learning tasks, input data is often contaminated with noise, making it essential to assess the robustness of operator learning models under such conditions. Understanding how models respond to increasing noise levels provides insights into their generalization ability and reliability in practical scenarios. This aspect has been explored in studies~\cite{pant2025mlps, shukla2024comprehensive} through the solution of the Darcy flow problem under varying levels of input noise. Table~\ref{Tab:ComparisonDeepOKAN} summarizes their results across different dimensions and model configurations. In the 1D case, as noise levels increase from 0\% to 10\%, DeepOKAN consistently maintains lower error compared to DeepONet. Specifically, in shallow architectures, DeepOKAN shows improvements of approximately 22\% at 5\% noise and around 9\% at 10\% noise. For deep models, the trend persists with DeepOKAN maintaining a smaller error, showing a roughly 7\% advantage at 5\% noise and 2\% at 10\% noise. These findings suggest that while both models experience degraded performance as noise increases, DeepOKAN demonstrates more robust behavior, especially in shallower settings. In the 2D Darcy flow problem, while DeepONet initially shows lower error under noise-free conditions, its performance degrades more sharply with added noise. In contrast, DeepOKAN exhibits stable accuracy across all tested noise levels, with only a slight increase in error, indicating better resilience to noisy inputs in this case.

Exploring different architectural designs is crucial for enhancing the flexibility and expressiveness of operator learning models. One promising direction involves integrating KANs with advanced structures such as Fourier layers. For example, Wang et al.~\cite{wang2025efkan} proposed EFKAN, which combines KAN with FNO. Their results show that EFKAN achieves higher accuracy in estimating both apparent resistivity and phase. Fig.~\ref{fig:DeepOKANaccuracy}(c) illustrates that both EFNO and EFKAN can produce reasonable results for apparent resistivity and phase. 
However, on average, EFKAN reduces the relative $\ell_1$-norm error by around 20–25\% across the different components, particularly improving phase accuracy without sacrificing performance on resistivity. Therefore, when integrated into more advanced operator frameworks like EFKAN, KANs are a compelling choice for tasks in computational physics.

Overall, DeepOKAN outperforms DeepONet in accuracy across multiple benchmarks, such as different network depths, sample sizes, and noise levels. Its superior performance is especially pronounced in shallow architectures, where the use of structured basis functions allows for efficient learning with fewer parameters. In contrast, DeepONet's reliance on MLPs results in greater sensitivity to data limitations and noise. Moreover, enhanced architectures like EFKAN, which integrate KANs with FNO, further enhance accuracy, particularly for complex tasks involving spatially varying physical properties. These findings indicate that DeepOKAN offers a more robust and precise solution for operator learning, making it a suitable choice for real-world physics-informed modeling where reliability and generalization are essential.

\begin{figure}
    \centering
    \includegraphics[width=1.0\linewidth]{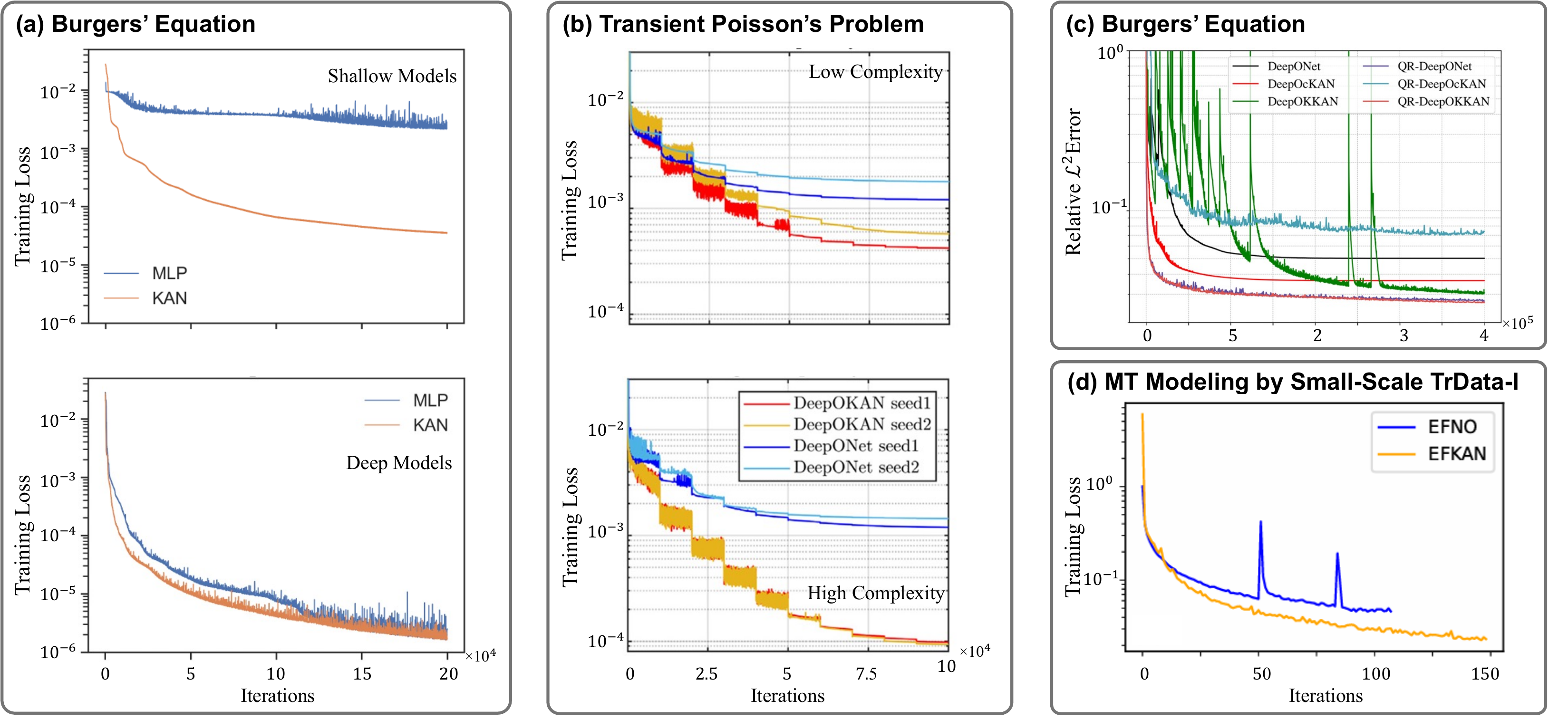}
    \caption{Convergence behavior of DeepOKAN and other operator networks across various problems. \textbf{(a)} Comparison of training loss for Burgers’ equation with fixed viscosity (0.01) and random initial conditions sampled from a Gaussian random field. The shallow network uses one hidden layer for both the branch and trunk, while the deep version uses 7 MLP layers and 4 KAN layers in each. This panel is reproduced from~\cite{pant2025mlps}. \textbf{(b)} Training loss curves for the transient Poisson equation with zero source and initial conditions. High-complexity DeepOKAN and DeepONet models have 50600 and 51204 parameters, respectively; low-complexity versions have 12650 and 13104 parameters. This panel is reproduced from~\cite{abueidda2025deepokan}. \textbf{(c)} Relative $\mathcal{L}^2$ error convergence on the test set for Burgers’ equation with lower viscosity ($1/(100\pi)$). DeepOKKAN, DeepOcKAN, and DeepONet models have 155920, 107328, and 131700 parameters, respectively; QR-based models use over 46000 parameters. This panel is reproduced from~\cite{toscano2025kkans}. \textbf{(d)} Training loss comparison of EFNO and EFKAN on the small-scale TrData dataset. This panel is reproduced from~\cite{wang2025efkan}.
    }
    \label{fig:DeepOKANconv}
\end{figure}

\subsubsection{Convergence Analysis}

Understanding convergence behavior is essential for evaluating how efficiently operator learning models approximate solutions during training.
To compare the learning efficiency of DeepOKAN and DeepONet, we analyze their convergence behavior across several settings.
First, we compare shallow and deep architectures to investigate how network depth influences the speed and stability of convergence.
Next, we explore different internal designs of KAN, such as variations in basis functions, to assess their impact on optimization dynamics.
In addition, we analyze an extended architecture where KAN is integrated into the FNO framework to investigate convergence behavior in a more expressive operator learning setting. Fig.~\ref{fig:DeepOKANconv} summarizes these convergence studies, highlighting how architectural choices and model compositions affect training efficiency across a range of scientific problems.

To understand the impact of network depth on convergence behavior, we compare the training dynamics of different complexities of models in operator learning. Many researchers have studied how network depth influences convergence behavior, such as~\cite{pant2025mlps, abueidda2025deepokan}, in the context of KAN and MLP-based models. Pant et al.~\cite{pant2025mlps} investigated the convergence behavior of shallow and deep networks by tracking the loss function during training. As shown in Fig.~\ref{fig:DeepOKANconv}(a), when solving the 1D Burgers’ equation, KAN-based models converge faster and reach lower final loss values compared to MLPs. In the shallow setting, MLPs show large loss values, strong oscillations, and early saturation, indicating unstable and inefficient learning. In contrast, shallow KANs not only converge rapidly but also achieve far lower and more stable loss levels. In the deep setting, while both models show reduced oscillations and more stable behavior, KAN still outperforms MLPs in terms of convergence speed and final loss. Abueidda et al.~\cite{abueidda2025deepokan} further examine convergence under varying levels of network complexity for both DeepOKAN and DeepONet models. Their study also includes two independent random seeds (seed1 and seed2) for weight initialization, ensuring robustness of the observations. Fig.~\ref{fig:DeepOKANconv}(b) presents the training loss profiles for solving the transient Poisson equation. Across all complexity levels and seeds, DeepOKAN consistently demonstrates lower training loss and faster convergence compared to DeepONet. While DeepONet tends to stabilize early, DeepOKAN maintains superior convergence speed and stability throughout training.

To evaluate how different internal designs of KANs influence model performance, several studies have explored architectural variations and their effects on convergence behavior in operator learning tasks~\cite{abueidda2025deepokan, shukla2024comprehensive, toscano2025kkans}. Toscano et al.~\cite{toscano2025kkans} investigate multiple extensions of the DeepONet framework by integrating MLPs, cKANs, and KKANs, leading to the development of models like DeepOKKAN. Their study incorporates RBFs within the KKAN structure and further enhances the models using the QR-DeepONet framework~\cite{lee2024training}, which improves training stability and convergence by reparameterizing the trunk network through QR decomposition. As shown in Fig.~\ref{fig:DeepOKANconv}(c), KKANs consistently outperform both cKANs and MLPs in terms of relative error convergence. While cKANs initially converge more rapidly, their performance plateaus over time. Notably, the QR formulation boosts the performance of MLPs and KKANs but has a detrimental effect on cKANs, suggesting that KKANs offer greater robustness and adaptability to optimization strategies. These findings underline the importance of architectural flexibility in achieving stable and efficient training within KAN-based operator learning models.

Integrating KAN into the FNO framework offers another perspective on improving convergence in operator learning. In the EFKAN architecture~\cite{wang2025efkan}, KAN replaces the MLP used in the traditional EFNO setup. As reported by Wang~\cite{wang2025efkan}, this modification leads to more stable training and improved convergence behavior. Fig.~\ref{fig:DeepOKANconv}(d) shows that EFNO tends to stop training early, potentially before reaching its optimal solution, while EFKAN continues to refine its predictions and ultimately achieves a lower loss. In addition to its improved learning dynamics, EFKAN also demonstrates faster computational performance compared to conventional numerical solvers like finite difference methods, indicating its potential as an efficient surrogate model for complex geophysical simulations such as magnetotelluric forward modeling.

The convergence analysis confirms that DeepOKAN consistently demonstrates faster and more stable training dynamics than DeepONet across various model depths, architectural designs, and integration frameworks. Whether in shallow or deep networks, DeepOKAN reaches lower loss values with fewer training iterations and demonstrates greater robustness to initialization and architectural changes. Enhanced variants of KAN, such as KKAN and EFKAN, further improve convergence properties by integrating advanced components. In contrast, DeepONet often shows slower convergence, early saturation, and less consistent performance. These results collectively demonstrate that DeepOKAN offers a more reliable and efficient optimization path, making it a strong candidate for operator learning tasks where training stability and convergence speed are critical.

\section{Challenges and Future Directions}
\label{sec:deepokan_challenges}

While the Kolmogorov–Arnold Network frameworks offer a mathematically grounded approach for approximating high-dimensional functions through compositions of simple univariate transformations, several challenges still hinder its broader adoption. These include scalability limitations due to computational cost, interpretability issues, sensitivity to hyperparameters, and difficulties in applying KANs to complex physical systems. In this section, we highlight key challenges and discuss promising directions for future research. These include strengthening the theoretical foundations, improving performance on complex geometries, and validating KANs on large-scale, real-world engineering and industrial applications.

\subsection{Methodological Challenges}

\begin{figure}[!h]
    \centering
    \includegraphics[width=1.0\linewidth]{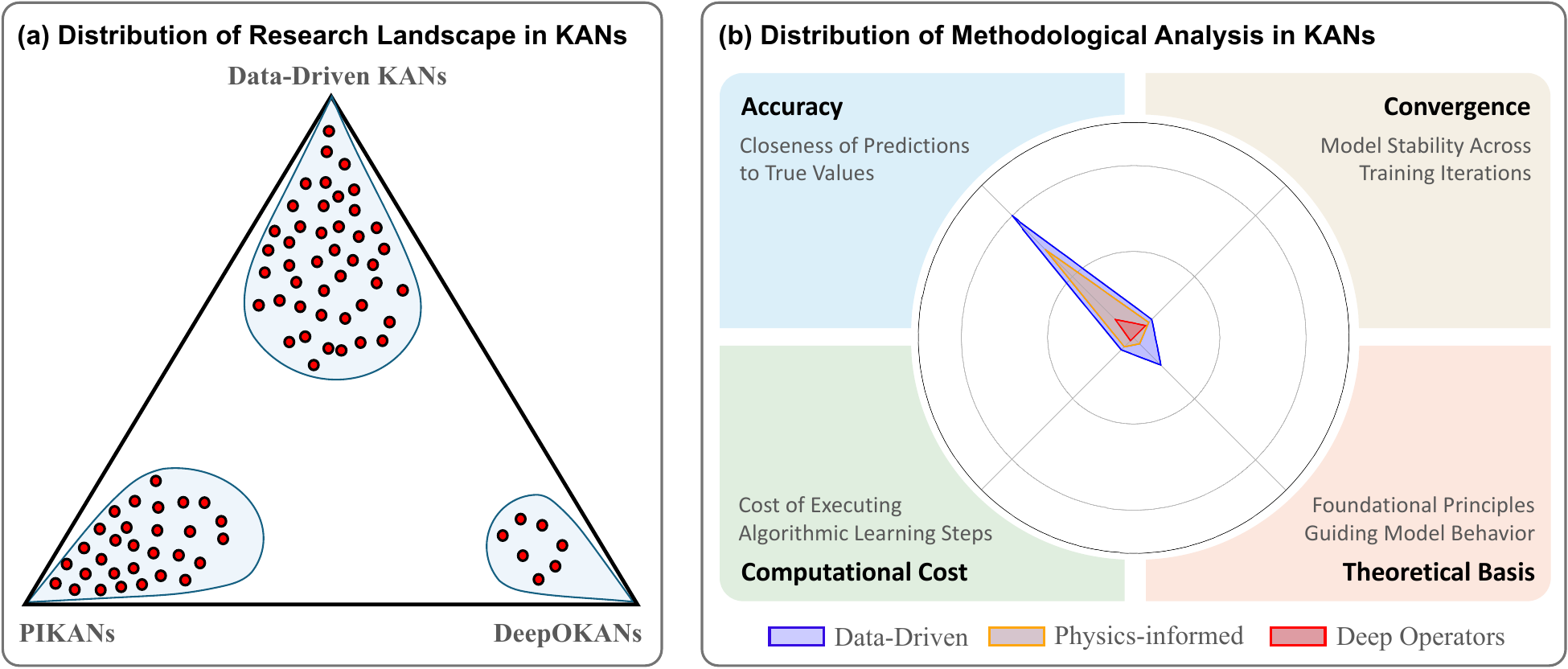}
    \caption{\textbf{(a)} Ternary plot showing the distribution of recent KAN-based studies across three main categories: purely Data-Driven KANs, PIKANs, and DeepOKANs. Most works emphasize data-driven modeling and the incorporation of physical constraints, with relatively limited attention to operator learning.
    \textbf{(b)} Comparative analysis of methodological challenges, accuracy, convergence, theoretical analysis, and computational cost, faced by each category. While most studies prioritize improving accuracy and, to a lesser extent, convergence, all three approaches still require significant advancements in theoretical foundations and reduction of computational cost.}
    \label{fig:FinalPic}
\end{figure}

It is essential to acknowledge the limitations of KANs when compared with MLPs. As comprehensively reviewed by Ji et al.~\cite{ji2024comprehensive}, several studies have shown that the performance of KANs is not universally superior. Zeng et al.~\cite{zeng2024kan} demonstrated that while KANs may exhibit faster convergence and improved fitting on certain irregular or noisy functions, they tend to underperform relative to MLPs for functions with jump discontinuities or non-differentiable points. Their results also indicate that KANs are more sensitive to noise, with test errors increasing noticeably under higher noise levels.  KANs generally require significantly greater computational cost and hardware resources, and in some tasks, the training time can exceed that of MLPs by up to two orders of magnitude. In addition, Pourkamali-Anaraki~\cite{pourkamali2024kolmogorov} reported that KANs experience substantial degradation in low-data regimes, particularly when data are scarce or unevenly distributed. Fig.~\ref{fig:FinalPic} provides an overview of the current landscape of KAN research and highlight key methodological challenges across different model categories. Among the most pressing issues are the high computational cost associated with KANs’ layered structure, the sensitivity of their performance to various hyperparameter choices, and the complexity of their algorithmic design, especially in terms of numerical stability and integration into standard machine learning environments. Table~\ref{Tab:summary} also presents a comparative summary of seminal KAN developments in terms of these four main methodological challenges across data-driven, physics-informed, and deep-operator learning paradigms. In what follows, we examine these challenges in detail and discuss recent approaches proposed to improve the efficiency, robustness, and usability of KANs across a range of real-world applications.

\begin{table}[h!]
\centering
\caption {Comparative summary of major KAN developments in the context of methodological challenges across data-driven, physics-informed, and deep-operator learning paradigms.}
\label{Tab:summary}
\footnotesize
\begin{tabular}{p{3.0cm}p{1.5cm} p{1.5cm} p{2.0cm} p{2.0cm} p{1.0cm} p{0.5cm}}
\toprule  
  {Model Name} & {Accuracy} & {Convergence}& {Theoretical Analysis}&{Computational Analysis} &{Code}& {Ref} \\ 
\midrule   
\addlinespace
\multicolumn{7}{l}{(I) Data-driven}\\
\addlinespace
\hline
\addlinespace
KAN &$\checkmark$ &$\checkmark$ &$\checkmark$ &$\checkmark$ &
\href{https://github.com/KindXiaoming/pykan}{\textcolor{blue}{\faGithub}} 
& \cite{liu2024kan} \\
\addlinespace
Chebyshev KAN &$\checkmark$ &$\checkmark$ &$\times$ &$\times$ &
\href{https://github.com/SynodicMonth/ChebyKAN}{\textcolor{blue}{\faGithub}} 
& \cite{ss2024chebyshev} \\
\addlinespace
FastKAN (RBF) &$\checkmark$ &$\times$ &$\times$ &$\checkmark$ &
\href{https://github.com/ZiyaoLi/fast-kan}{\textcolor{blue}{\faGithub}} 
& \cite{li2024kolmogorov} \\
\addlinespace
Wav-KAN &$\checkmark$ &$\times$ &$\times$ &$\times$ &
\href{https://github.com/zavareh1/Wav-KAN}{\textcolor{blue}{\faGithub}} 
& \cite{bozorgasl2405wav} \\
\addlinespace
\hline
\addlinespace
\multicolumn{7}{l}{(II) Physics-informed} \\
\addlinespace
\hline
\addlinespace
PIKAN& 
$\checkmark$ & $\checkmark$ & $\checkmark$ & $\checkmark$ &
\href{https://github.com/yizheng-wang/Research-on-Solving-Partial-Differential-Equations-of-Solid-Mechanics-Based-on-PINN}{\textcolor{blue}{\faGithub}} 
& \cite{wang2024kolmogorov} \\
\addlinespace
SPIKAN& 
$\checkmark$ & $\checkmark$ & $\checkmark$ & $\checkmark$ &
\href{https://github.com/pnnl/spikans}{\textcolor{blue}{\faGithub}} 
& \cite{jacob2024spikans} \\
\addlinespace
Scaled-cPIKAN& 
$\checkmark$ & $\checkmark$ & $\checkmark$ & $\times$ &
\href{https://github.com/sfaroughi3/Pub_Scaled_cPIKAN}{\textcolor{blue}{\faGithub}} 
& \cite{mostajeran2025scaled} \\
\addlinespace
KKANs& 
$\checkmark$ & $\checkmark$ & $\times$ & $\checkmark$ &
\href{https://github.com/jdtoscano94/Kurkova_Kolmogorov_Arnold_Networks_KKANs}{\textcolor{blue}{\faGithub}}
& \cite{toscano2025kkans} \\
\addlinespace
\hline
\addlinespace
\multicolumn{7}{l}{(III) Deep-operators}\\
\addlinespace
\hline
\addlinespace
DeepOKAN& 
$\checkmark$ & $\checkmark$ & $\times$ & $\times$ &
\href{https://github.com/DiabAbu/DeepOKAN}{\textcolor{blue}{\faGithub}}
& \cite{abueidda2025deepokan} \\
\addlinespace
Hybrid Operators& 
$\checkmark$ & $\checkmark$ & $\times$ & $\times$ &
$\times$ 
& \cite{kiyani2025predicting} \\
\bottomrule
\end{tabular}
\end{table}

One significant challenge in implementing the KAN lies in its high computational cost, which arises from the complexity of learning multivariate functions through the composition and addition of univariate functions. This formulation leads to intricate hierarchical structures and adaptive activation mechanisms that require intensive optimization. As a result, training KANs often becomes time-consuming, particularly when applied to high-dimensional problems that demand precise approximations. Fine-grained adjustments and iterative refinements are typically needed to capture the target mappings accurately, further increasing computational overhead. These demands limit the scalability of KANs in large-scale or real-time applications, where computational efficiency is essential. To address these challenges, several recent studies have proposed strategies aimed at reducing both training time and memory usage. For example, the development of separable PIKANs (SPIKANs)~\cite{jacob2024spikans} has shown that leveraging the separability of physical laws allows for simplified network architectures, resulting in significantly lower computational costs. Another promising direction is domain decomposition~\cite{howard2024finite, faroughi2025neural}, where the problem is divided into smaller subdomains, each handled by a reduced-size network. This approach enables parallel or distributed training and decreases the overall complexity of the model, making KANs more suitable for large or distributed systems.
Moreover, within deep-operator frameworks, Wu \& Lin~\cite{wu2025po} introduced the Chunk-rational KAN (CKAN) and integrated it into the Physics-informed deep-operator architecture to enhance scalability and enforce physical consistency in PDE-based learning. Their study demonstrated that conventional KANs exhibit quadratic parameter growth and extremely high floating-point operations (FLOPs). 
By incorporating chunk-wise parameter sharing, 
CKAN achieves a reduction in FLOPs by approximately three orders of magnitude while preserving parameter complexity comparable to MLP baselines and maintaining, or even improving, predictive accuracy, with only minor increases in inference latency.
While these advances have led to meaningful improvements, achieving broader scalability and faster training remains an active challenge. Continued refinement of architectural principles, training procedures, and hardware-aware implementations will likely play an important role in unlocking the full potential of the Kolmogorov–Arnold framework in applied machine learning contexts.

Another key challenge in KANs lies in their strong sensitivity to hyperparameters, especially those that govern the design and refinement of their compositional architecture. Unlike conventional neural networks, which typically rely on fixed activation functions and standard layer configurations, KANs require precise control over the complexity of univariate basis functions, their adaptive combinations, and the depth of hierarchical compositions. Without well-defined heuristics, setting these hyperparameters often involves a manual and computationally expensive trial-and-error process. Poor choices can result in underfitting due to insufficient representation power or, conversely, lead to excessive training costs due to over-parameterization. In addition, the evolving nature of function approximation within KANs introduces the need for continual adjustment, as the ideal hyperparameter values may shift during training in response to the model’s internal dynamics. To support more efficient and systematic hyperparameter tuning, several strategies rooted in automatic optimization have been proposed. Bayesian optimization techniques~\cite{snoek2012practical}, for instance, offer a probabilistic approach to exploring the hyperparameter space while minimizing the number of costly model evaluations. These methods are particularly well-suited to KANs, where each training run can be resource-intensive. Similarly, gradient-based hyperparameter optimization frameworks~\cite{lorraine2020optimizing} have shown potential for adapting hyperparameters by leveraging gradient signals from the training loss, which could be customized to align with KANs’ compositional structure. Such methods reduce the reliance on exhaustive grid searches and improve efficiency by guiding tuning based on the network’s evolving behavior. By combining these algorithmic tools with a deeper understanding of KAN-specific design principles, researchers are beginning to develop more robust and adaptive approaches to managing complexity in training. Nevertheless, refining these strategies to better match the unique architectural features of KANs remains an area with considerable potential for further refinement and insight.

The third challenge in the practical use of KANs is their lack of built-in support in mainstream deep learning frameworks, such as TensorFlow and PyTorch, which complicates their integration into standard deep learning workflows and limits computational efficiency. The recursive, loop-intensive nature of KANs, stemming from the layered composition of univariate functions, introduces practical difficulties during implementation, particularly when scaling to large models or deploying them in parallel computing environments. These recursive operations often lead to non-trivial memory access patterns and inefficient execution on modern hardware accelerators, which are optimized for regular, feedforward computations. This not only exacerbates challenges related to training time and hyperparameter tuning but also limits the ability to exploit GPU-level parallelism effectively. One promising approach to mitigate this issue involves reformulating recursive expressions using algebraically equivalent, non-recursive structures. For instance, in the case of Chebyshev polynomials, a triangular matrix formulation can replace recursive evaluations, enabling more efficient and parallelizable computations.
However, these functions, while theoretically valuable for representing complex nonlinearities, can suffer from numerical instability, particularly when evaluated outside stable intervals or in the presence of noisy gradients. Such issues are especially pronounced in physics-informed applications or differential equation solvers, where accuracy is critical. To improve robustness, researchers have introduced normalization and rescaling techniques that map input variables to compact domains such as $[-1,1]$. This strategy reduces numerical error and maintains stability during training, especially in models involving trigonometric components or sensitive differential operators~\cite{mostajeran2025scaled, faroughi2025neural}. On the implementation side, efforts have also been made to develop more practical and accessible KAN tools. For example, the construction of dedicated KAN modules with optimized computational kernels can streamline training and inference, while minimizing unnecessary overhead. Integration with widely-used libraries through user-friendly application programming interfaces (APIs) and visual debugging tools would further facilitate adoption among practitioners and researchers alike. These enhancements are essential not only for improving usability but also for encouraging more widespread experimentation with KANs in diverse application areas.

\subsection{Future Direction}

Future research directions aim to build on the current progress to further enhance the impact of PIKANs and DeepOKANs.
We identify three directions to support the advancement.
The first focuses on establishing stronger theoretical foundations, including tools to analyze generalization, convergence, and interpretability. The second aims to improve the handling of complex geometries and mesh representations, enabling these models to operate effectively on irregular domains common in real-world applications. The third direction emphasizes industrial-scale validation, where large-scale simulations, computational efficiency, and physical consistency must be addressed to move from academic studies to practical deployment. Together, these directions identify critical areas where continued development could make KAN-based models more robust, scalable, and widely applicable.

The first research direction focuses on developing a rigorous theoretical foundation to better understand the generalization behavior and learning dynamics of KAN-based models.
Studies investigating the feature-learning properties of KANs, mainly through NTK analysis, remain limited and are mostly confined to specific cases~\cite{mostajeran2025scaled, faroughi2025neural, rigas2025initialization}.
Prior work such as~\cite{faroughi2025neural} has investigated the training behavior of cPIKANs using the Neural Tangent Kernel framework, providing valuable insights into how architectural choices influence optimization. However, this analysis is limited to a specific case and does not extend to the broader family of PIKAN models  or more complex variants like DeepOKANs.A deeper theoretical understanding is required to characterize how the NTK evolves with model complexity and to develop formal connections between the NTK spectrum and prediction error. In addition, NTK-based perspectives represent only one possible lens; other theoretical approaches, such as approximation theory, stability analysis, or generalization error bounds, are essential for a more complete understanding of these models. Beyond NTK-based analyses, some studies have explored the theoretical foundations of practical KANs from an approximation-theoretic perspective, establishing optimal approximation guarantees~\cite{kratsios2025kolmogorov}, explicit error bounds~\cite{warin2024p1}, and constructive results that mitigate the curse of dimensionality for various spline- and polynomial-based architectures~\cite{fakhoury2025expressivity}. Similarly, studies that establish formal generalization bounds for PINNs in inverse PDE settings~\cite{mishra2022estimates} or derive approximation error estimates for operator networks like FNOs~\cite{kovachki2021universal} illustrate the type of theoretical analysis that is still lacking for PIKANs and DeepOKANs. In particular, formal characterizations of how model complexity influences generalization, convergence rates, and expressiveness remain largely underdeveloped. Building such theoretical tools could lead to a deeper understanding of KANs and guide the principled design, analysis, and application of these networks across a wide variety of tasks.

The second research direction addresses the challenge of applying KAN-based models to irregular domains and complex geometries that frequently arise in scientific and engineering contexts. The second research direction focuses on enhancing the ability of KAN-based models to operate effectively on irregular domains and complex geometries commonly encountered in scientific and engineering problems. While the Kolmogorov–Arnold structure at the core of PIKANs and DeepOKANs is inherently geometry-agnostic, practical implementations often rely on regular, rectangular grid structures, limiting performance on complex domains. As shown in~\cite{wang2024kolmogorov}, PIKANs demonstrates limited improvement over traditional MLP-based approaches when applied to irregular geometries, despite its effectiveness on structured ones. A key constraint stems from the rectangular nature of the KAN grid in high dimensions, which aligns more naturally with simple geometries. Moreover, a trade-off exists between capturing geometric detail and maintaining manageable resolution: the more intricate the domain, the finer the grid must be, increasing computational cost and complexity. This tension complicates the balance between geometric fidelity and model simplicity. To improve performance on complex domains, Wang et al.~\cite{wang2024kolmogorov} propose several strategies drawn from finite element methods, including h-refinement to locally adapt the grid, isoparametric transformations to map curved domains to simpler reference shapes, and conformal mappings to reparameterize irregular geometries into more manageable ones. While such techniques are promising for PIKANs, handling geometry in DeepOKANs remains relatively underexplored. In particular, although KAN layers are geometry-agnostic, practical deployment in deep-operator networks often requires coordinate mappings for trunk inputs or boundary interpolation for the branch network. Embedding richer geometry encodings, such as signed distance functions or shape parameterizations, could help bridge this gap. Furthermore, combining these approaches with domain decomposition or overlapping local expansions may offer a pathway toward more robust and scalable solutions in complex geometrical settings.

The third research direction aims to bridge the gap between current academic implementations of KAN-based models and their deployment in large-scale, real-world engineering and industrial systems. Most current evaluations are conducted on academic-scale PDEs or controlled simulations, often in low-dimensional settings. To transition these models toward real-world deployment, extensive validation on high-performance computing (HPC) platforms is essential, particularly to assess scalability, memory efficiency, parallel training performance, and long-term robustness under real-world conditions. Some early progress in this direction has been made through the introduction of ICKANs~\cite{thakolkaran2025can}, which were successfully used to model hyperelastic material behavior directly from measurable data such as strain fields and reaction forces. These models demonstrated strong generalization in finite element simulations involving complex geometries and noise-prone data, confirming the potential of physics-informed KAN-based models in engineering applications. However, broader adoption in large-scale simulations, such as those encountered in digital twin systems~\cite{abueidda2025deepokan}, design optimization, and uncertainty quantification, will require continued effort in algorithm-hardware alignment, model generality, and physical consistency across diverse scenarios. Advancing this line of research could enable KAN-based models to move beyond academic proof-of-concept and toward robust, scalable tools for industrial use.

\section{Conclusions}\label{Sec.Conclusion}

The integration of Kolmogorov–Arnold Networks into scientific machine learning represents a shift toward models that are structurally aligned with the nature of the problems they aim to solve. This idea, rooted in a classical mathematical theorem, leads to models that are often more interpretable and structured than standard neural networks. KANs have been successfully applied in data-driven tasks, physics-informed settings, and operator learning frameworks, showing promising results in a range of scientific problems. A key advantage of KANs is that their structure can be aligned with the nature of the problem, rather than relying on a fixed network design or purely numerical optimization. This allows the model to better capture underlying patterns, such as physical laws or compositional relationships, in a more meaningful way. This perspective opens the door to more customized, efficient, and explainable models in science and engineering. However, their advantages come with new challenges. Training KANs requires a more careful design, additional computational resources, and a deeper understanding of how their structure affects performance. These challenges are different from those of traditional neural networks and require new tools and methods for efficient implementation and analysis. In general, KANs offer a solid basis for constructing models that are both well-structured and physically coherent in scientific machine learning. Ongoing research will be essential for advancing concepts underlying KANs, such as modular architectures, the integration of prior knowledge, and the ability to precisely control function approximations. These innovations may also pave the way for novel model types that extend beyond the scope of KANs.

\section{Acknowledgements}

S.A.F. acknowledges support from the U.S. Department of Energy, Office of Environmental Management (Award No. DE-EM0005314), and the U.S. National Science Foundation under the Collaborations in Artificial Intelligence and Geosciences (CAIG, Award No. DE-2530611).

\section{Conflict of Interest}
The authors declare no conflict of interest.

\def\mybibdoicolor{\color{black}}
\newcommand*{\doi}[1]{\href{\detokenize{#1}} {\raggedright\mybibdoicolor{DOI: \detokenize{#1}}}}

\bibliographystyle{elsarticle-num}
\bibliography{references_new}


\end{document}